\documentclass[lettersize,journal]{IEEEtran}
\usepackage{amsmath,amsfonts}
\usepackage{algorithmic}
\usepackage{array}
\usepackage[table]{xcolor} 
\usepackage{tabularx}   
\usepackage{multirow}      
\usepackage{booktabs}      
\usepackage{bigstrut}      
\usepackage[caption=false,font=normalsize,labelfont=sf,textfont=sf]{subfig}
\usepackage{textcomp}
\usepackage{stfloats}
\usepackage{url}
\usepackage{verbatim}
\usepackage{graphicx}
\newcommand{\eg}{\textit{e.g.},~}
\newcommand{\ie}{\textit{i.e.},~}
\newcommand{\etal}{\textit{et al.}~}

\def\BibTeX{{\rm B\kern-.05em{\sc i\kern-.025em b}\kern-.08em
    T\kern-.1667em\lower.7ex\hbox{E}\kern-.125emX}}
\usepackage{balance}
\begin{document}

\title{A Tri-Modal Dataset and a Baseline System for Tracking Unmanned Aerial Vehicles }
\author{Tianyang Xu,
        Jinjie Gu,
        Xuefeng Zhu,
        Xiao-Jun Wu, and~Josef~Kittler,~\IEEEmembership{Life~Member,~IEEE}
\thanks{T. Xu, J. Gu, X. Zhu and X.-J. Wu are with the School of Artificial Intelligence and Computer Science, Jiangnan University, Wuxi, Jiangsu, China (E-mail: 
\{tianyang.xu; xuefeng.zhu; wu\_xiaojun\}@jiangnan.edu.cn;
6243112025@stu.jiangnan.edu.cn).}
\thanks{J. Kittler is with the School of Computer Science and Electronic Engineering and the Centre for Vision, Speech and Signal Processing, University of Surrey, Guildford, GU2 7XH, UK. (e-mail: j.kittler@surrey.ac.uk)}
}


\maketitle

\begin{abstract}
With the proliferation of low altitude unmanned aerial vehicles (UAVs), 
visual multi-object tracking 
is becoming a critical security technology, demanding significant robustness even in complex environmental conditions. 
However, tracking UAVs using a single visual modality often fails in challenging scenarios, \eg such as low illumination, cluttered backgrounds, and rapid motion. 
Although multi-modal multi-object UAV tracking is more resilient, the development of effective solutions has been hindered by the absence of dedicated public datasets. 
To bridge this gap, we release \textit{MM-UAV}, the first large-scale benchmark for Multi-Modal UAV Tracking, integrating three key sensing modalities, \ie RGB, infrared (IR), and event signals. 
The dataset spans over 30 challenging scenarios, with 1,321 synchronised multi-modal sequences, and more than 2.8 million annotated frames. 
Accompanying the dataset, we provide a novel multi-modal multi-UAV tracking framework, designed specifically for UAV tracking applications and serving as a baseline for future research. 
Our framework incorporates two key technical innovations, \ie an offset-guided adaptive alignment module to resolve spatio mismatches across sensors, and an adaptive dynamic fusion module to balance complementary information conveyed by different modalities. 
Furthermore, to overcome the limitations of conventional appearance modelling in multi-object tracking, we introduce an event-enhanced association mechanism that leverages motion cues from the event modality for more reliable identity maintenance.
Comprehensive experiments demonstrate that the proposed framework consistently outperforms state-of-the-art methods, particularly in challenging visual conditions, achieving breakthrough performance in complex low-altitude environments. 
To foster further research in multi-modal UAV tracking, both the dataset and source code will be made publicly available at \url{https://xuefeng-zhu5.github.io/MM-UAV/}.
\end{abstract}


\section{INTRODUCTION}
\IEEEPARstart{T}{he} rapid proliferation of unmanned aerial vehicles (UAVs) has significantly impacted various sectors, including civilian logistics, aerial cinematography, infrastructure inspection, emergency response, and environmental monitoring. 
Their adaptability and operational efficiency have rendered them indispensable across both commercial and industrial domains. 
However, this expansive use introduces critical security vulnerabilities, raising substantial concerns regarding personal privacy, public safety, and national security. 
Consequently, the development of robust UAV detection and tracking systems has emerged as an urgent technological imperative~\cite{DUT-Anti-UAV}.

As a fundamental component of low-altitude security systems, UAV tracking demands increasingly robust algorithms to counter evolving and sophisticated UAV threats. 
The core objective is to achieve continuous, real-time, and stable trajectory estimation of UAVs across diverse and challenging environments. 
In contrast to traditional radar~\cite{radar1} or radio frequency~\cite{RF} based countermeasures, vision-based UAV tracking has gained prominence, largely driven by significant advances in computer vision~\cite{YOLOX,detr,cheng2025one} and visual tracking algorithms~\cite{DeepSORT,bytetrack,botsort,trackformer,xu2020accelerated,kristan2023first}.

To address these challenges, a growing body of research~\cite{Anti-UAV,Anti-UAV410,Anti-UAV600,2th-antiuav,3th-antiuav,SiamSTA,GLTF-MA,UTTracker} has been dedicated to developing visual object tracking methods tailored for UAV-targeted applications. 
Nevertheless, the majority of these efforts remain confined to single visual modalities and single-object tracking (SOT) frameworks. 
Although multi-UAV tracking has recently attracted attention~\cite{MOT-FLY,Dist-Tracker,strong-baseline}, existing approaches still rely predominantly on a single type of sensor data.
Single-modality tracking is inherently limited by sensor-specific constraints, leading to notable performance bottlenecks in practical scenarios. 
For instance, RGB cameras, which operate in the visible spectrum, are highly susceptible to performance degradation under poor illumination conditions, such as at night, in heavy fog, or during rain, leading to significant noise and loss of detail. 
In contrast, infrared (IR) sensors detect thermal radiation and can operate in darkness, but are often susceptible to false positives caused by birds or other heat-emitting objects, while lacking the ability to discern textual or fine-grained visual features of UAVs. 
These limitations substantially reduce the operational effectiveness of single-modality trackers in real-world anti-UAV operations.

Similarly, SOT paradigms, which typically require manual initialisation of a single target in the first frame, preserving tracking of individual objects but are ill-suited for scenarios involving variable numbers of UAVs. 
In realistic UAV missions, threats may emerge, disappear, or re-enter the field of view abruptly and without prior knowledge, making manual initialisation impractical. 
Although methods such as~\cite{Anti-UAV600} attempt to reduce reliance on manual prior information, they remain limited to single-modality SOT tracking, thus cannot handle multi-object scenarios.
Therefore, multi-modal multi-object tracking (MM-MOT) presents a promising direction by leveraging complementary sensor information to enhance robustness across diverse conditions, while also supporting autonomous detection and tracking of multiple uncertain UAV targets. 
This capability is critical for responsive and scalable anti-UAV systems in dynamic and complex environments.

\begin{table*}[!t] 
\centering
\caption{Summary of UAV-targeted datasets, which encompasses datasets applicable to detection tasks (Det), single-object tracking tasks (SOT), and multiple-object tracking tasks (MOT).}
\label{tab:uav_datasets}
\renewcommand{\arraystretch}{1.5}

\begin{tabular}{
>{\centering\arraybackslash}m{3.4cm}  
>{\centering\arraybackslash}m{1.1cm}  
>{\centering\arraybackslash}m{1.9cm}    
>{\centering\arraybackslash}m{0.6cm}  
>{\centering\arraybackslash}m{0.7cm}    
>{\centering\arraybackslash}m{2.6cm}  
>{\centering\arraybackslash}m{1.8cm}  
>{\centering\arraybackslash}m{2.6cm}
}
\hline
\rowcolor{gray!5} 
\textbf{Name} & \textbf{Task} & \textbf{Modality} & \textbf{Seq} & \textbf{Frames} & \textbf{Resolution} & \textbf{Avg. UAV size} & \textbf{Years \& Venue} \\
\hline

\rowcolor{green!5} 
FL-Drones \cite{FL-Drones}          & Det   & RGB       & 14  & 39K  & 640\texttimes480-752\texttimes480 & -- & 2016 TPAMI \\
\rowcolor{green!5} 
NPS-Drones \cite{NPS-Drones}         & Det   & RGB       & 50  & 70K  & 1280\texttimes760-1920\texttimes1280 & -- & 2016 IROS \\
\rowcolor{green!5} 
Det-Fly \cite{Det-Fly}            & Det   & RGB       & --  & 13K  & 3840\texttimes2160 & -- & 2021 RA-L \\
\rowcolor{green!5} 
Drone-vs-Bird \cite{Drone-vs-Bird}      & Det   & RGB       & 77  & 105K & 720\texttimes576-3840\texttimes2160 & 34\texttimes23 & 2021 AVSS \\
\rowcolor{green!5} 
UETT4K Anti-UAV \cite{UETT4K-Anti-UAV}    & Det   & RGB       & 18  & 33K  & 3840\texttimes2160 & -- & 2025 IEEE Access \\
\hline

\rowcolor{yellow!5} 
USC Drone \cite{USC-Drone}          & SOT   & RGB       & 60  & 20K  & 1920\texttimes1080 & 114\texttimes83 & 2017 APSIPA ASC \\
\rowcolor{yellow!5} 
Anti-UAV \cite{Anti-UAV}          & SOT   & RGB, IR   & 348 & 297K & 1920\texttimes1080 (RGB)

640\texttimes512 (IR)& 125\texttimes63 (RGB)

52\texttimes30 (IR) & 2021 TMM \\
\rowcolor{yellow!5} 
Anti-UAV600 \cite{Anti-UAV600}       & SOT   & IR        & 600 & 723K & 640\texttimes512 & 30\texttimes19 & 2023 CoRR\\
\rowcolor{yellow!5} 
Anti-UAV410 \cite{Anti-UAV410}       & SOT   & IR        & 410 & 438K & 640\texttimes512 & -- & 2024 TPAMI \\
\rowcolor{yellow!5} 
Halmstad Drone \cite{Halmstad-Drone}    & Det/SOT & RGB, IR  & 650 & 203K & 640\texttimes512 (RGB)

320\texttimes256 (IR) & -- & 2021 ICPR \\
\rowcolor{yellow!5} 
DUT Anti-UAV \cite{DUT-Anti-UAV}      & Det/SOT & RGB       & 20  & 35K  & 160\texttimes240-3744\texttimes5616 & 70\texttimes31 & 2022 T-ITS \\
\hline

\rowcolor{blue!5} 
UAVSwarm \cite{UAVSwarm}          & MOT   & RGB       & 72  & 12K  & 446\texttimes276-1919\texttimes1079 & -- & 2022 Remote Sensing \\
\rowcolor{blue!5} 
MOT-Fly \cite{MOT-FLY}           & MOT   & RGB       & 16  & 11K  & 1920\texttimes1080 & -- & 2023 ICUS \\
\rowcolor{blue!5} 
4th Anti-UAV challenge \cite{4th-Anti-UAV-workshop} & MOT & IR        & 300 & 225K & 640\texttimes512 & -- & 2025 CVPR workshop \\
\hline
\textbf{MM-UAV (ours)}            & \textbf{MOT/SOT}   & \textbf{RGB,IR,Event} & \textbf{1321} & \textbf{2.8M} & \textbf{640}\texttimes\textbf{360} \textbf{(RGB)}

\textbf{640}\texttimes\textbf{512 (IR)}

\textbf{346}\texttimes\textbf{260} \textbf{(Event)} & \textbf{12}\texttimes\textbf{5} \textbf{(RGB)}

\textbf{15}\texttimes7 \textbf{(IR)} & \textbf{2025} \\
\hline

\end{tabular}
\end{table*}

The name of anti-UAV qualifying these datasets has understandably been coined in \cite{DUT-Anti-UAV,Anti-UAV,Anti-UAV410,Anti-UAV600,2th-antiuav,3th-antiuav,SiamSTA,GLTF-MA} to emphasise the defensive purpose of UAV tracking.
However, we shall not perpetuate the use of this term as it is ambiguous and counterintuitive. 
The task is to track simultaneously several very challenging objects, and fundamentally, it is a multi-object tracking problem. 
If tracked successfully, an appropriate action can be taken to neutralise any threat posed by the objects.
However, the discussion of any such action is beyond the scope of this paper. 
We focus simply on the topic of multi-UAV tracking, and its peculiarities, to advance multi-object tracking in general, and multi-UAV tracking in particular. 
In this context, we have collected the first large-scale tri-modal UAV tracking dataset. 
It can be used for multimodal tracking per se, but it can as well be used for related research topics, such as multimodal UAV detection, for studies of adversarial attacks on tracking systems, and for adversarial attack detection.

In general, multi-object tracking (MOT), aims to detect and localise multiple targets across video sequences while maintaining consistent identities over time~\cite{DeepSORT}. 
Compared to conventional MOT tasks such as pedestrian tracking~\cite{mot15,mot20}, multi-UAV tracking presents significantly greater challenges in identity preservation~\cite{Dist-Tracker,strong-baseline}. 
These difficulties stem from three primary factors.
First, UAVs often exhibit fast and irregular motion patterns, causing conventional Kalman filters~\cite{KF} to accumulate large prediction errors, which leads to identity mismatches in IoU-based association modules. 
Second, unlike pedestrians that often possess distinctive appearance features, UAVs typically share highly similar or identical visual characteristics, which greatly limits the discriminative power of appearance-based Re-ID models.
Third, UAVs typically appear as small objects with limited texture details, making them prone to being confused with background clutter, which results in elevated false negatives and false positives.

Due to these challenges, conventional MOT frameworks~\cite{DeepSORT,bytetrack,botsort,boostTrack,boostTrack++} that rely primarily on Kalman filtering and appearance embeddings exhibit considerable limitations in multi-UAV scenarios. 
In response, the emerging event modality visual sensing technology, capturing asynchronous, sparse motion streams with high temporal resolution~\cite{ceutrack,fe108,eventvot,TENet}, has recently gained increasing attention in visual tracking. 
By directly encoding intensity changes over time, event cameras enable robust motion-based reasoning that helps disambiguate small, fast-moving objects such as UAVs. 
This modality provides strong motion cues that complement conventional RGB or IR sensors, thereby reducing identity switches caused by appearance ambiguity or erratic motion patterns.

Given the current absence of multi-modal frameworks in existing multi-UAV tracking studies~\cite{MOT-FLY,strong-baseline,Dist-Tracker}, we propose a novel multi-modal multi-object tracking approach designed specifically to address the identity discrimination challenges in UAV scenarios. 
Our method, termed Multi-Modal-Aware Simple Online and Realtime Tracking (MMA-SORT),
effectively integrates RGB, IR, and event modalities within a unified tracking framework. 
Beyond conventional motion prediction using Kalman filters and appearance modelling via Re-ID features, we introduce an additional motion embedding derived directly from event data. 
This embedding captures fine-grained temporal dynamics that are complementary to appearance cues, significantly enhancing the discrimination of UAV targets with similar appearance and reducing identity switches caused by background clutter or erratic motion.
Notably, the proposed motion embedding is extracted in a training-free manner directly from raw event streams, avoiding the computational overhead associated with deep feature extraction, thereby guaranteeing high inference speed. 
The MMA-SORT framework effectively combines the strengths of three complementary sensing modalities, offering robust performance in challenging conditions where single-modality or conventional dual-modality trackers tend to fail. 

Furthermore, in multi-modal visual tasks that demand high localisation accuracy, spatio-temporal misalignment across modalities presents a critical challenge. 
Such misalignment arises from differences in imaging mechanisms, field of view, data transmission latency, and resolution among heterogeneous sensors~\cite{CSOM,align2}. 
While often tolerable for larger objects, even slight misalignments cause severe localisation errors for small UAV targets, which may span only a few pixels, significantly undermining the effectiveness of multi-modal fusion.
To tackle this issue, we propose an Offset-Guided Adaptive Alignment Module (OGAA), which performs feature-level alignment across modalities before fusion. 
OGAA employs two complementary strategies, \ie deformable convolution~\cite{def-conv} and spatial transformer networks~\cite{STN}, to dynamically adjust feature maps and reduce spatial discrepancies. 
Following alignment, a lightweight adaptive dynamic fusion module is introduced to balance the contributions of different modalities, further enhancing fusion quality and robustness. 
Together, these components form a cohesive fusion pipeline that significantly improves tracking accuracy for small and fast-moving UAV targets.

Existing benchmark datasets for visual UAV tracking and detection tasks, such as those summarised in Table~\ref{tab:uav_datasets}, focus on single-modal detection or single-object tracking~\cite{Anti-UAV,Anti-UAV410,Anti-UAV600}. 
Although a few datasets, including AntiUAV and Halmstad Drone, incorporate multiple modalities (\eg RGB and infrared), they remain confined to single-object tracking scenarios. 
As a result, there is a conspicuous absence of large-scale, publicly available datasets supporting multi-modal multi-object tracking for UAVs, hindering the development of robust solutions.
To bridge this gap, we introduce MM-UAV, the first large-scale Multi-Modal UAV Tracking dataset, designed specifically for multi-object tracking across three sensing modalities, \ie RGB, infrared, and event data. 
Comprising 1,321 sequences (1,200 for training and 121 for testing), with approximately 2.8 million frames per modality, MM-UAV offers unprecedented scale and diversity. 
Notably, we adopt a stricter trajectory annotation protocol, \ie UAVs that leave and re-enter the scene retain their original identity, avoiding identity fragmentation common in other datasets. 
The key characteristics of MM-UAV can be summarised as: (1) smaller target sizes, (2) larger data scale, (3) richer modality variety, and (4) more consistent identity annotations. 
To the best of our knowledge, it is the largest and most comprehensive dataset of its kind to date. 
A detailed description and statistical analysis of the dataset are provided in Sec.~\ref{dataset}.
In summary, this work presents the following key contributions:
\begin{itemize}
\item The first large-scale multi-modal multi-object tracking dataset of flying UAVs. It contains three visual modalities, 1321 video sequences, and over 2.8 million frames per modality, featuring a larger scale, smaller targets, extended modalities, and stricter trajectories.

\item The first multi-modal multi-UAV tracking framework, serving as the benchmark baseline for this dataset. At the detection end of the network, an offset-guided adaptive alignment module and an adaptive dynamic fusion module are introduced to achieve multi-modal feature alignment and fusion. At the tracking end, a new tracker (MMA-SORT) incorporating motion embedding is proposed, which effectively alleviates incorrect ID associations caused by false detections in complex backgrounds by utilising motion embedding.

\item The first attempt to introduce the Event modality into multi-modal multi-object tracking leverages its motion-aware capability to enhance the ability to distinguish targets that are extremely small in size, and move irregularly, such as UAVs.

\item Extensive experiments demonstrate that our proposed novel multi-modal multi-object framework outperforms current single-modal multi-object tracking SOTA methods, particularly in scenarios such as low illumination conditions and fast-moving contexts, providing a rigorous baseline and technical foundation for future research in multi-UAV tracking.
\end{itemize}

\section{RELATED WORK}

\subsection{UAV-Targeted Tracking Solutions}
Unlike UAV-view tracking~\cite{fan2020visdrone}, where the UAV serves as an aerial observer to perform air-to-ground tracking of objects such as pedestrians and vehicles within ground scenes, UAV-targeted tracking considers UAVs as objects of interest, aiming to achieve continuous and stable trajectory tracking of UAVs in complex low-altitude scenarios. 
Compared to general single-object tracking tasks, which rely on distinctive appearance cues and moderate scale variations, UAV targets are typically small, visually indistinct, fast-moving, and often concealed. 
As a result, in UAV-targeted scenarios, single-object tracking is required to maintain the trajectory of an initialised target across video frames as well as to overcome additional challenges such as weak appearance discriminability, frequent disappearance, and intermittent reappearance.
Most existing solutions are based on the Siamese network architecture, achieving target localisation through cross-frame template feature matching~\cite{xu2020accelerated}. 
Representative examples include early CNN-based solutions like SiamFC~\cite{siamfc}, SiamRPN~\cite{siamRPN}, and SiamRPN++~\cite{siamRPN++}, as well as Transformer-based trackers such as Stark~\cite{stark}, OSTrack~\cite{ostrack}, and MixFormer~\cite{mixformer}. 
However, these trackers are optimised primarily on large-scale general tracking datasets, such as GOT-10k~\cite{got10k}, TrackingNet~\cite{trackingnet}, and LaSOT~\cite{lasot}, lacking specific descriptions for UAVs.

Given the specific appearance and the long-term nature of the UAV tracking tasks, many anti-UAV SOT solutions employ a ``local tracking + global re-detection'' framework. 
For instance, SiamSTA~\cite{SiamSTA} leverages a spatio-temporal attention mechanism and a three-stage global re-detection strategy to track fast-moving UAVs. 
UTTracker~\cite{UTTracker} performs multi-region local tracking with temporal cues, combined with global detection and background correction modules, to handle drastic scale variations, frequent disappearance, and camera motion. 
Despite their achievements, these solutions are limited to single-object tracking and cannot handle complex scenarios involving collaborative intrusion by unknown numbers of UAVs. 
Thus, multi-UAV tracking has become a focus to meet practical demands.

In general, Multi-Object Tracking focuses on detecting and tracking multiple targets across frames. 
Building on this paradigm, multi-UAV tracking aims to associate trajectories of multiple UAVs without prior target initialisation, representing a crucial step toward practical anti-UAV technologies. 
Current mainstream MOT frameworks fall into two paradigms: tracking-by-detection (\eg DeepSORT~\cite{DeepSORT}, ByteTrack~\cite{bytetrack}, BoTSORT~\cite{botsort}, BoostTrack~\cite{boostTrack}) and tracking-by-query (\eg TransTrack~\cite{transtrack}, Trackformer~\cite{trackformer}, MOTR~\cite{motr}, MeMOTR~\cite{memotr}). 
However, UAV-specific challenges significantly degrade their performance. First, high appearance similarity reduces the effectiveness of traditional ReID-based embeddings.
Second, fast and irregular motion violates linear motion assumptions in Kalman filtering, leading to prediction errors and ID switches.
Third, the small UAV target size increases detection failures, further complicating data association.

These limitations highlight the need for UAV-oriented solutions in multi-object tracking paradigms. 
Drawing on this, recent studies have explored adapting generic MOT pipelines to small targets, particularly in thermal infrared videos. 
Dist-Tracker~\cite{Dist-Tracker} combines a scale-shape-quality detector based on YOLOv12~\cite{yolov12} with a fusion of L2-IoU tracker to handle low contrast, scale variations, and motion instability.
Strong Baseline~\cite{strong-baseline} achieves competitive results using the YOLOv12 and BoT-SORT framework with customised training/inference strategies. 
Nevertheless, the progress in multi-UAV tracking has been limited, with most approaches confined to single-modality inputs, resulting in insufficient robustness under challenging conditions.

\subsection{UAV-Targeted Tracking Benchmarks}
While significant progress has been made in both SOT and MOT paradigms, their development has been constrained by the availability of suitable datasets, motivating the design of UAV-specific benchmarks. 

For single-object tracking, Chen \etal\cite{USC-Drone} introduced USC Drone, a dataset with 60 sequences and 20K frames, accompanied by a Fast-RCNN + MDNet system where the residual image input improves adaptation to a fast UAV motion. 
Huang \etal\cite{Anti-UAV410} constructed Anti-UAV410, a large-scale thermal infrared benchmark with 410 sequences and 438K frames, and proposed SiamDT with dual-semantic features for tiny targets. 
Zhu \etal\cite{Anti-UAV600} released Anti-UAV600, comprising 600 thermal sequences and 723K frames without prior bounding-box initialisation, and designed a ``global detection + local tracking'' framework that avoids reliance on the initial frame information through evidential collaboration. 
Despite the progress, these benchmarks remain limited to SOT and cannot depict scenarios involving multiple UAVs.

For multi-UAV tracking, UAVSwarm~\cite{UAVSwarm} builds a visible-light dataset with 72 sequences (12K frames) and validates a Faster-RCNN/YOLOX + ByteTrack baseline. 
MOT-FLY~\cite{MOT-FLY} presents an air-to-air high-resolution dataset (1920×1080) with 16 sequences (11K frames) and benchmarked advanced MOT frameworks. 
Recently, the 4th Anti-UAV Challenge~\cite{4th-Anti-UAV-workshop} released the first large-scale infrared multi-UAV benchmark with 300 sequences and 225K frames.

Notably, existing multi-UAV tracking research remains limited: datasets are small-scale and merely provide single modalities (RGB or IR). 
In general, single-modality sensors inherently struggle in complex scenarios: visible-light performance degrades under low light or adverse weather, while infrared, though all-weather capable, suffers from low target-background contrast and missing details. 
These limitations severely restrict the robustness of multi-UAV tracking systems in complex low-altitude environments, motivating multi-modal fusion as a crucial enabling solution.

\begin{figure*}
    \centering
    \includegraphics[width=0.8\linewidth]{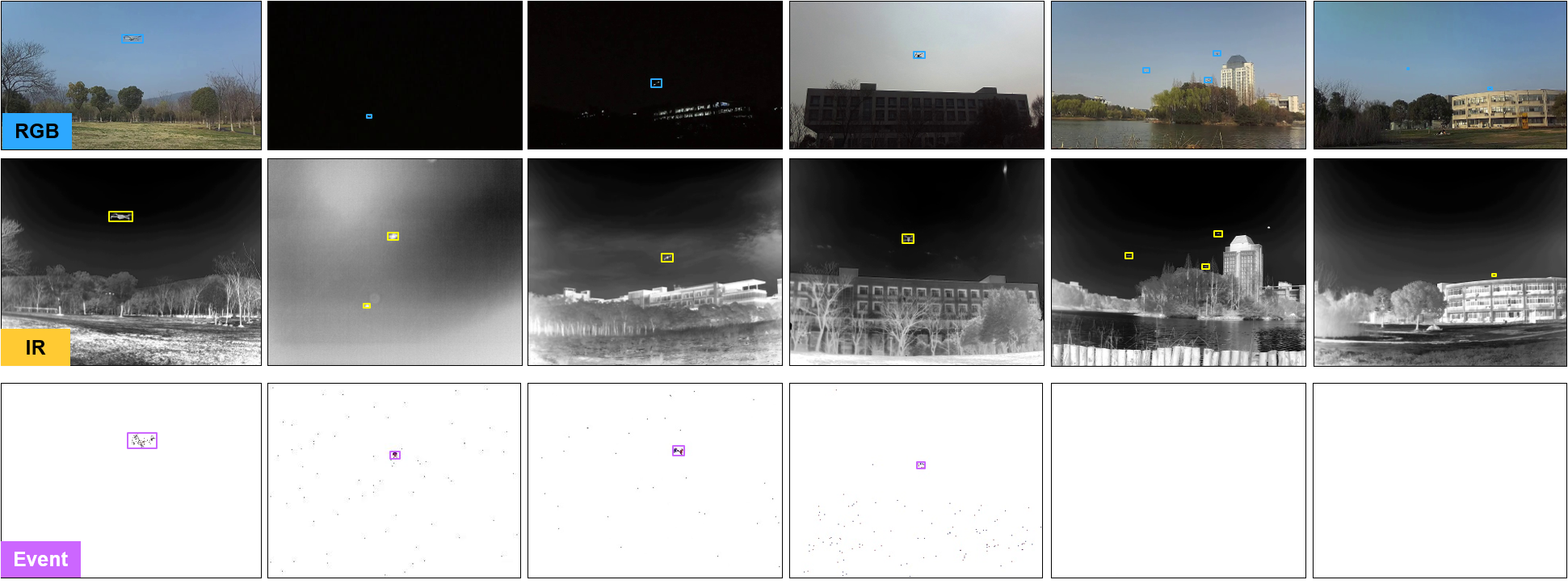}
    \caption{An example of the released MM-UAV dataset. The respective imaging characteristics of different sensors provide modal complementarity that is critical for robust operation in different scenarios. Different modalities exhibit varying degrees of visual misalignment. For instance, in low-light scenarios, the IR (infrared) modality can provide more discriminative clues.}
    \label{fig:dataset-sample}
\end{figure*}

\subsection{Multi-Modal Tracking Potential}
The performance bottleneck exhibited by a single visual modality in complex scenarios have driven the development of multi-modal fusion technologies~\cite{cheng2025one}, such as RGB-T (RGB + thermal infrared), RGB-D (RGB + depth), and RGB-E (RGB + event). 
Integrating the complementary information conveyed by different  modalities significantly enhances the tracking robustness.

However, multi-modal research in anti-UAV tasks remains scarce. 
The exceptions include the Anti-UAV dataset~\cite{Anti-UAV}, which was the first to introduce an RGB-T dataset, with 348 sequences and 297K frames. As a baseline, \cite{Anti-UAV} proposed a dual-stream feature semantic consistency training strategy. 
Yet, due to the modalities being unaligned, it cannot exploit the real potential of the cross-modal complementarity, instead using them independently. 
Similarly, the Halmstad Drone dataset~\cite{Halmstad-Drone} includes 650 videos (365 for IR and 285 for RGB) along with 90 audio clips, but focuses on building automated multi-sensor systems rather than investigating deep learning-based fusion. 
Its incomplete correspondence between IR and visible-light data also limits the utility of multi-modal anti-UAV research. 
Moreover, these datasets primarily target detection or SOT, leaving a void in multi-modal multi-UAV tracking benchmarks and methods.

A fundamental challenge in multi-modal tracking lies in modality misalignment. 
While most studies assume the data being spatio-temporally aligned, real-world data often suffers from misalignment due to differences in imaging principles, field-of-view, resolution, and transmission delays.
This problem is especially critical for UAV tracking, where the targets are small and fast-moving, and even slight offsets undermine effective fusion. 
Traditional pixel-level solutions rely on feature-point matching and affine/perspective transformations to achieve alignment, but these methods incur considerable preprocessing latency and are unsuitable for real-time multi-UAV tracking. 
Recent studies, therefore, emphasise learnable feature alignment.
\cite{ARCNN} proposes an aligned region CNN to align RGB and thermal regions using a Region Feature Alignment Module with RoI jitter.
\cite{CSOM} proposes cross-modality spatial offset modelling, together with an offset-guided deformable alignment and fusion, achieving robust UAV RGB-IR detection without strict pre-alignment. 
Inspired by these advances, we adopt two feature-level alignment strategies for UAV tracking: (1) learning sampling offsets via deformable convolution~\cite{def-conv}; and (2) predicting affine transformation matrices via a spatial transformer network~\cite{STN}. 
Both strategies are end-to-end trainable without additional losses, enabling efficient mitigation of misalignment in multi-modal fusion.

Additionally, given the small size and irregular motion of UAVs, as well as the inability of traditional RGB or IR modalities to distinguish targets from background under similar appearance or temperature, this study further introduces the Event modality. 
As a cutting-edge visual sensor, event cameras capture asynchronous event streams by recording changes in light intensity. 
They offer unique advantages such as high dynamic range and microsecond-level temporal resolution, making them particularly suited for extreme scenarios. 
Existing event-based tracking research has produced datasets such as VisEvent~\cite{visevent}, FE108~\cite{fe108}, EventVOT~\cite{eventvot}, and COESOT~\cite{ceutrack}, along with methods like CEUTrack~\cite{ceutrack}, TENet~\cite{TENet}, ViPT~\cite{vipt}, and AFNet~\cite{afnet}, but are restricted to the single-object tracking task, leaving a critical gap in event-based multi-UAV tracking.

\section{MM-UAV BENCHMARK}\label{dataset}

\begin{figure*}
    \centering
    \subfloat{\includegraphics[width=0.325\linewidth]{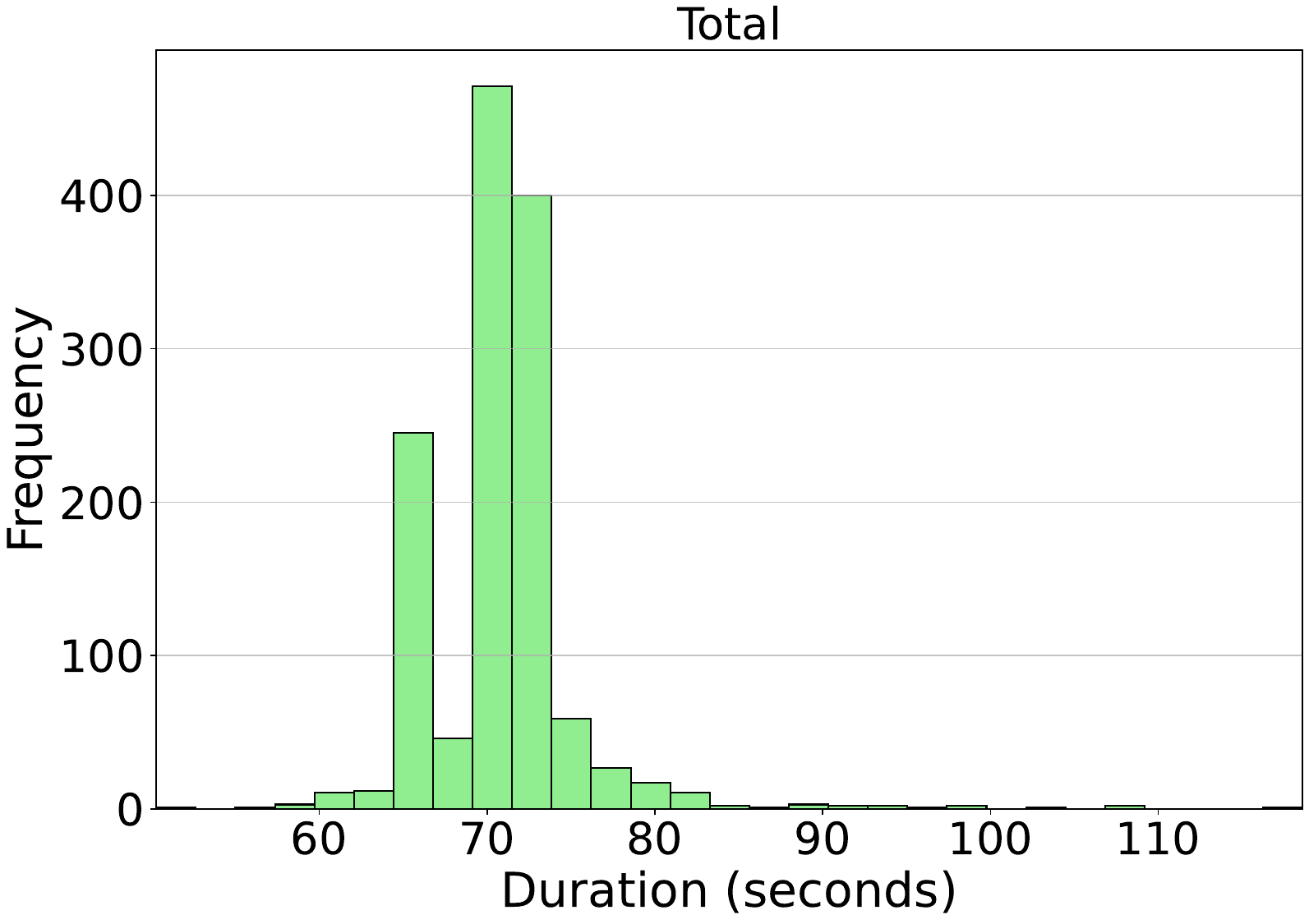}}
    \subfloat{\includegraphics[width=0.325\linewidth]{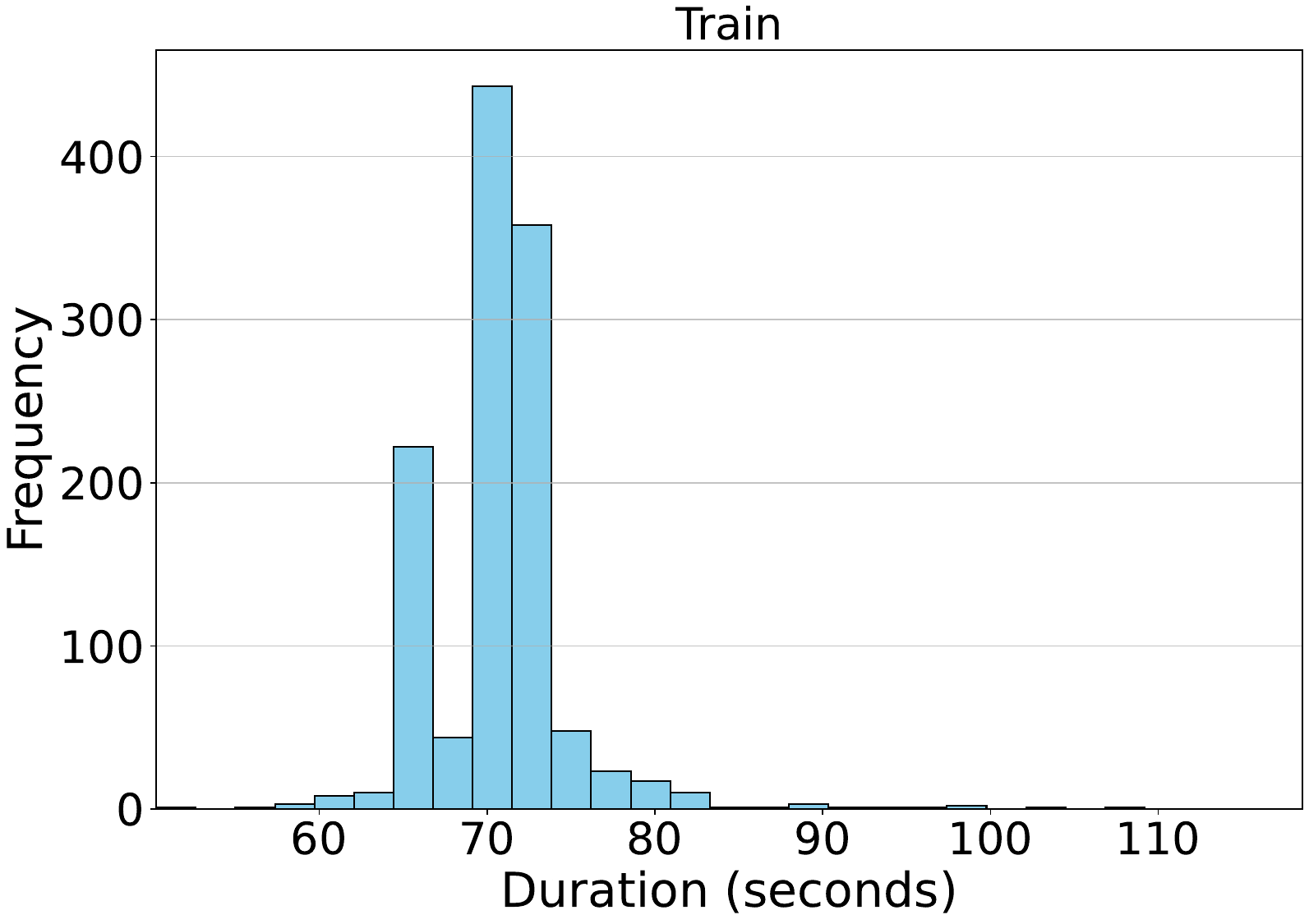}}
    \subfloat{\includegraphics[width=0.325\linewidth]{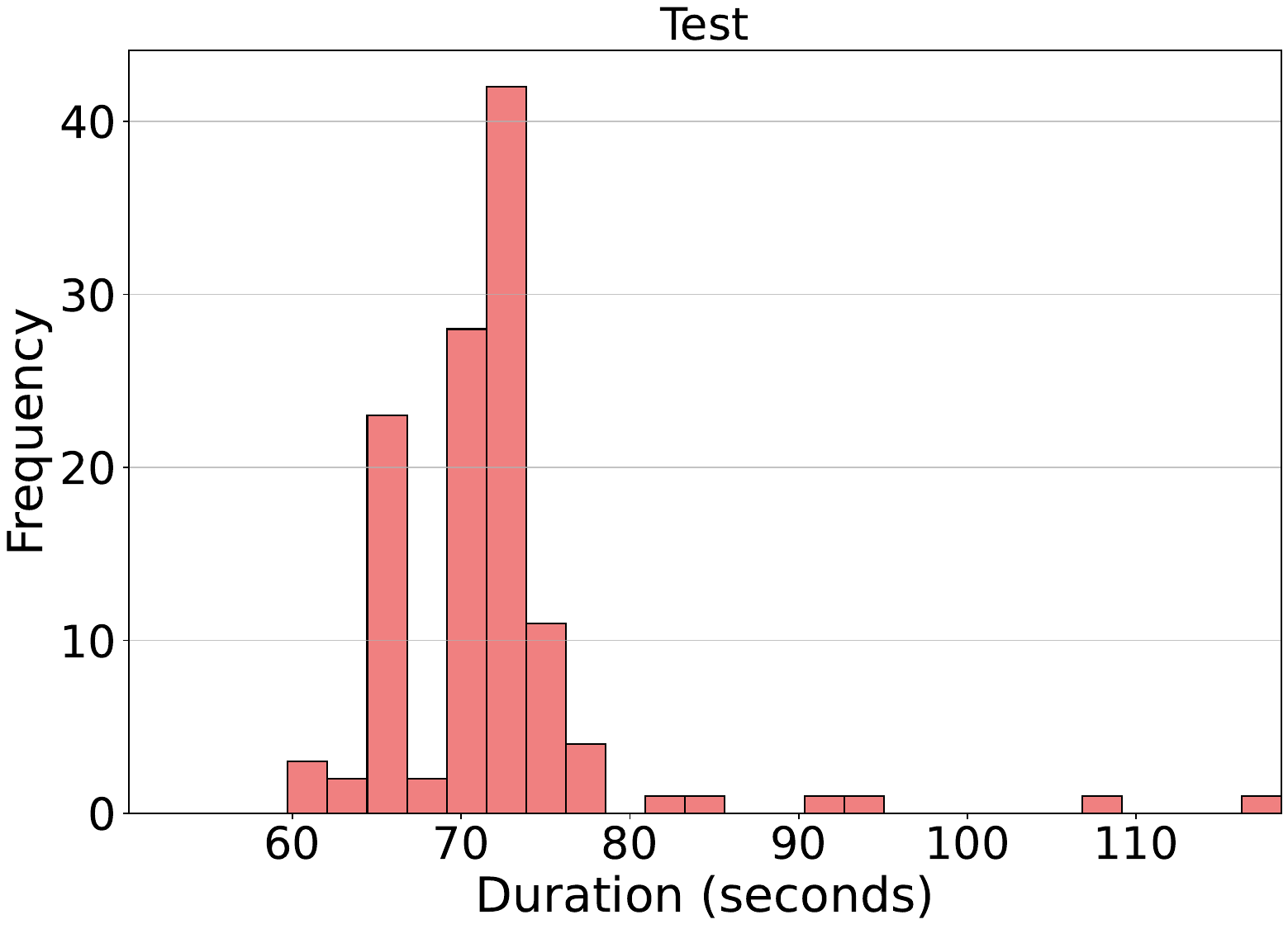}}
    \caption{Temporal distribution. We analyse the temporal distribution of sequences in the training set and in the test set separately. The figure details the number of sequences corresponding to each time interval. Notably, all modalities in this dataset share identical sequence lengths, so we do not distinguish between modalities.}
    \label{fig:dataset-time-distribution}
\end{figure*}
\begin{figure*}
    \centering
    \subfloat[Train]{
        \fbox{
            \begin{minipage}{0.46\linewidth}
                \includegraphics[width=0.48\linewidth]{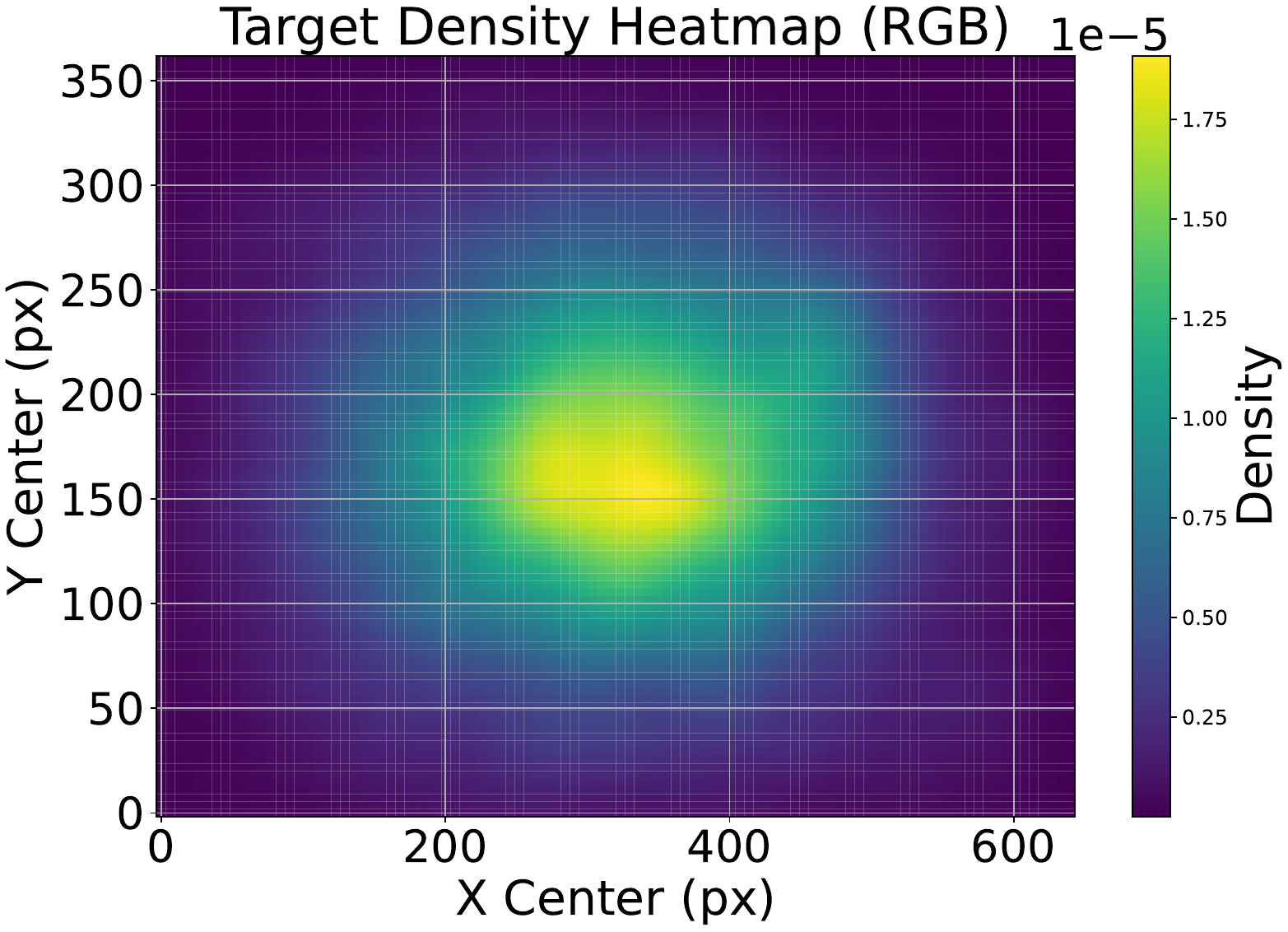}
                \includegraphics[width=0.48\linewidth]{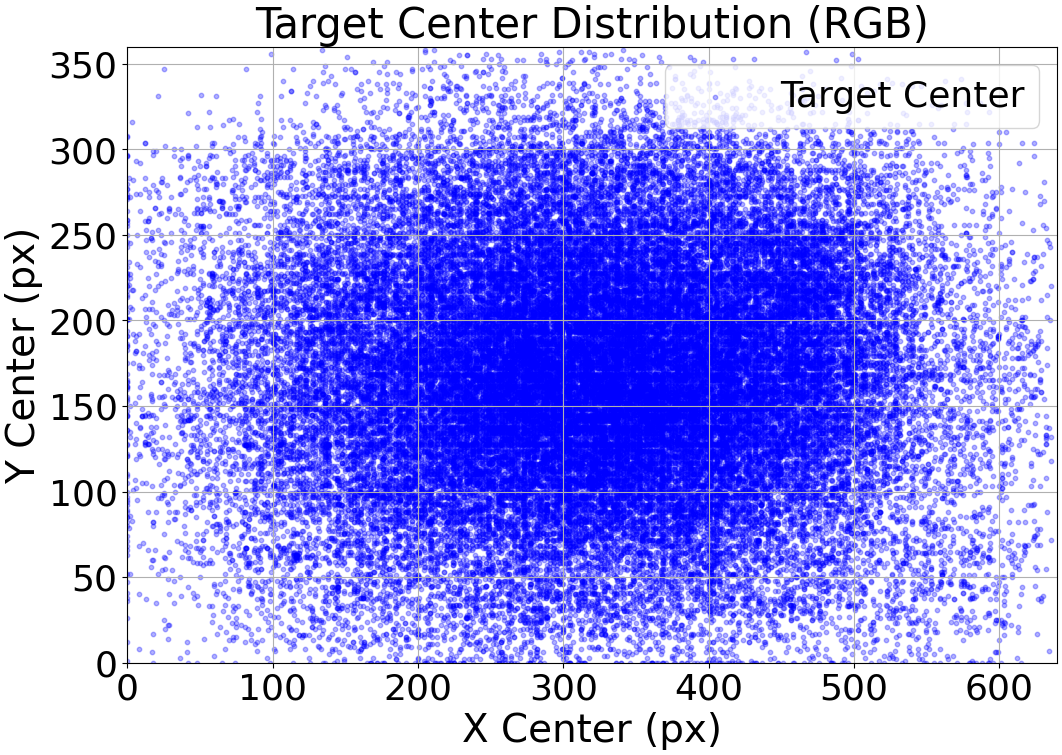}\\
                \includegraphics[width=0.48\linewidth]{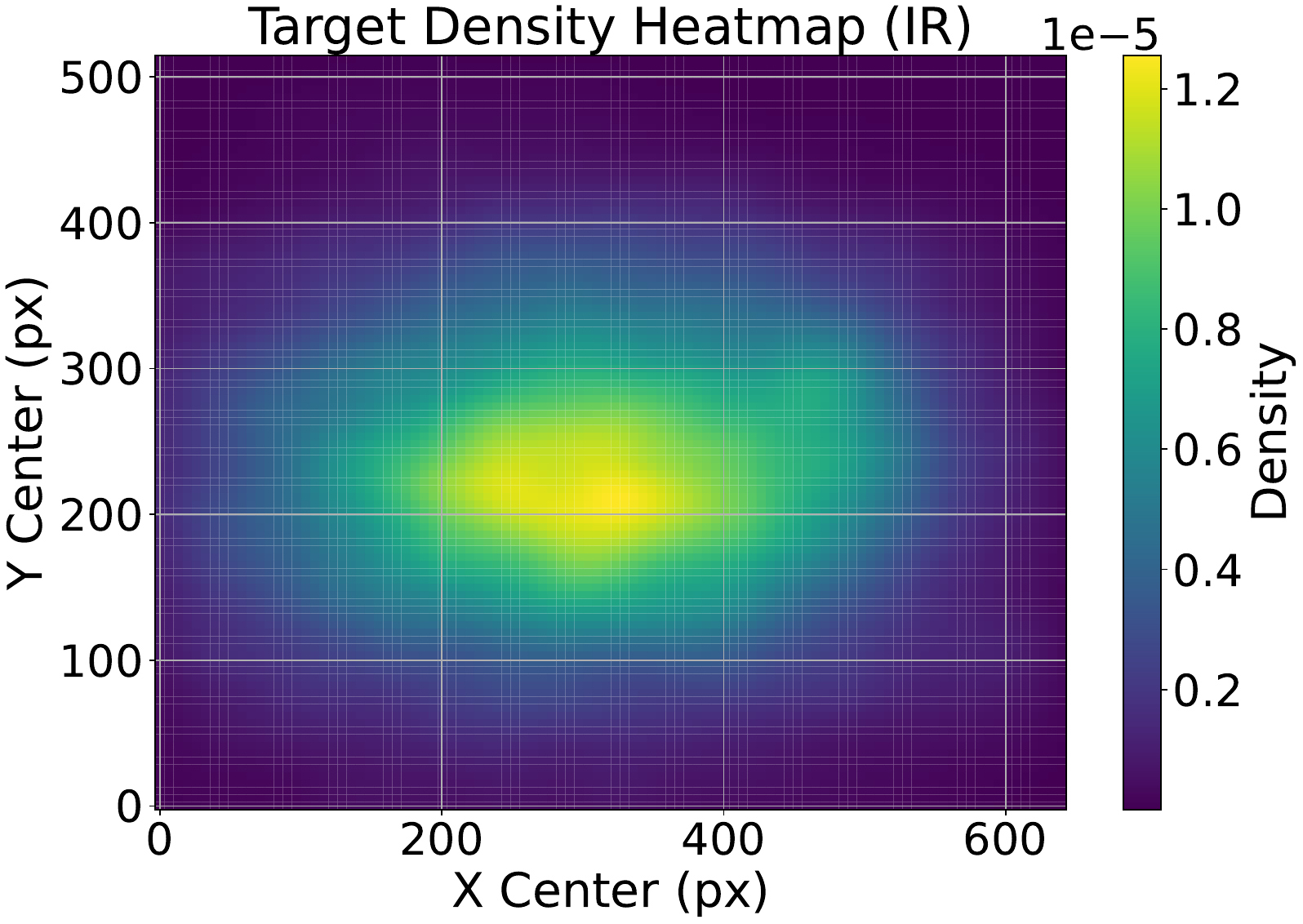}
                \includegraphics[width=0.48\linewidth]{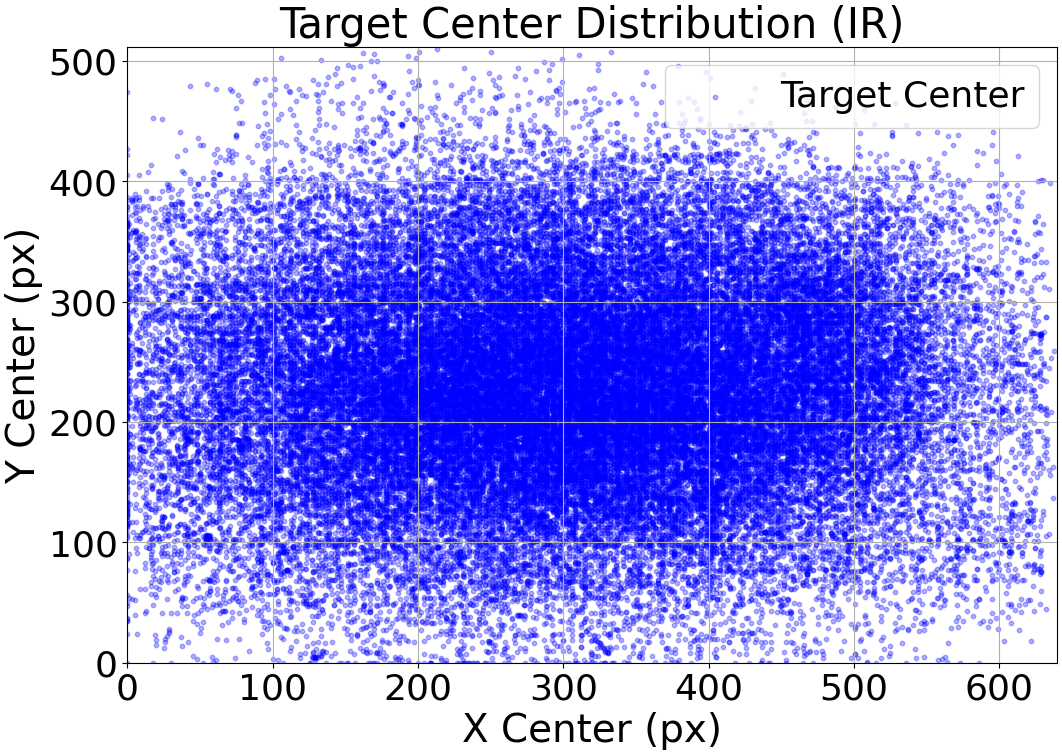}
            \end{minipage}
        }
    }
    \subfloat[Test]{
        \fbox{
            \begin{minipage}{0.46\linewidth}
            \includegraphics[width=0.48\linewidth]{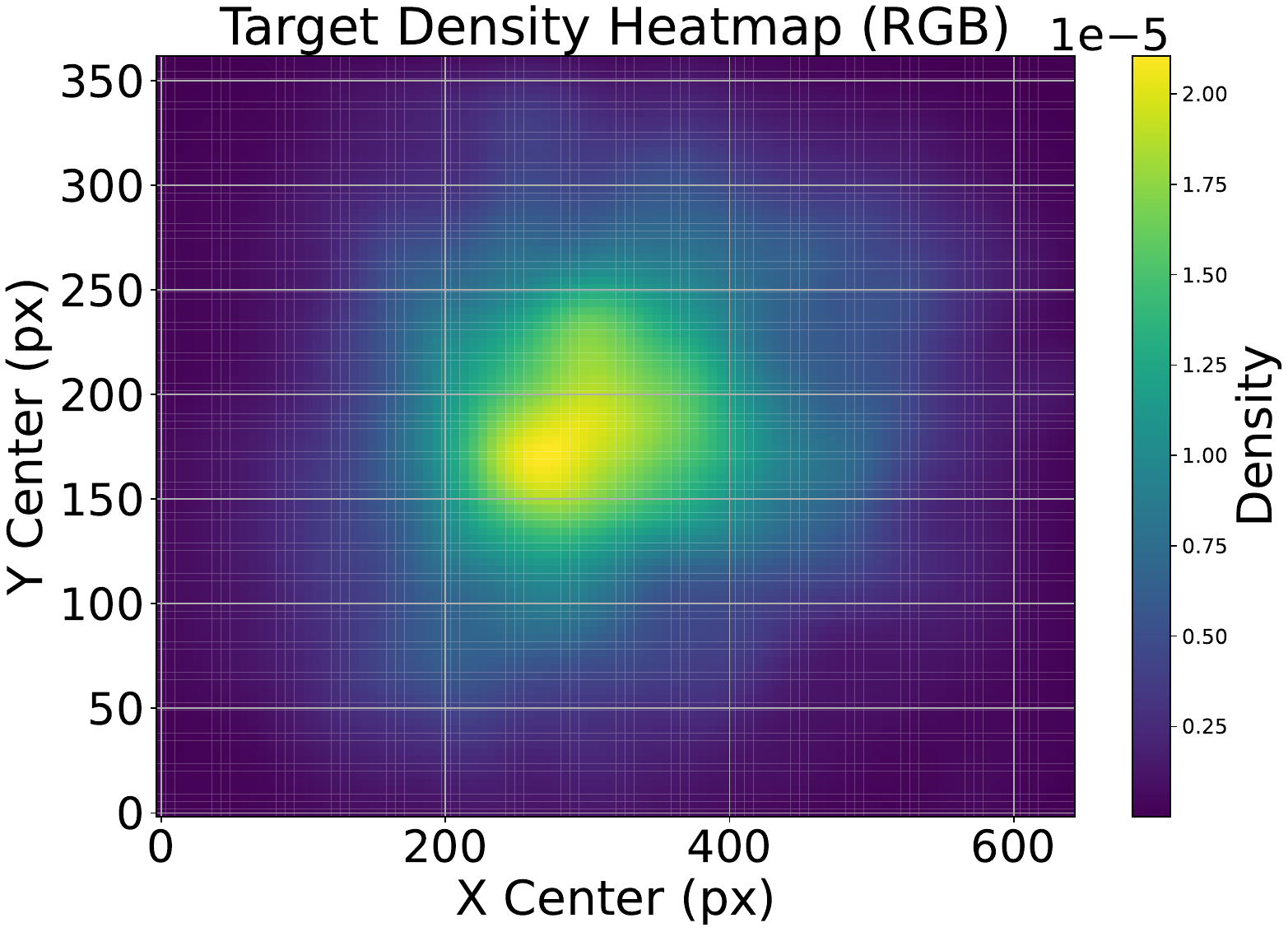}
            \includegraphics[width=0.48\linewidth]{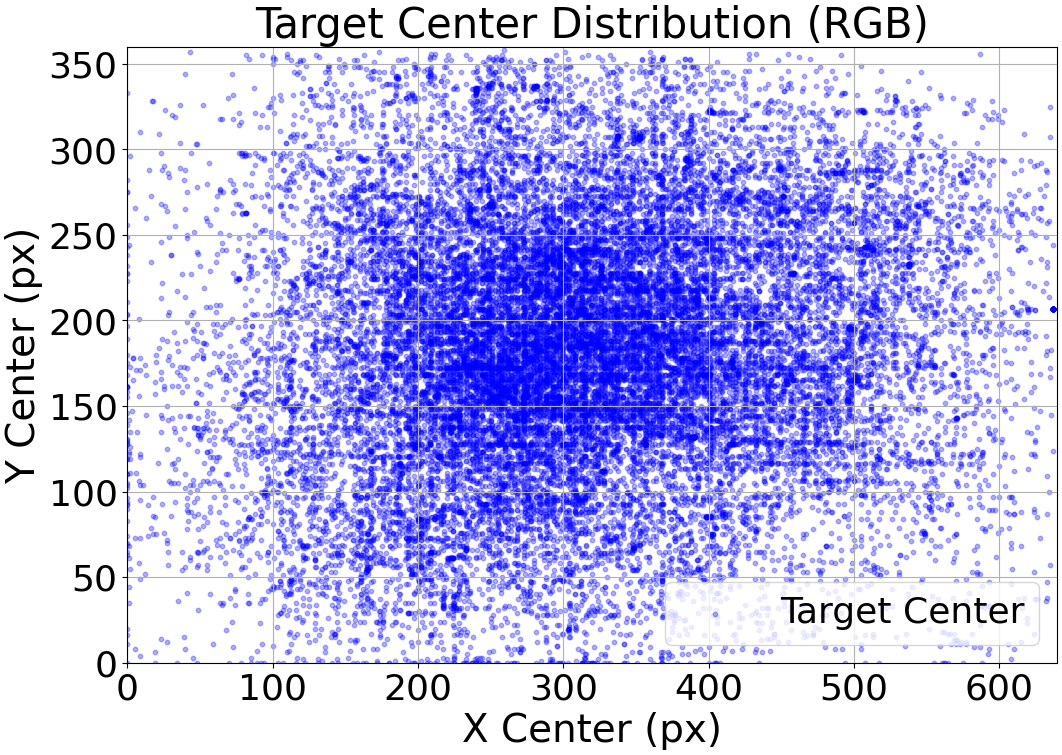}\\
            \includegraphics[width=0.48\linewidth]{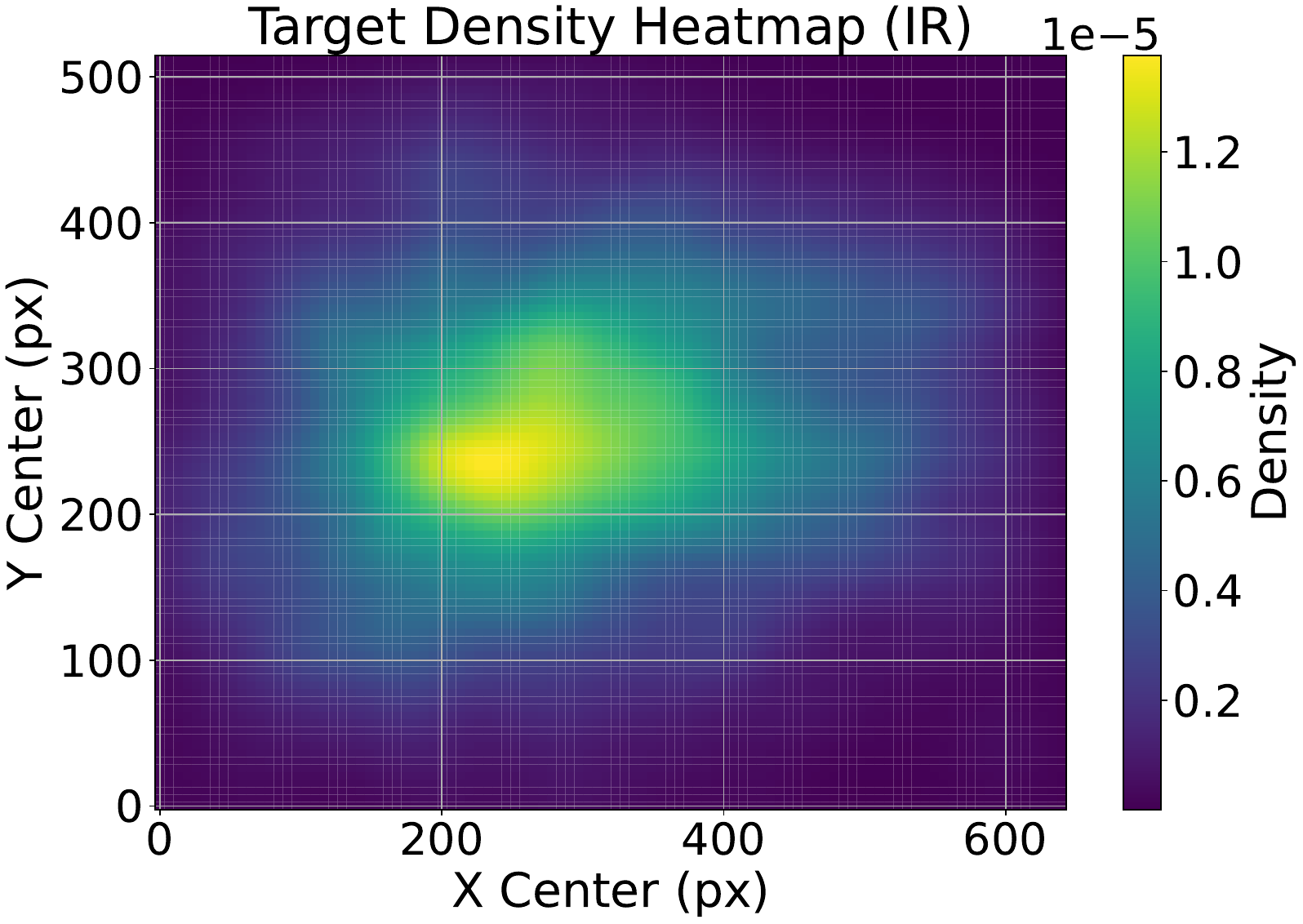}
            \includegraphics[width=0.48\linewidth]{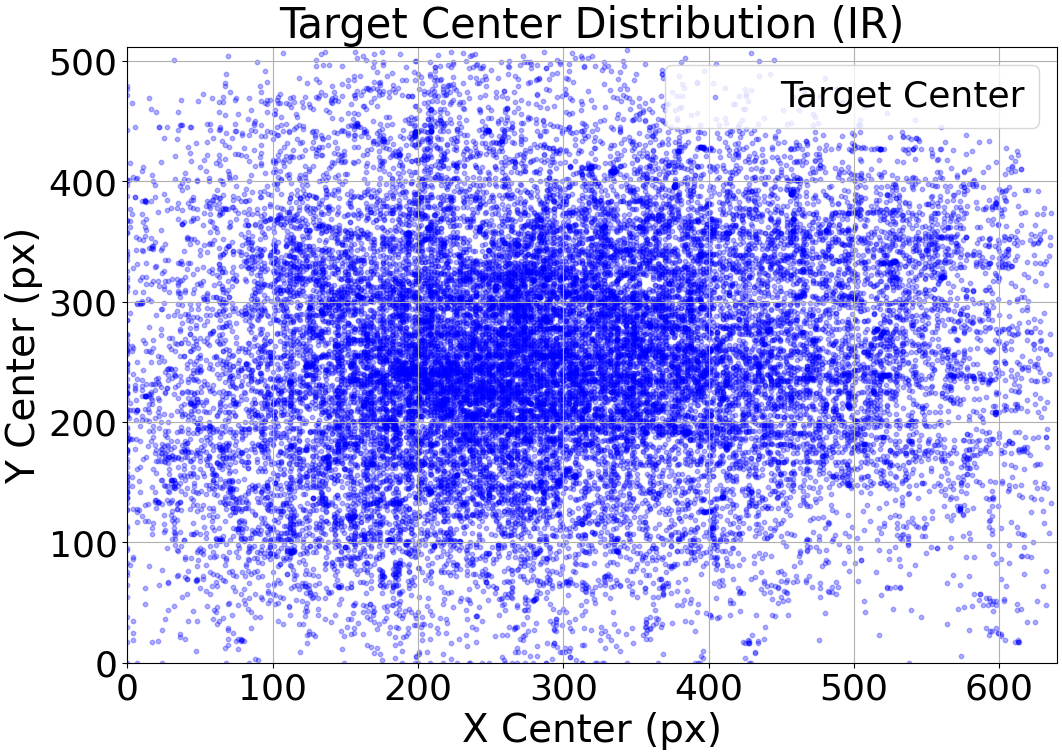}
            \end{minipage}
        }
    }    
    \caption{Spatial distribution. The spatial distributions of the targets in the training and test sets are examined and visualised using scatter plots and heatmaps. Since the two modalities may exhibit discrepancies in the target distributions due to differences in visibility and other factors, the spatial distributions are computed and presented separately for each modality.}
    \label{fig:dataset-spatial}
\end{figure*}
\begin{figure*}
    \centering
    \subfloat{%
        \begin{minipage}{0.32\linewidth}
            \centering
            \includegraphics[width=\linewidth]{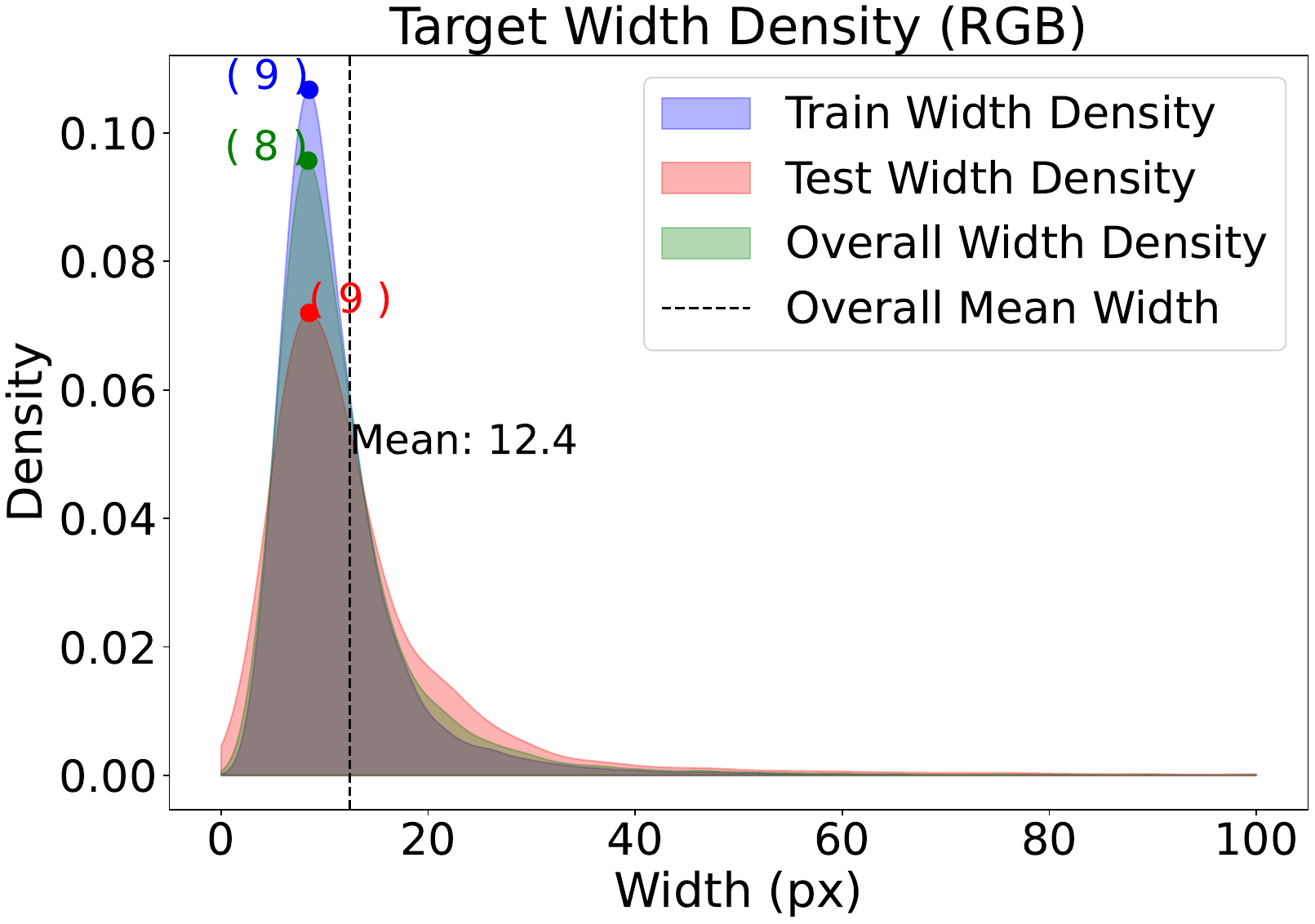}
        \end{minipage}
    }
    \hfill
    \subfloat{%
        \begin{minipage}{0.32\linewidth}
            \centering
            \includegraphics[width=\linewidth]{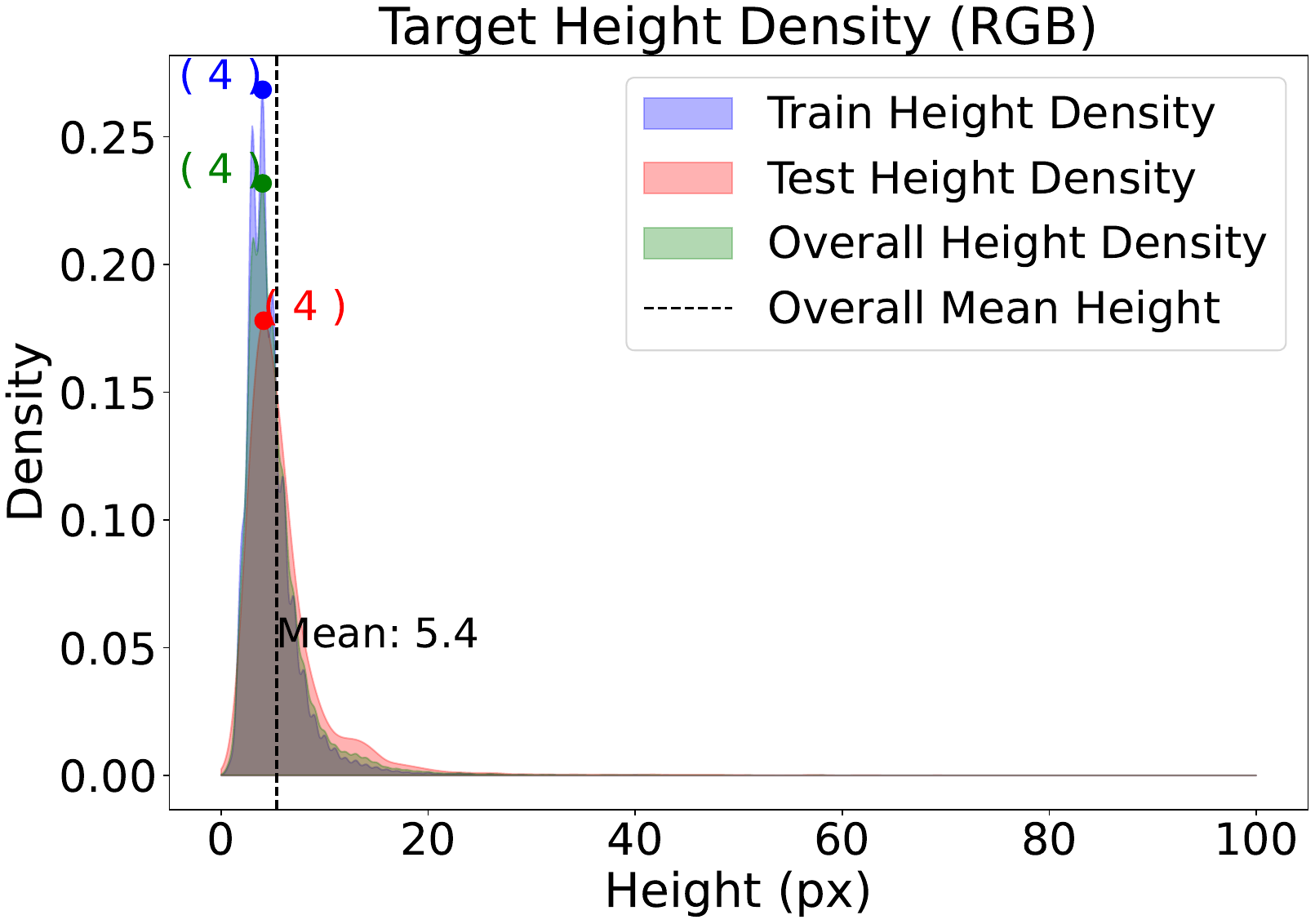}
        \end{minipage}
    }
    \hfill
    \subfloat{%
        \begin{minipage}{0.32\linewidth}
            \centering
            \includegraphics[width=\linewidth]{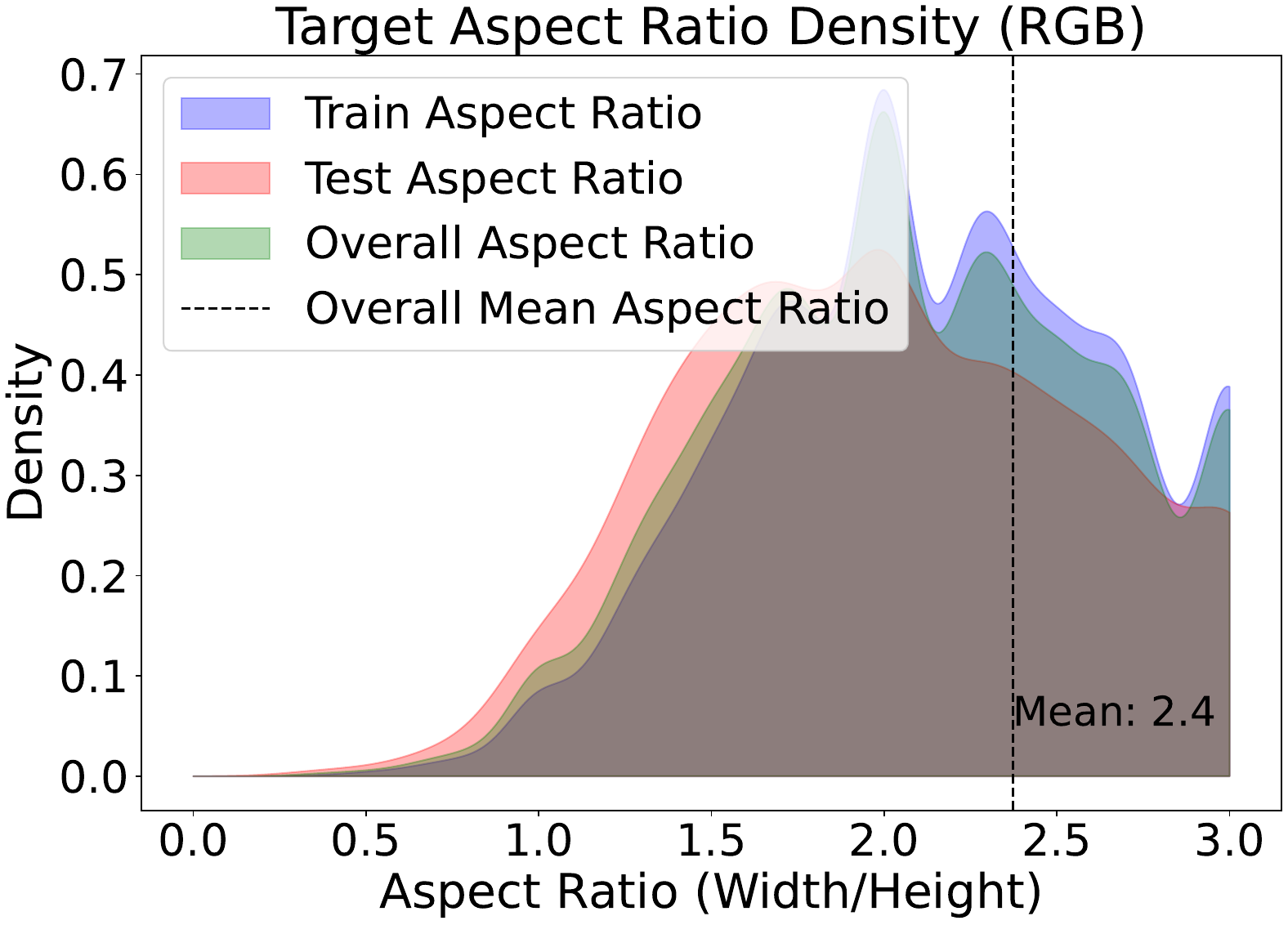}
        \end{minipage}
    }

    \subfloat{%
        \begin{minipage}{0.32\linewidth}
            \centering
            \includegraphics[width=\linewidth]{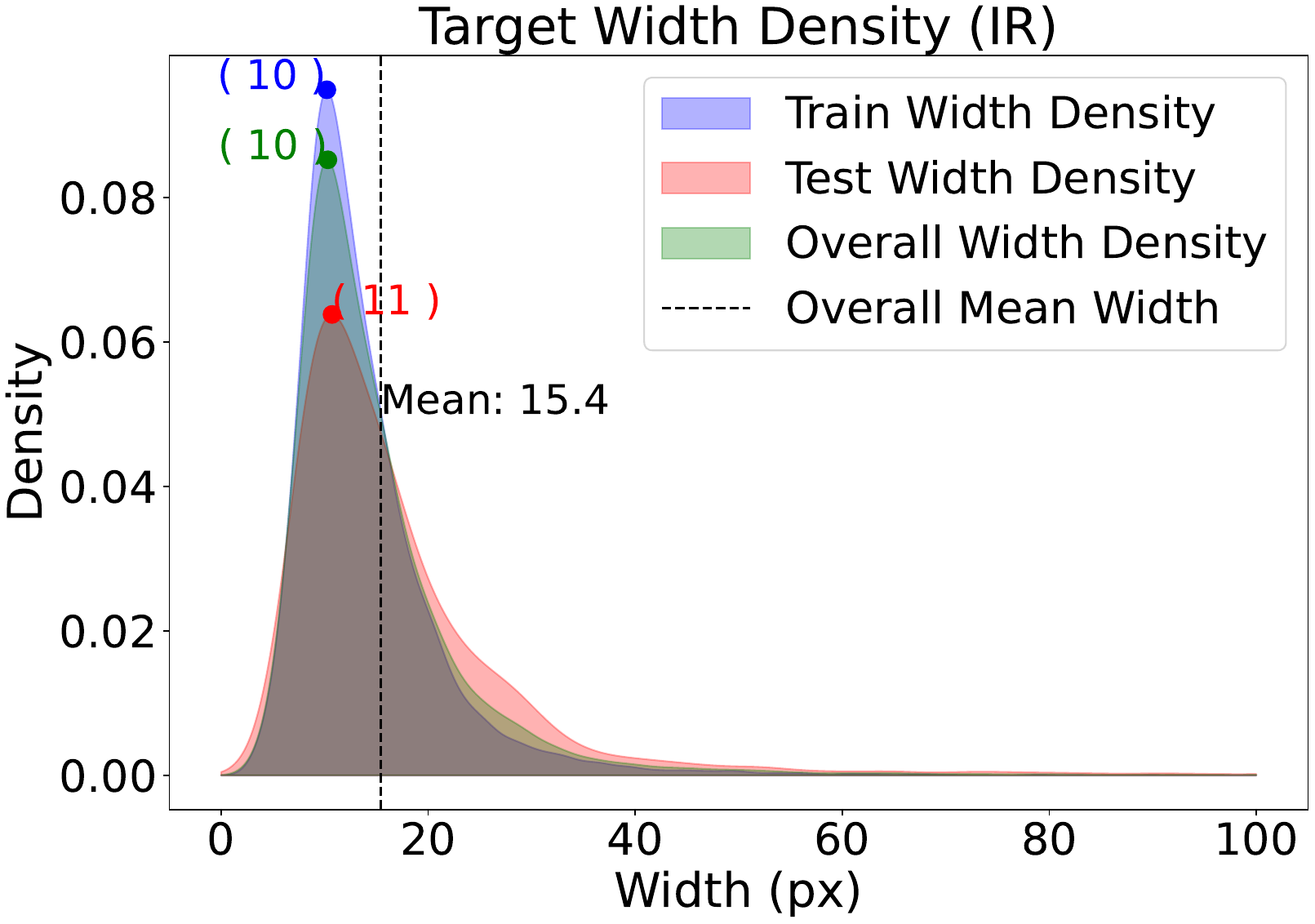}
        \end{minipage}
    }
    \hfill
    \subfloat{%
        \begin{minipage}{0.32\linewidth}
            \centering
            \includegraphics[width=\linewidth]{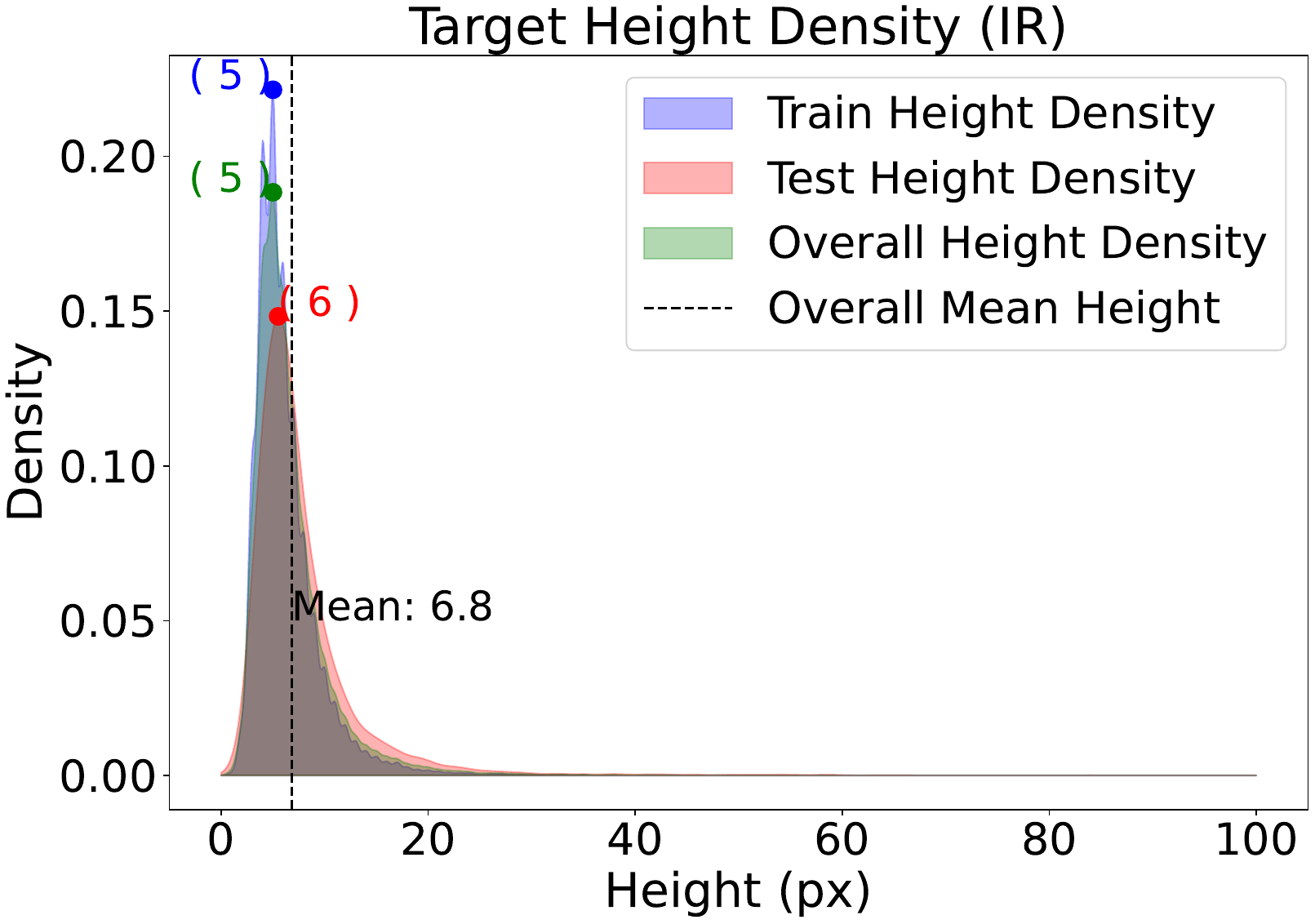}
        \end{minipage}
    }
    \hfill
    \subfloat{%
        \begin{minipage}{0.32\linewidth}
            \centering
            \includegraphics[width=\linewidth]{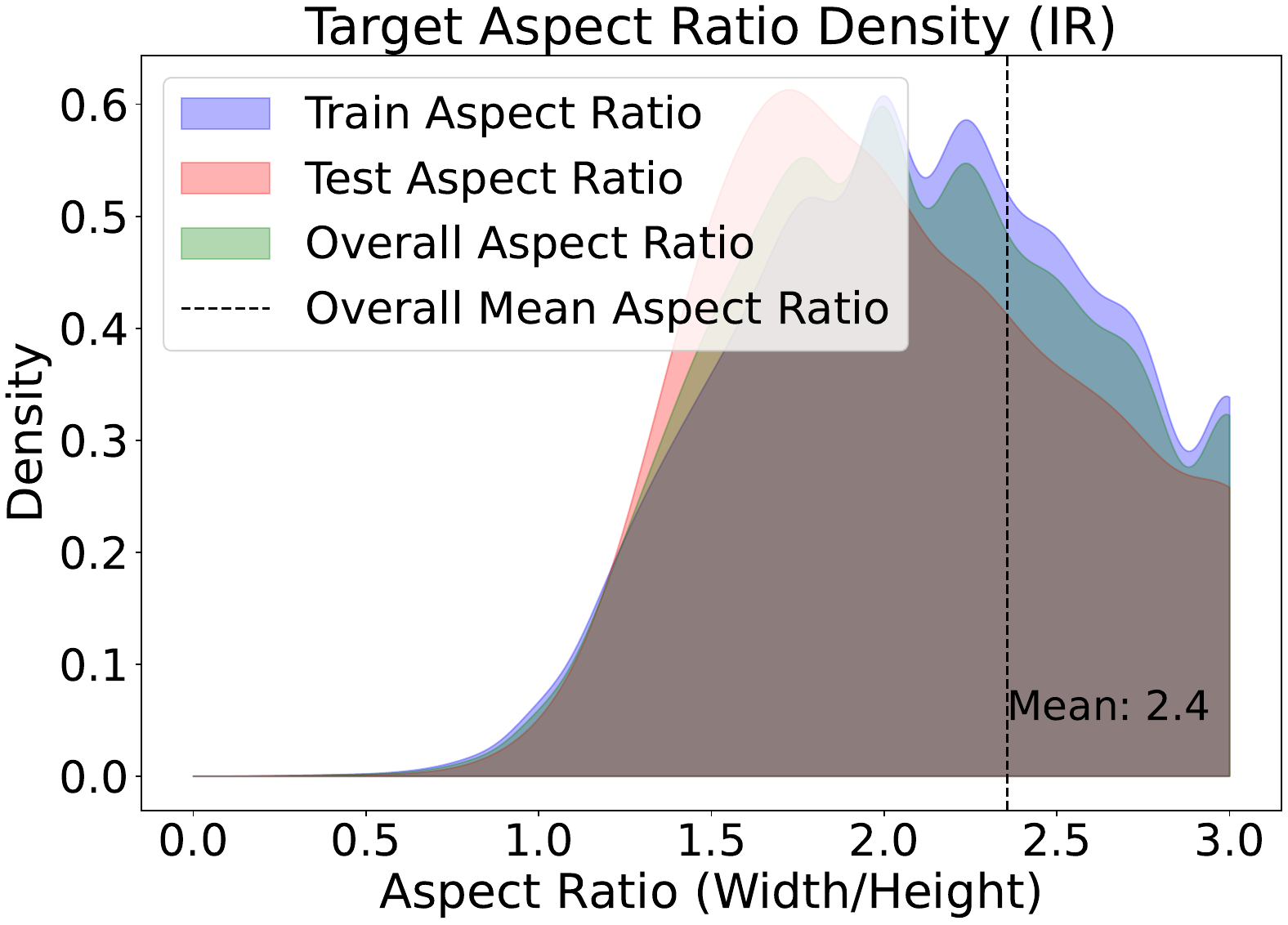}
        \end{minipage}
    }
    
    \caption{Size distribution. Although UAVs, as rigid bodies, do not undergo non-rigid deformation, the complexity of their flight postures leads to substantial variations in the aspect ratios of their bounding boxes.} 
    \label{fig:dataset-size}
\end{figure*}

The absence of a dedicated multi-modal multi-UAV tracking dataset to support anti-UAV research has significantly impeded the study of multi-modal fusion technologies. 
To address this, this paper presents the first large-scale tri-modal multi-UAV tracking dataset, with each modality containing 1321 sequences (2.8 million frames), aiming to fill this critical research void. 
This section gives a comprehensive overview of the dataset, covering core aspects such as data acquisition, annotation protocols, statistical distributions, and challenging attributes. 

\subsection{Data Collection}
In the MM-UAT dataset, the UAV targets are simultaneously captured using an infrared camera (infiRay LG6122), an RGB camera (Stereolabs ZED), and an event camera (DAVIS 346), with all modalities synchronised to generate video sequences at 30 frames per second (FPS). 
Specifically, the resolutions of RGB, IR, and event modalities are 640$\times$360, 640$\times$512, and 346$\times$260 pixels, respectively. 
The entire dataset comprises 1321 sequences across over 30 scenarios spanning multiple time periods, with 1200 sequences allocated to the training set and 121 to the test set. 
Sample sequences are provided in Fig.~\ref{fig:dataset-sample} to give an intuitive feeling for the collected data.

\subsection{Data Annotation}
Due to different imaging principles, field of view, and resolutions among sensor types, there are two critical challenging issues, \ie the size and visibility of the same target varies across modalities.
In particular, for small UAVs, this discrepancy is exacerbated, rendering shared annotations in terms of bounding box depiction across modalities extremely difficult.
For this reason, the annotation for the RGB and IR modalities in this dataset are provided independently.

Note that the event modality, designed to capture illumination intensity changes, exhibits characteristics such as information instability, high noise, and sparsity. 
These properties make it impractical to localise the targets directly with a precise size and position using the event modality alone. 
Accordingly, no annotations are provided for the event modality, which functions solely as an auxiliary input.

To facilitate efficient bounding box labelling while preserving the appearance diversity, a sparse annotation strategy is adopted.
Each training sequence is annotated every 100 frames, while the sequences from the 121 test set are annotated every 20 frames. 
Specifically, each annotation includes a unique ID and precise spatial coordinates of the target. 
Importantly, in our annotation setting, a strict ID persistence rule is enforced.
Within a sequence, a UAV retains its unique ID regardless of how many times it exits and re-enters the field of view.

\subsection{Statistics}
This section elaborates  various statistical characteristics of our dataset to facilitate an in-depth understanding of its properties.

\noindent \textbf{Temporal Distribution}.
As shown in Fig.~\ref{fig:dataset-time-distribution}, the total duration of the 1321 sequences is 26 hours, with the training set accounting for 23.6 hours and the test set for 2.4 hours. 
In particular, individual sequence durations range from 50 to 118 seconds, with an average of 70.96 seconds.

\noindent \textbf{Spatial Distribution}.
The spatial distribution of the targets in the training and test sets is visualised in Fig.~\ref{fig:dataset-spatial}. 
The UAV targets are widely distributed across the entire field of view, with the density increasing toward the centre region.

\noindent \textbf{Size Distribution}.
In principle, the UAV targets are characterised by their small size and high probability of concealment. 
In our MM-UAV, as shown in Fig.~\ref{fig:dataset-size}, the average size of UAVs is 12.4$\times$5.4 pixels in the RGB modality, and 15.4$\times$6.8 pixels in the IR modality, with both modalities exhibiting an average aspect ratio of 2.4. 
The aspect ratio distribution across modalities indicates that UAVs, due to their fast and erratic movement, undergo frequent and drastic posture changes.

\begin{figure*}
    \centering
    \subfloat{\includegraphics[width=0.325\linewidth]{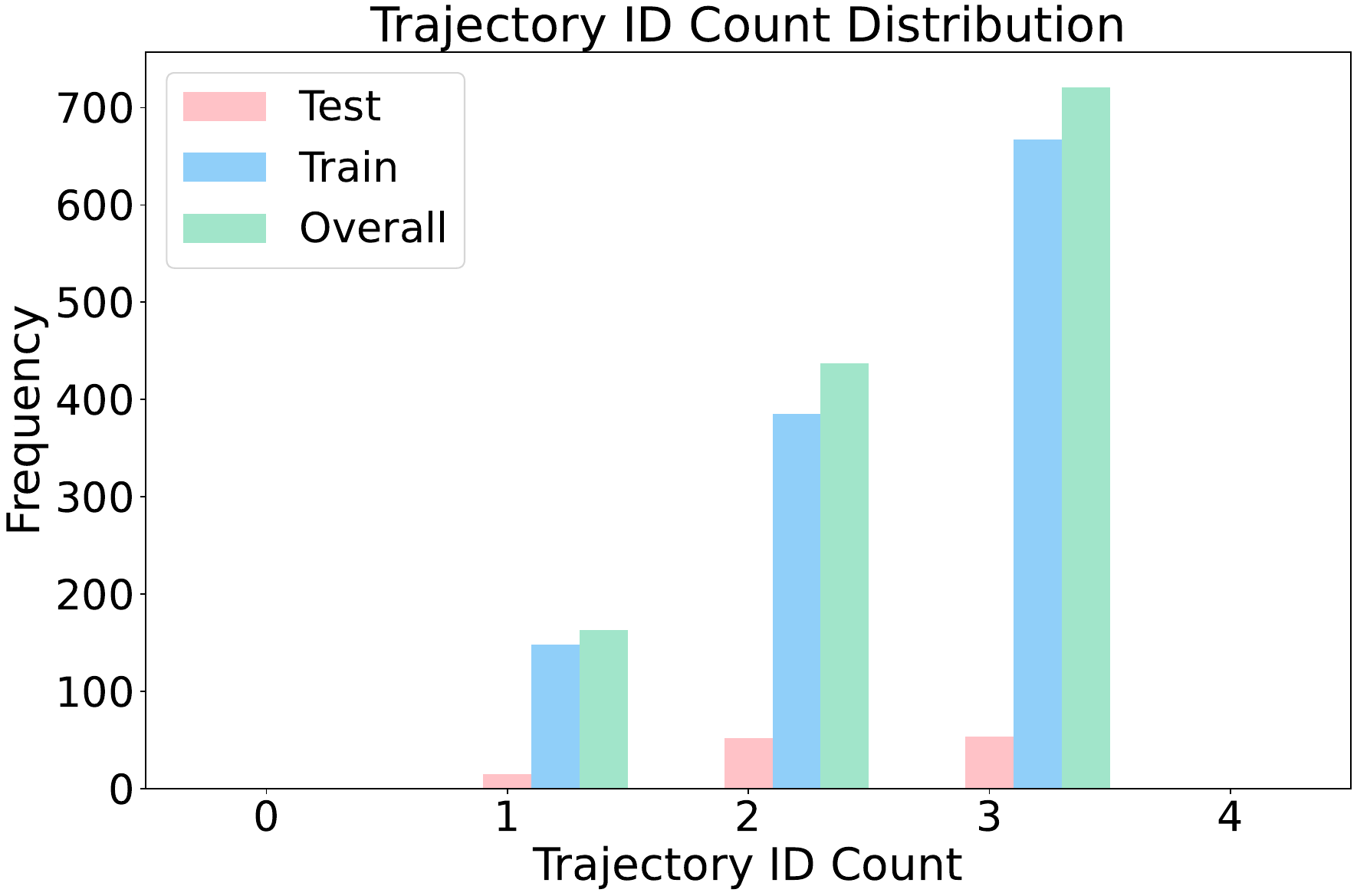}}
    \subfloat{\includegraphics[width=0.325\linewidth]{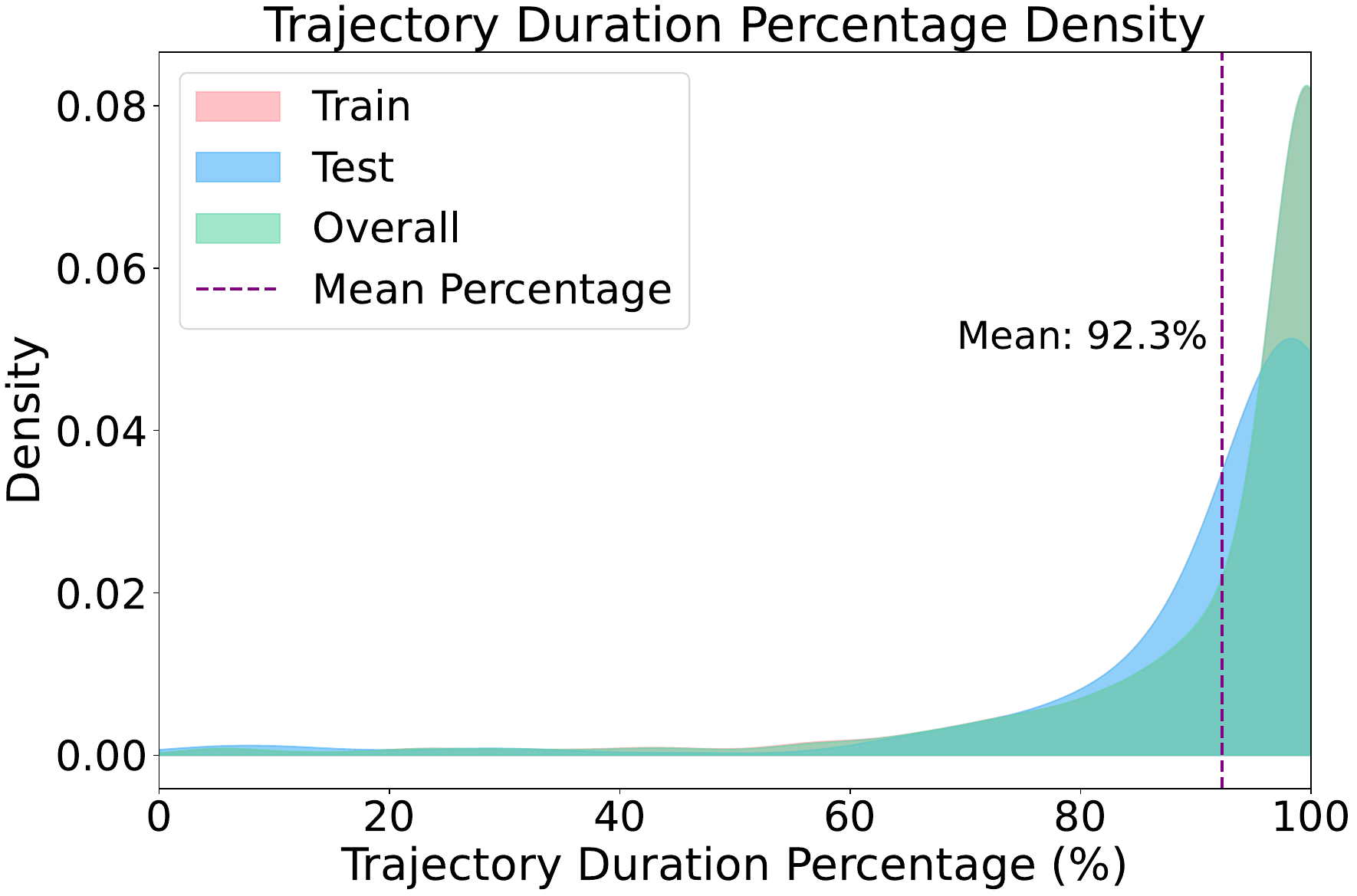}}
    \subfloat{\includegraphics[width=0.325\linewidth]{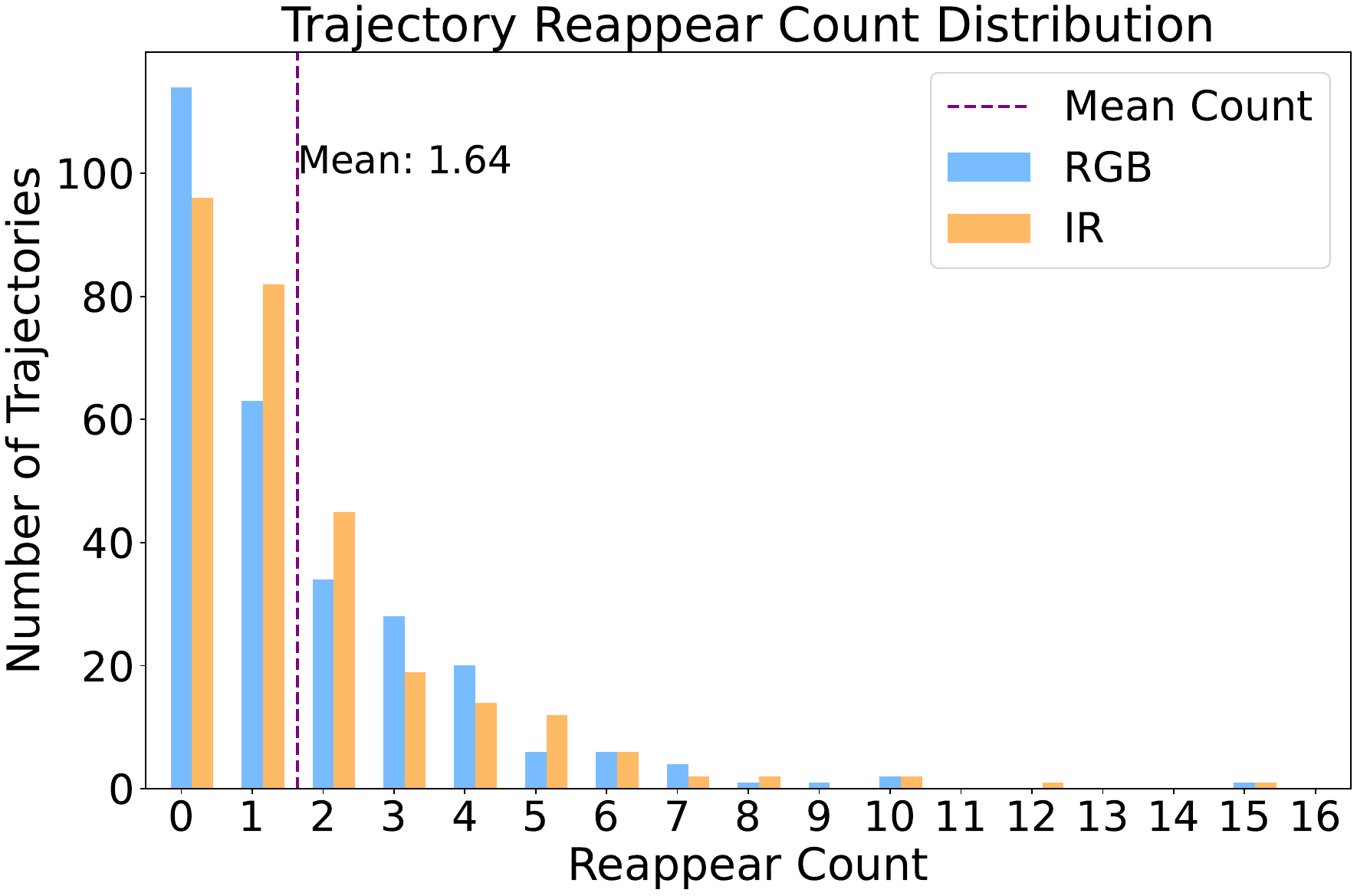}}
    \caption{The trajectory statistics. We provide the trajectory-level statistics across the entire dataset, including the number of target trajectories per sequence, the proportion of visible duration for each trajectory, and the frequency with which trajectories disappear and reappear.}
    \label{fig:dataset-tracjetory-distribution}
\end{figure*}
\begin{figure}
    \centering
    \includegraphics[width=0.95\linewidth]{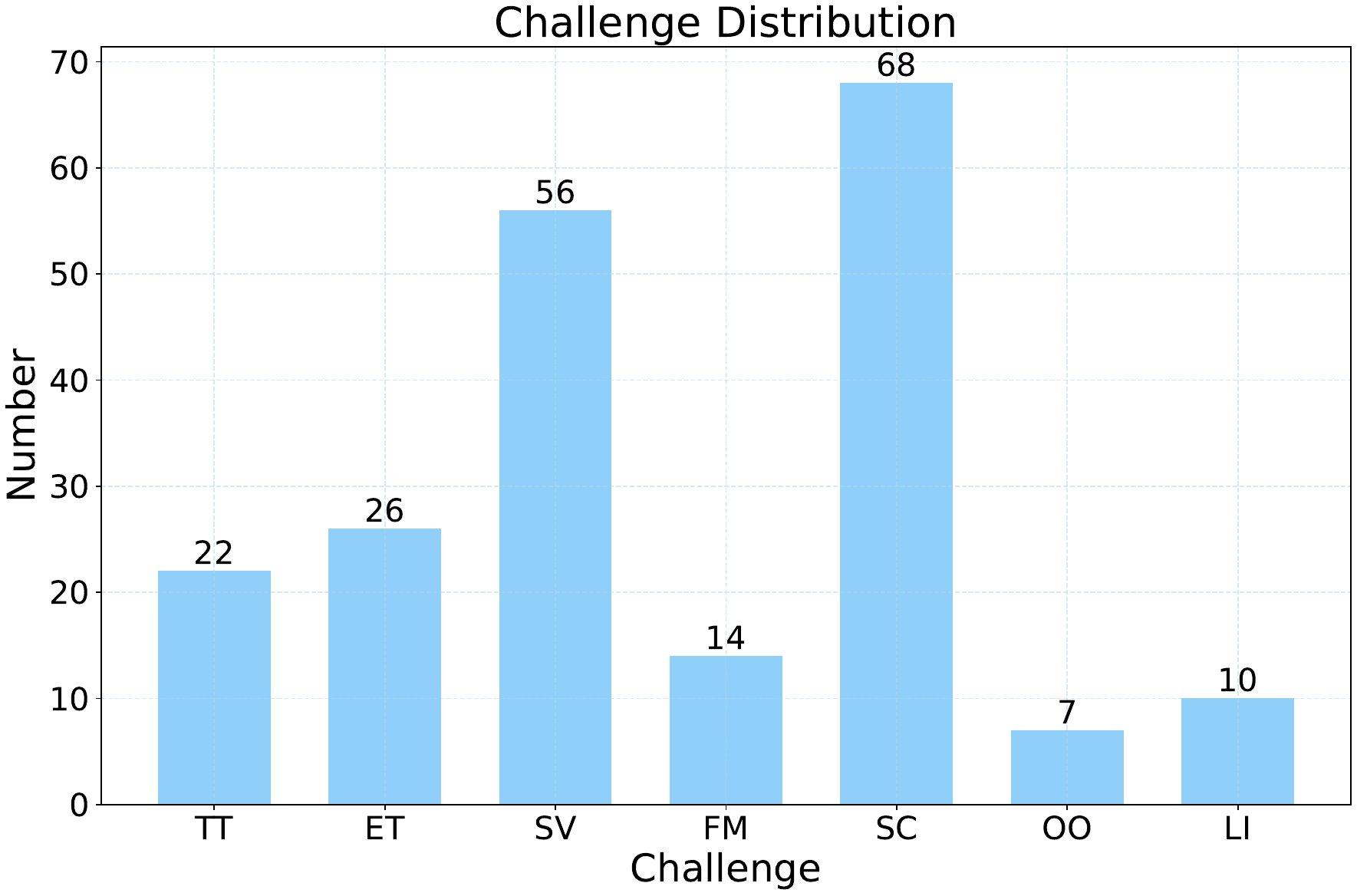}
    \caption{The distribution of challenging attributes in the Test set.}
    \label{fig:dataset-challenge}
\end{figure}

\noindent \textbf{Trajectory Distribution}.
In contrast to existing datasets that primarily focus on high-density UAV clusters, this dataset emphasises the stealth characteristics of UAVs, aiming to support robust multi-object tracking under sparse distribution and unknown quantities. 
In particular, each sequence comprises 1 to 4 UAV targets, with an average of 2.42 trajectories per sequence, as reported in  Fig.~\ref{fig:dataset-tracjetory-distribution}. 
The entire trajectories exhibit an average visibility duration of 92.32\%, with a minimum of 2.78\% and a maximum of 100\%. 
On average, each trajectory undergoes 1.55 exit-reentry events, reaching up to 15 instances of target exit and re-entry in the most extreme cases. 
These characteristics underscore the necessity for the tracking algorithms to have the capability to handle frequent occlusion and reappearance scenarios.

\subsection{Challenging Attributes}
To evaluate and analyse tracking performance across scenarios comprehensively, the collected sequences are annotated with seven challenging attributes:
\begin{figure*}[t]
    \centering
    {\setlength{\fboxrule}{0.8pt}   
     \setlength{\fboxsep}{2pt}      
    \begin{minipage}{0.39\linewidth}
        \centering
        \subfloat[TT]{\fbox{\includegraphics[width=\linewidth]{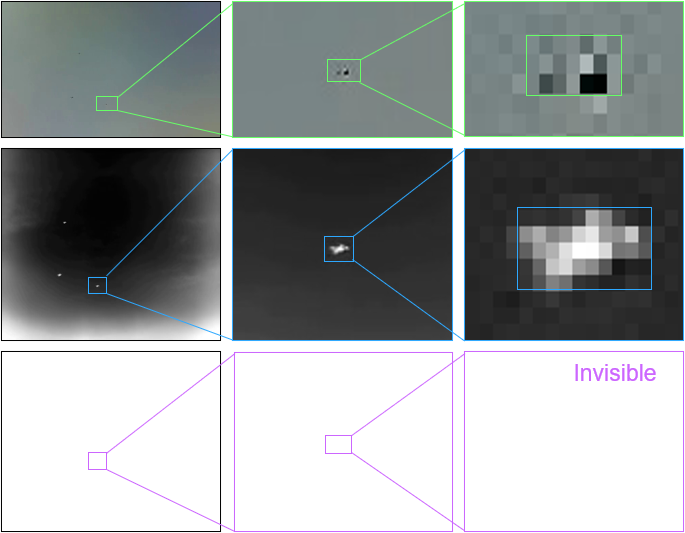}}}
    \end{minipage}\hfill
    \begin{minipage}{0.57\linewidth}
        \centering
        \subfloat[SV]{\fbox{\includegraphics[width=\linewidth]{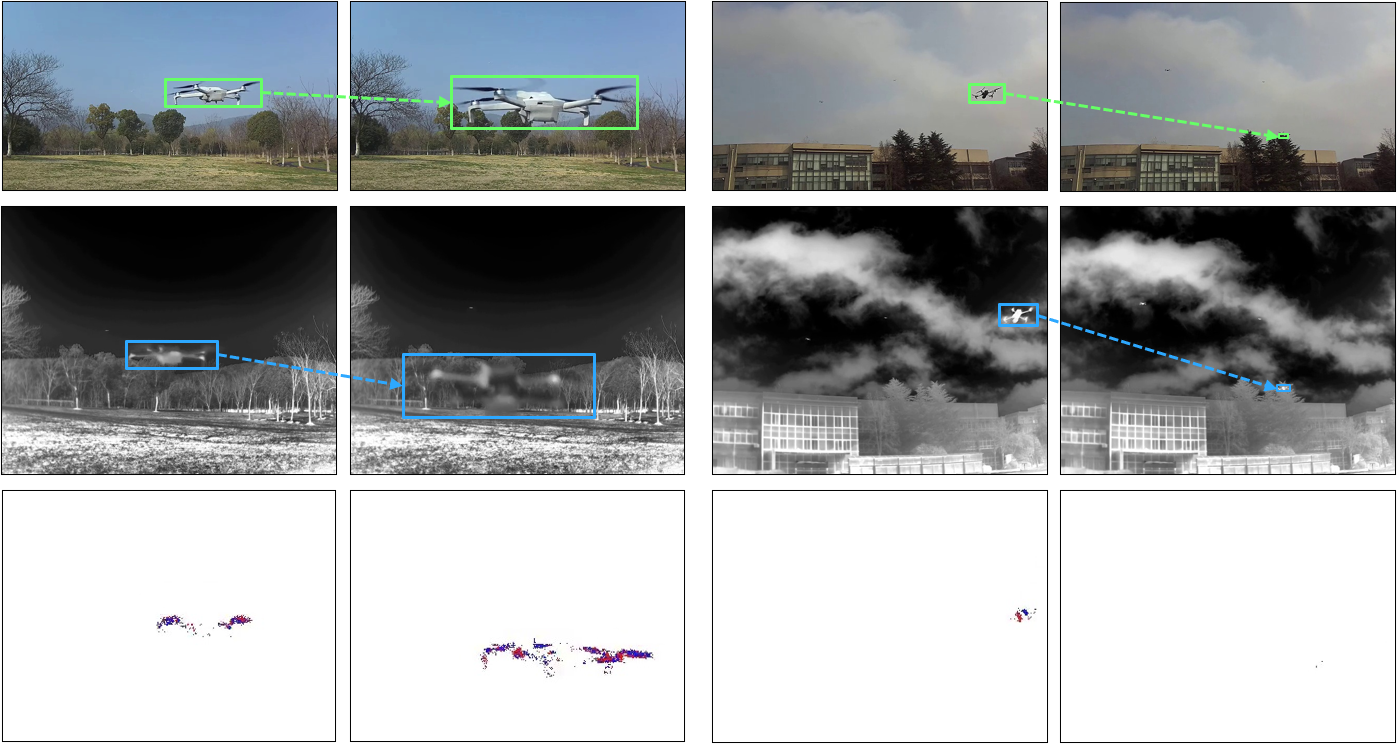}}}
    \end{minipage}

    \begin{minipage}{0.27\linewidth}
        \centering
        \subfloat[SC]{\fbox{\includegraphics[width=\linewidth]{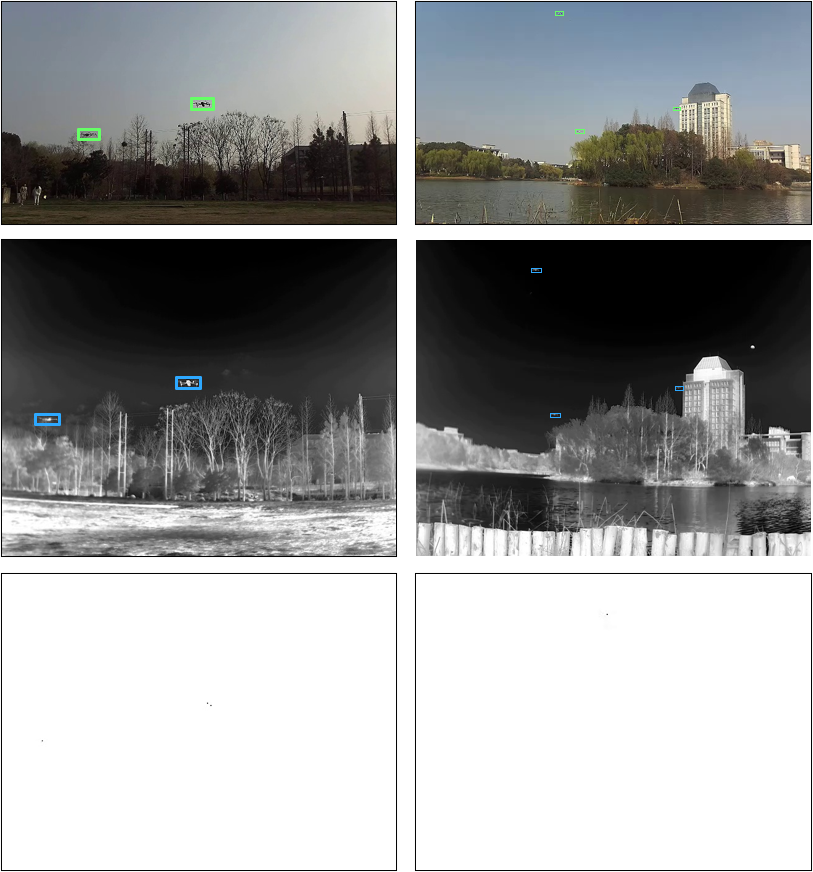}}}
    \end{minipage}\hfill
    \begin{minipage}{0.37\linewidth}
        \centering
        \subfloat[ET]{\fbox{\includegraphics[width=\linewidth]{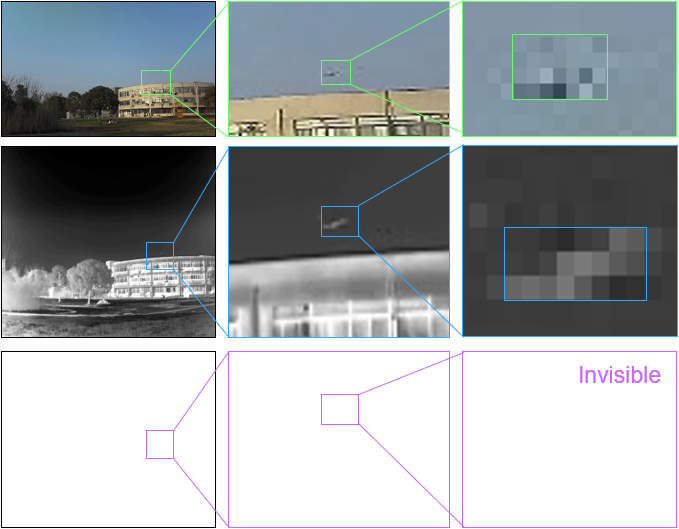}}}
    \end{minipage}\hfill
    \begin{minipage}{0.27\linewidth}
        \centering
        \subfloat[OO]{\fbox{\includegraphics[width=\linewidth]{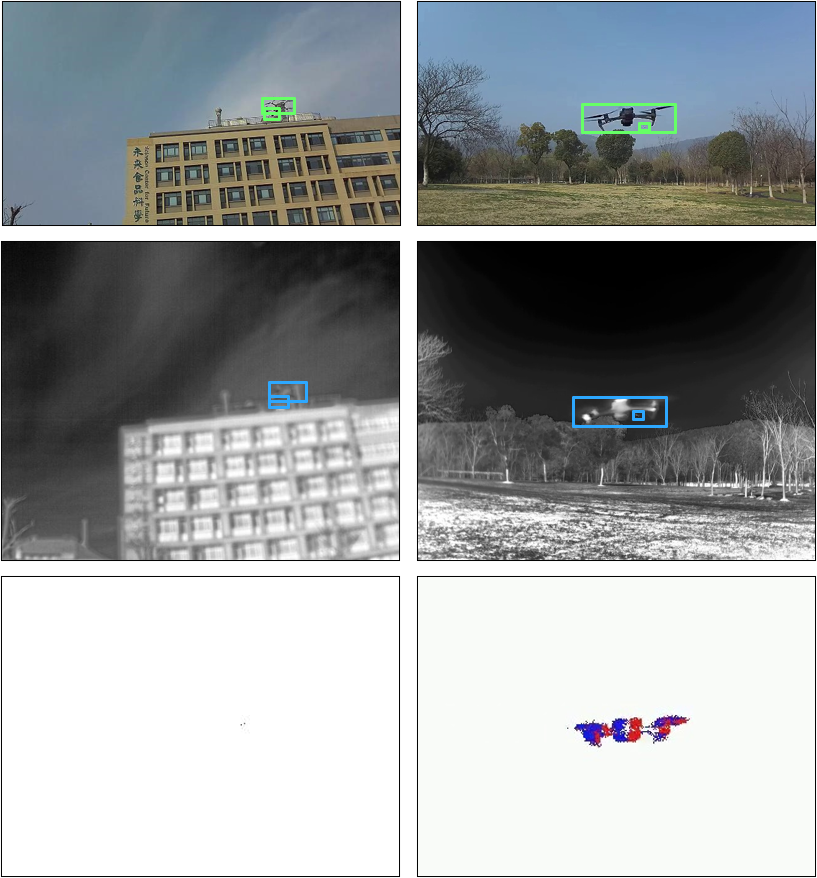}}}
    \end{minipage}

    \begin{minipage}{0.55\linewidth}
        \centering
        \subfloat[FM]{\fbox{\includegraphics[width=\linewidth]{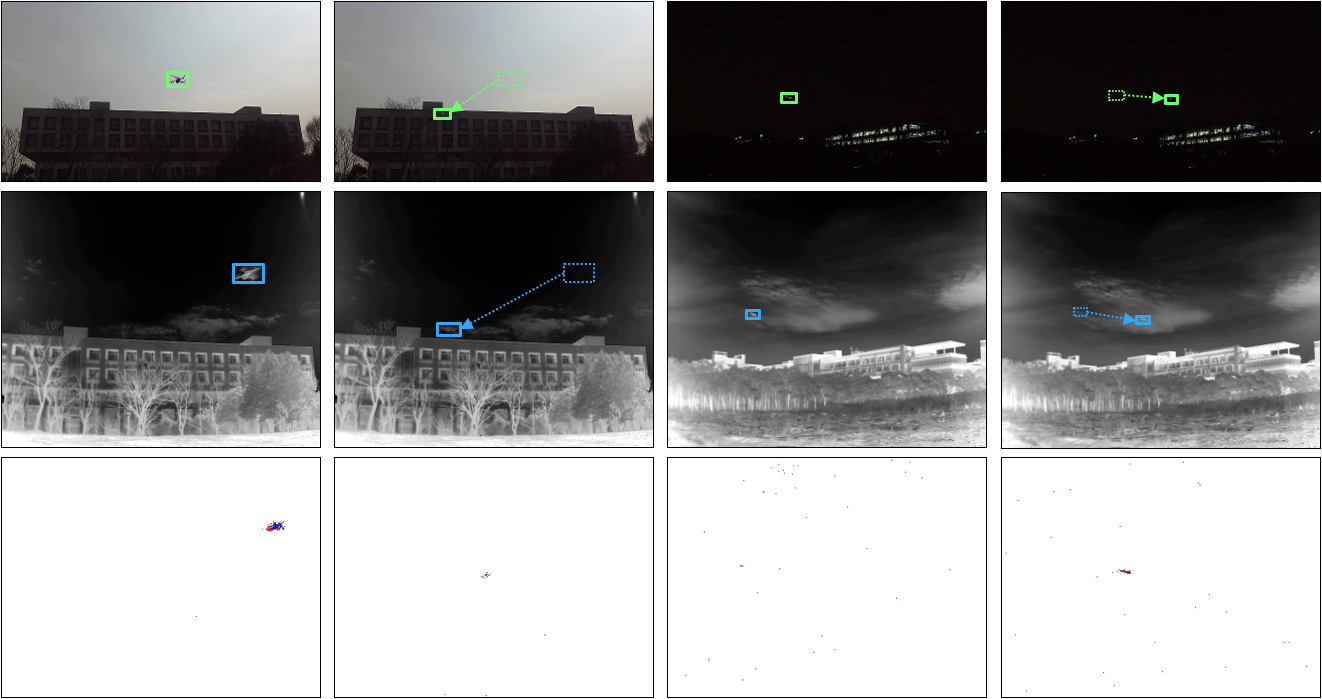}}}
    \end{minipage}\hfill
    \begin{minipage}{0.42\linewidth}
        \centering
        \subfloat[LI]{\fbox{\includegraphics[width=\linewidth]{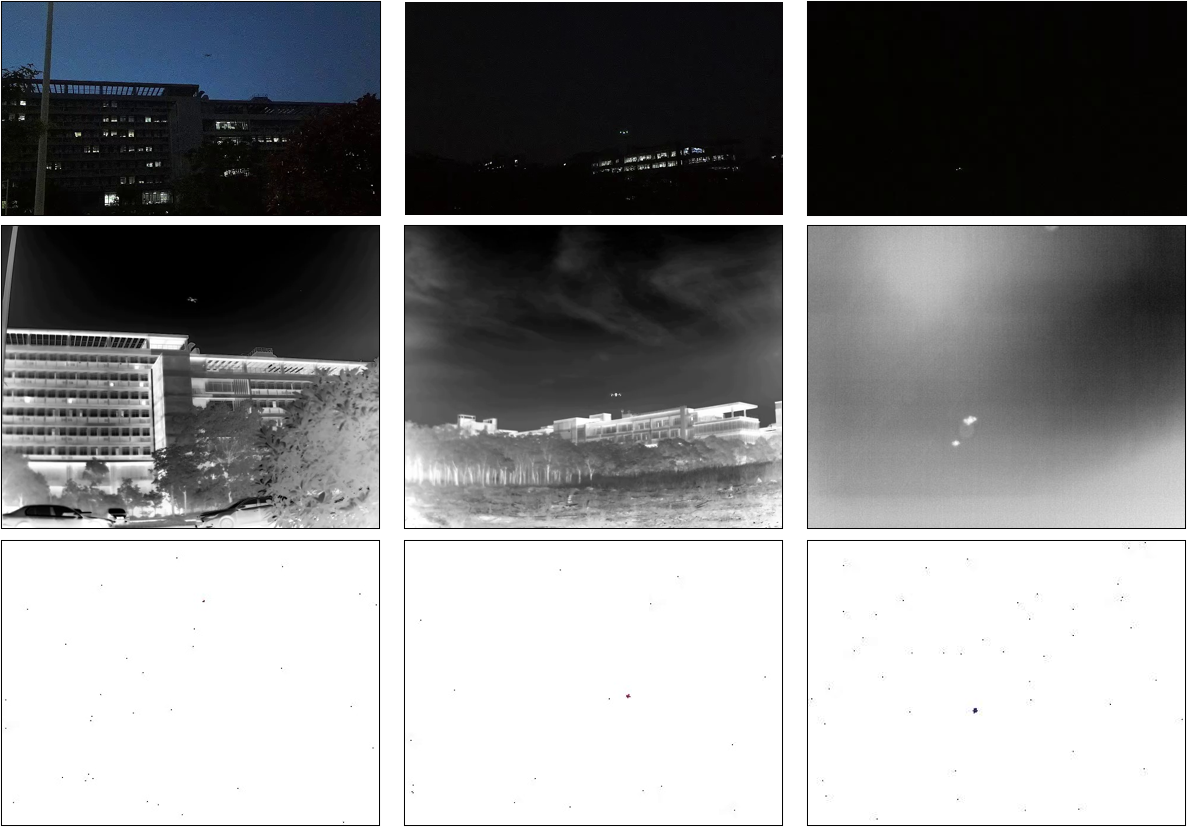}}}
    \end{minipage}
    }

    \caption{Examples of challenging attributes.}
    \label{fig:dataset-challenge-vis}
\end{figure*}

\noindent \textbf{Tiny target (TT)}: To assess the ability to discriminate tiny targets 
an attribute coined as "tiny target" is defined for the UAV size between 8$\times$8 and 16$\times$16 pixels. 
Specifically, a sequence is annotated with this attribute if such targets appear in over 50\% of the entire frames. 
Given the independent annotations for RGB and IR, the "tiny target" attribute is assigned only if both modalities reach this criterion.

\noindent \textbf{Extremely tiny target (ET)}: Building on the definition of "tiny target", "extremely tiny target" denotes those UAVs smaller than 8$\times$8 pixels. 
These targets typically occupy only 0.01\%–0.02\% of the image area, with an almost complete loss of texture and contour features, rendering them highly prone to confusion with background noise or small distractors (\ie bird fragments, light spots). 
To this end, robust discriminative feature extraction capabilities and strong resilience to background clutter are required to preserve trajectory continuity and accuracy. 
Consistent with the tiny target criterion, a sequence is assigned as "extremely tiny target" attribute only if both RGB and IR modalities contain extremely tiny targets in over 50\% of the entire video frames.

\noindent \textbf{Scale variation (SV)}: This attribute is assigned if, for any trajectory, the ratio between the maximum and minimum scale of the same target within a 200-frame window exceeds 4:1. 
Such scenarios arise from abrupt changes in UAV-sensor distance (\ie a sudden departure from a close-range position or rapid long-range approach), visually causing targets to transition from tiny to medium size, or vice versa, in a short time period.
This attribute demands strong scale adaptability in the anti-UAV design, including the use of multi-scale discriminative features during the extraction and effective modelling of non-linear scale changes in the motion prediction modules, to avoid a breakup  of the trajectory or ID confusion.

\noindent \textbf{Fast motion (FM)}: This attribute applies if the distance between the target centroids in two consecutive bounding boxes exceeds 60 pixels, with such frames accounting for over 30\% of the entire sequence. 
High UAV manoeuvrability (\eg diving, sharp turns) often causes inter-frame motion blur and contour distortion, rendering traditional association strategies based on smooth motion assumptions (\eg IOU matching) ineffective.

\noindent \textbf{Low illumination (LI)}: This attribute is assigned when the average brightness of RGB images in a sequence is $\leq$50 (pixel intensity). 
Low illumination typically introduces increased noise, reduced contrast, and loss of texture details in RGB frames, leading to a high risk of detection failure in systems relying on colour or texture features. 
Such scenarios highlight the limitations of using a stand alone  RGB modality and serve as a critical testbed for effective multi-modal fusion.

\noindent \textbf{Similarity clutter (SC)}: A sequence receives this attribute if over 50\% of its frames contain multiple targets with similar scales (bounding box scale ratios between targets do not exceed 2:1). 
UAV swarms often share a similar appearance (\eg uniform rotor structures, comparable flight postures), and combined with near-identical scales, this reduces the discriminability of ReID-extracted appearance embeddings, increasing the risk of target misidentification.

\noindent \textbf{Object overlap (OO)}: 
A sequence is annotated as this attribute if there is more than one instance of bounding box overlap between two trajectories. 
Overlap causes a mutual feature occlusion and local information loss, which can lead to ID association errors, especially for UAVs with similar appearance.

The distributions of these challenging attributes and examples of each attribute are shown in Fig.~\ref{fig:dataset-challenge} and Fig.~\ref{fig:dataset-challenge-vis}, respectively.

\begin{figure*}
    \centering
    \includegraphics[trim={00mm 95mm 30mm 00mm},clip, width=1.0\linewidth]{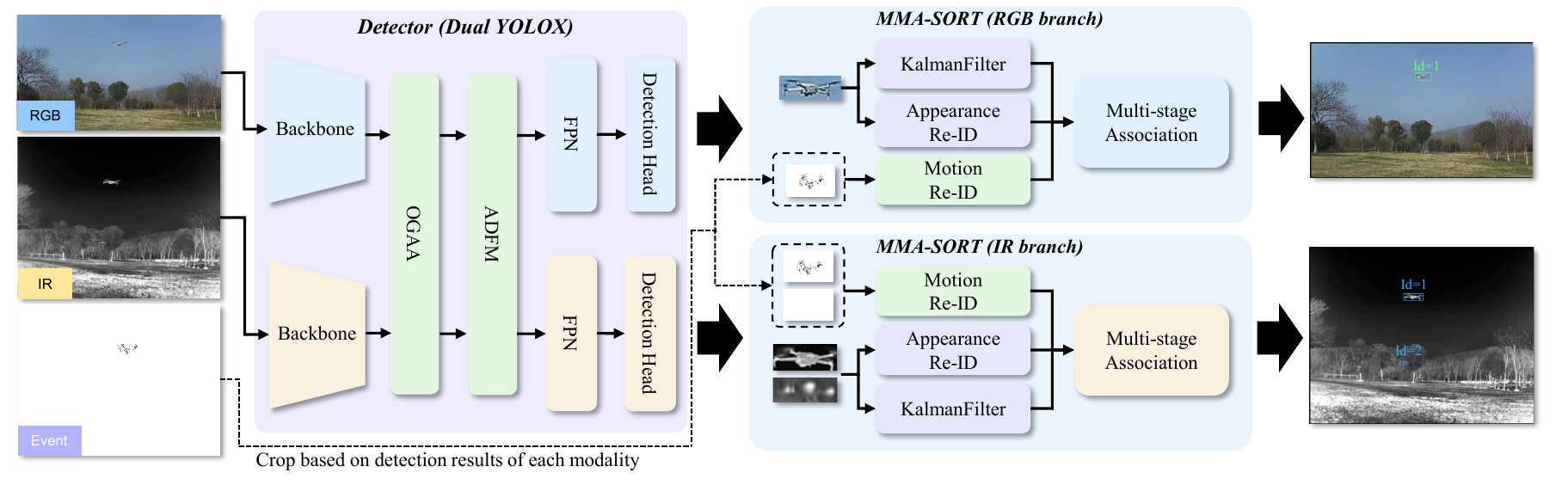}
    \caption{The Overall Framework of MMA-SORT. In the detection stage, the RGB and IR modalities are aligned and fused to exploit their complementary information, tuned to perceive tiny UAV targets. In the tracking stage, the Event modality is incorporated to help to discriminate different UAVs.}
    \label{fig:total-framework}
\end{figure*}
 
\section{Approach}
\subsection{Overview}
Given the real-time response requirements of anti-UAV tracking systems, we adopt the mainstream "Tracking-By-Detection" paradigm. 
In general, regarding the modality fusion strategies, we opt for feature-level fusion between the highly complementary RGB and IR modalities at the detection stage, while incorporating the Event modality for decision-level fusion at the tracking stage. 
This strategic choice stems from the distinct properties of each modality, \ie both RGB and IR modalities can provide relatively reliable and stable representations of the entire scene. 
In contrast, the Event modality, which captures pixel-level illumination intensity changes to precisely characterise the rapid motion trajectories and the instantaneous pose variations of the targets, inherently provides  sparser and less stable information compared to RGB or IR. 
Consequently, it is better to leverage its strengths in motion state modelling at the tracking stage, thereby complementing the traditional appearance features derived from RGB and IR.

The overall framework is illustrated in Fig.~\ref{fig:total-framework}. 
Specifically, we employ a two-stream YOLOX architecture at the detection stage, which receives frames from the RGB modality and IR modality. 
The features produced by their respective independent backbones \(\mathbf{X}_{\text{rgb}}\) and \(\mathbf{X}_{\text{ir}}\) are first aligned via the Offset-Guided Adaptive Alignment Module (OGAA) to produce \(\mathbf{X}'_{\text{rgb}}\) and \(\mathbf{X}'_{\text{ir}}\). 
These aligned features are then fused via the Adaptive Dynamic Fusion Module (ADFM), which adaptively combines each aligned feature \(\mathbf{X}'_{\text{rgb}}\)/\(\mathbf{X}'_{\text{ir}}\) with its corresponding raw counterpart \(\mathbf{X}_{\text{rgb}}\)/\(\mathbf{X}_{\text{ir}}\), to generate the fused features \(\mathbf{X}_{\text{rgb-fused}}\) and \(\mathbf{X}_{\text{ir-fused}}\).

The fused features \(\mathbf{X}_{\text{rgb-fused}}\) and \(\mathbf{X}_{\text{ir-fused}}\) are then fed into the detection head to perceive potential UAVs. 
The detection results are transmitted to the MMA-SORT at the tracking stage to enable multi-object tracking. 
Notably, since annotated data is provided for both modalities, our two-stream framework outputs detection results in the respective coordinate systems of the two modalities simultaneously, offering comprehensive references for applications with unaligned multi-sensor cameras. 
Details of the alignment module, the fusion mechanism, and the MMA-SORT are presented in subsequent sections.

\subsection{Offset-Guided Adaptive Alignment}
This module aims to align the features of one modality to the coordinate system of another, such as aligning \(\mathbf{X}_{\text{rgb}}\) to \(\mathbf{X}_{\text{ir}}\) (or vice versa). 
We propose two alignment strategies, \ie one implicitly learns the offsets relative to the original sampling points through deformable convolutions, while the other explicitly transforms feature coordinates via spatial transformer networks (STN) based on affine transformations.

\begin{figure*}
    \centering
    \subfloat[OGAA (DefConv)]{\includegraphics[trim={00mm 50mm 180mm 00mm},clip, width=0.5\linewidth]{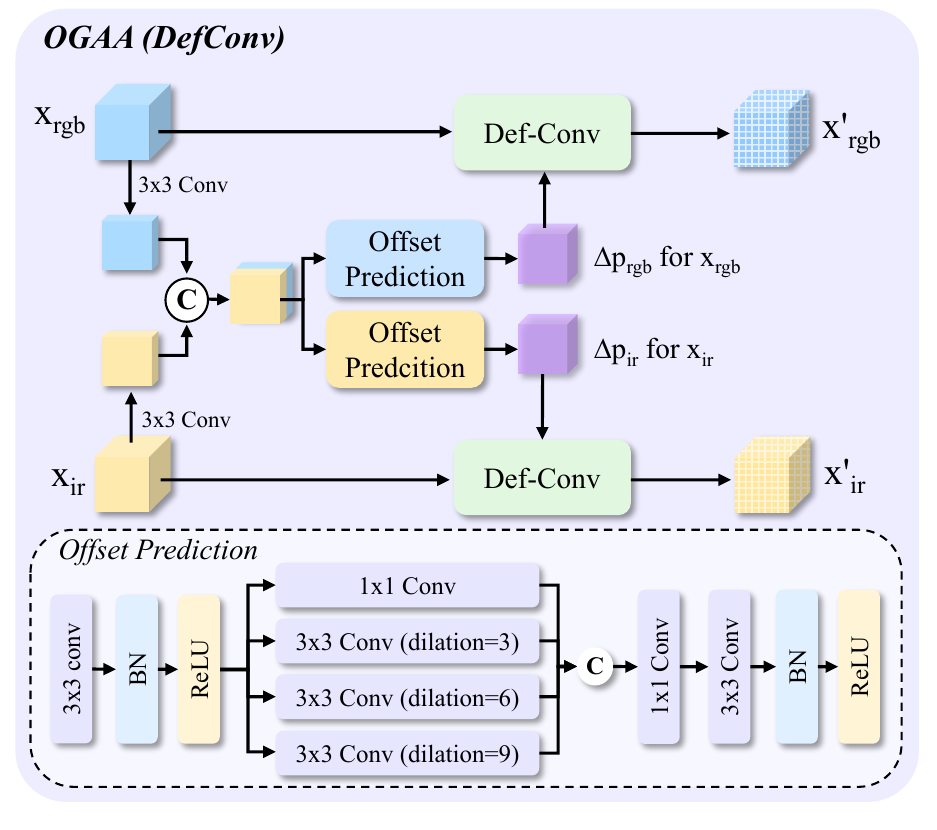}}
    \subfloat[OGAA (STN)]{\includegraphics[trim={00mm 50mm 180mm 00mm},clip, width=0.5\linewidth]{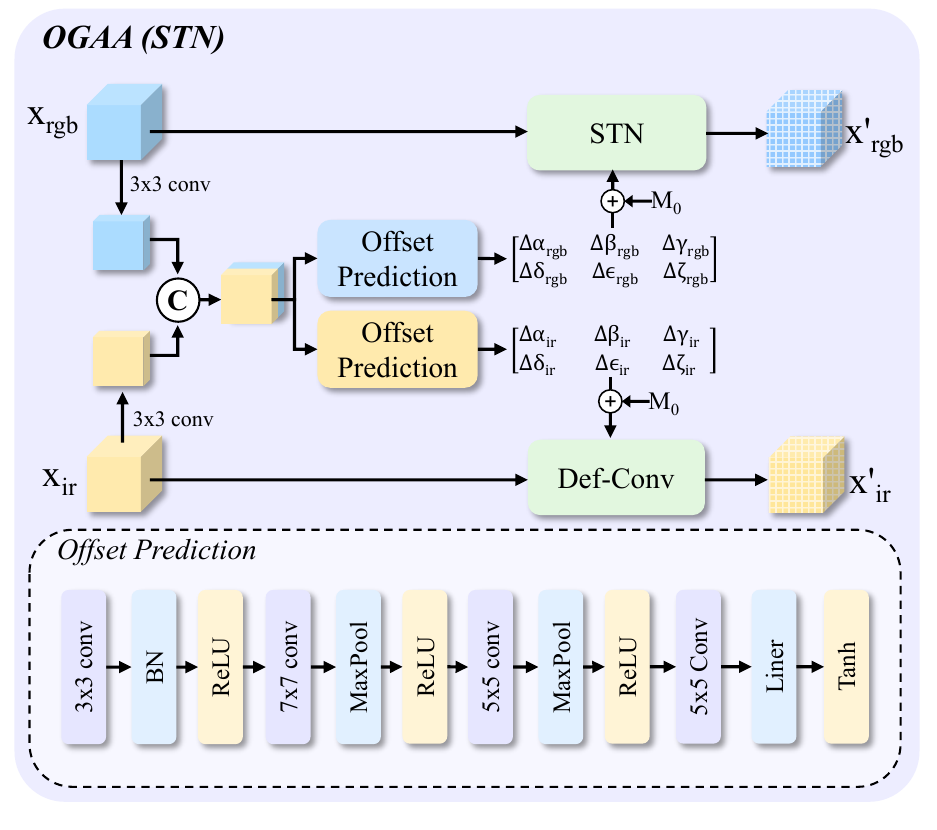}}
    \caption{Offset-Guided Adaptive Alignment Module. (a) shows the OGAA based on the deformable convolution alignment strategy, while (b) presents the OGAA based on the spatial transformation network.}
    \label{fig:align-framework}
\end{figure*}

The deformable convolution-based alignment scheme, shown in Fig.~\ref{fig:align-framework}(a), employs a dual-branch architecture to achieve mutual alignment between the modalities. 
For \(\mathbf{X}_{\text{rgb}}\) and \(\mathbf{X}_{\text{ir}}\) extracted by the backbone network, we first apply a shared \(3\times3\) convolutional layer to reduce the channel dimension and capture common feature representations.


The processed features are then concatenated and fed into two separate branches. 
Each branch incorporates a multi-scale dilated convolution module (dilation rates of 3, 6, and 9) to aggregate contextual information across different receptive fields, providing spatial cues for offset prediction.
Following a dimensionality reduction via \(1\times1\) convolution and local aggregation via \(3\times3\) convolution, the network outputs the offsets \(\Delta \mathbf{p}\) for deformable convolutions.
Finally, \(\mathbf{X}_{\text{rgb}}\) is aligned to the coordinate system of \(\mathbf{X}_{\text{ir}}\) using the offsets \(\Delta \mathbf{p}_{\text{ir}}\) generated by the IR branch, resulting in the aligned feature \(\mathbf{X}'_{\text{rgb}}\). 
Simultaneously, \(\mathbf{X}_{\text{ir}}\) is adapted to the RGB coordinate system using \(\Delta \mathbf{p}_{\text{rgb}}\), yielding \(\mathbf{X}'_{\text{ir}}\). 
The outputs can be formulated as:
\begin{equation}
\left\{
\begin{aligned}
    \mathbf{X}'_{\text{rgb}} =& \text{DefConv}^{\text{rgb}}(\mathbf{X}_{\text{rgb}}, \Delta \mathbf{p}_{\text{ir}})\\
    \mathbf{X}'_{\text{ir}} =& \text{DefConv}^{\text{ir}}(\mathbf{X}_{\text{ir}}, \Delta \mathbf{p}_{\text{rgb}})
    \end{aligned}\right..
\end{equation}
This implicit alignment mechanism learns optimal sampling strategies without explicitly computing a transformation matrix.

The STN-based alignment scheme, shown in Fig.~\ref{fig:align-framework}(b), leverages prior knowledge from the dataset to simplify the learning process. 
Considering the difficulty of directly learning a full affine transformation, we precompute an initial affine matrix \(\mathbf{M}_0\) and train the network to predict offsets \(\Delta\mathbf{M}\) relative to this matrix. 
Specifically, \(\mathbf{X}_{\text{rgb}}\) and \(\mathbf{X}_{\text{ir}}\) undergo shared \(3\times3\) convolution for feature extraction and dimension reduction, followed by concatenation before being input into the dual branches. This process is the same as the deformable convolution-based alignment scheme.
Each branch employs multiple convolutional layers and downsampling operations to refine spatial transformation features, culminating in the prediction of six offset parameters \([\Delta\alpha, \Delta\beta, \Delta\gamma, \Delta\delta, \Delta\epsilon, \Delta\zeta]\) via a Tanh activation function.
The final affine transformation matrices are computed as:
\begin{equation}
\left\{
\begin{aligned}
\mathbf{M}_{\text{rgb}} =& 
\underbrace{\begin{bmatrix} a_{\text{rgb}} & b_{\text{rgb}} & c_{\text{rgb}} \\ d_{\text{rgb}} & e_{\text{rgb}} & f_{\text{rgb}} \end{bmatrix}}_{\mathbf{M}_{0,\text{rgb}}} 
+ 
\underbrace{\begin{bmatrix} \Delta\alpha_{\text{rgb}} & \Delta\beta_{\text{rgb}} & \Delta\gamma_{\text{rgb}} \\ \Delta\delta_{\text{rgb}} & \Delta\epsilon_{\text{rgb}} & \Delta\zeta_{\text{rgb}} \end{bmatrix}}_{\Delta\mathbf{M}_{\text{rgb}}}\\
\mathbf{M}_{\text{ir}} =& 
\underbrace{\begin{bmatrix} a_{\text{ir}} & b_{\text{ir}} & c_{\text{ir}} \\ d_{\text{ir}} & e_{\text{ir}} & f_{\text{ir}} \end{bmatrix}}_{\mathbf{M}_{0,\text{ir}}} 
+ 
\underbrace{\begin{bmatrix} \Delta\alpha_{\text{ir}} & \Delta\beta_{\text{ir}} & \Delta\gamma_{\text{ir}} \\ \Delta\delta_{\text{ir}} & \Delta\epsilon_{\text{ir}} & \Delta\zeta_{\text{ir}} \end{bmatrix}}_{\Delta\mathbf{M}_{\text{ir}}}
\end{aligned}\right..
\end{equation}
These matrices drive the STN modules to transform \(\mathbf{X}_{\text{rgb}}\) and \(\mathbf{X}_{\text{ir}}\), generating aligned features \(\mathbf{X}'_{\text{rgb}}\) and \(\mathbf{X}'_{\text{ir}}\):
\begin{equation}\left\{
\begin{aligned}
    \mathbf{X}'_{\text{rgb}} =& \text{STN}(\mathbf{X}_{\text{rgb}}, \mathbf{M}_{\text{rgb}})\\
    \mathbf{X}'_{\text{ir}} =& \text{STN}(\mathbf{X}_{\text{ir}}, \mathbf{M}_{\text{ir}})
\end{aligned}\right..
\end{equation}
It should be noted that this explicit alignment mechanism effectively compensates for spatial misalignment by modelling global coordinate transformations.

\begin{figure}
    \centering
    \includegraphics[trim={00mm 110mm 115mm 00mm},clip,width=1.0\linewidth]{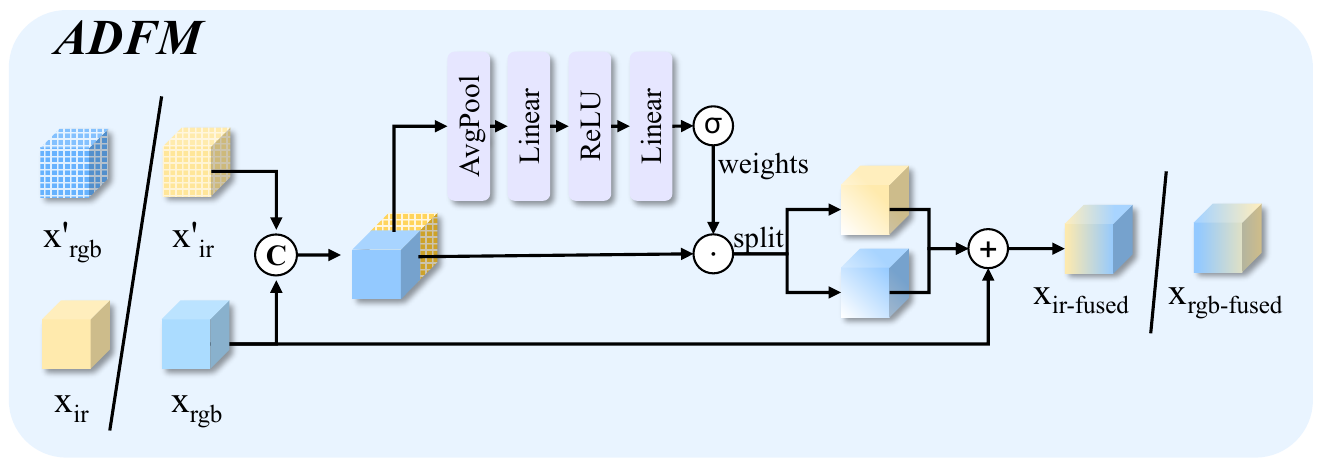}
    \caption{The Adaptive Dynamic Fusion Module. For the RGB branch, the module takes the original RGB feature $X_{\textbf{rgb}}$ (before OGAA) and the IR feature aligned to the RGB space $X'_{\text{ir}}$, and outputs the fused RGB feature $X_{\text{rgb-fused}}$. For the IR branch, it takes the original IR feature $X_{\text{ir}}$ (before OGAA) and the RGB feature aligned to the IR space $X'_{\text{rgb}}$, and outputs the fused IR feature $X_{\text{ir-fused}}$.}
    \label{fig:AWFM}
\end{figure}

\subsection{Adaptive Dynamic Fusion Module}
To dynamically balance the contributions of the two modalities, we adopt a simple attention-weighted channel fusion mechanism. 
For each branch of the two modalities, the module takes two inputs: one is the original input feature of the current modality, and the other is the feature of the other modality which has been aligned to the current one. 

Taking the RGB branch as an example, ADFM receives the original feature \(\mathbf{X}_{\text{rgb}}\) and the aligned feature \(\mathbf{X}'_{\text{ir}}\). 
These two features are first concatenated along the channel dimension:
\begin{equation}
    \mathbf{F}_{\text{concat}, rgb} = \text{Concat}\left( \mathbf{X}_\text{rgb}, \mathbf{X}'_\text{ir} \right),
\end{equation}
and then input to an additional channel attention branch, where the channel weights \(W_\text{rgb}^{\text{rgb}}\) and \(W_\text{ir}^{\text{rgb}}\) are obtained by:
\begin{equation}\left\{
\begin{aligned}
    \mathbf{F}_{\text{gap}, rgb} =& \text{ReLU}\left( \text{Linear}\left( \text{AvgPool}\left( F_{\text{concat}, rgb} \right) \right) \right)\\
    \mathbf{W}_\text{rgb}^{\text{rgb}} =& \text{Sigmoid}\left( \text{Liner}\left( \mathbf{F}_{\text{gap}, rgb} \right) \right)\\
    \mathbf{W}_\text{ir}^{\text{rgb}} =& \text{Sigmoid}\left( \text{Liner}\left( \mathbf{F}_{\text{gap}, rgb} \right) \right)
    \end{aligned}\right..
\end{equation}
The original feature \(\mathbf{X}_{\text{rgb}}\) is multiplied by \(W_\text{rgb}^{\text{rgb}}\), and the aligned feature \(\mathbf{X}'_{\text{ir}}\) is multiplied by \(W_\text{ir}^{\text{rgb}}\).
To preserve the original feature information, residual connections are introduced. The final fused feature is then given by:
\begin{equation}
    \mathbf{X}_\text{rgb-fused} = \left( \mathbf{X}_\text{rgb} \odot W_\text{rgb}^{\text{rgb}} \right) + \left( \mathbf{X}_\text{ir}' \odot W_\text{ir}^{\text{rgb}} \right) + \mathbf{X}_{\text{rgb}}
\end{equation}
Similarly, the operation for the IR modality branch is as follows:
\begin{equation}
    \mathbf{X}_\text{ir-fused} = \left( \mathbf{X}_\text{ir} \odot W_\text{rgb}^{\text{ir}} \right) + \left( \mathbf{X}_\text{rgb}' \odot W_\text{ir}^{\text{ir}} \right) + \mathbf{X}_{\text{ir}}
\end{equation}
This design dynamically balances the contributions of the aligned and raw features, ensuring robust fusion, while mitigating information loss.

\subsection{MMA-SORT}
\begin{figure*}
    \centering
    \includegraphics[trim={00mm 100mm 40mm 00mm},clip,width=1.0\linewidth]{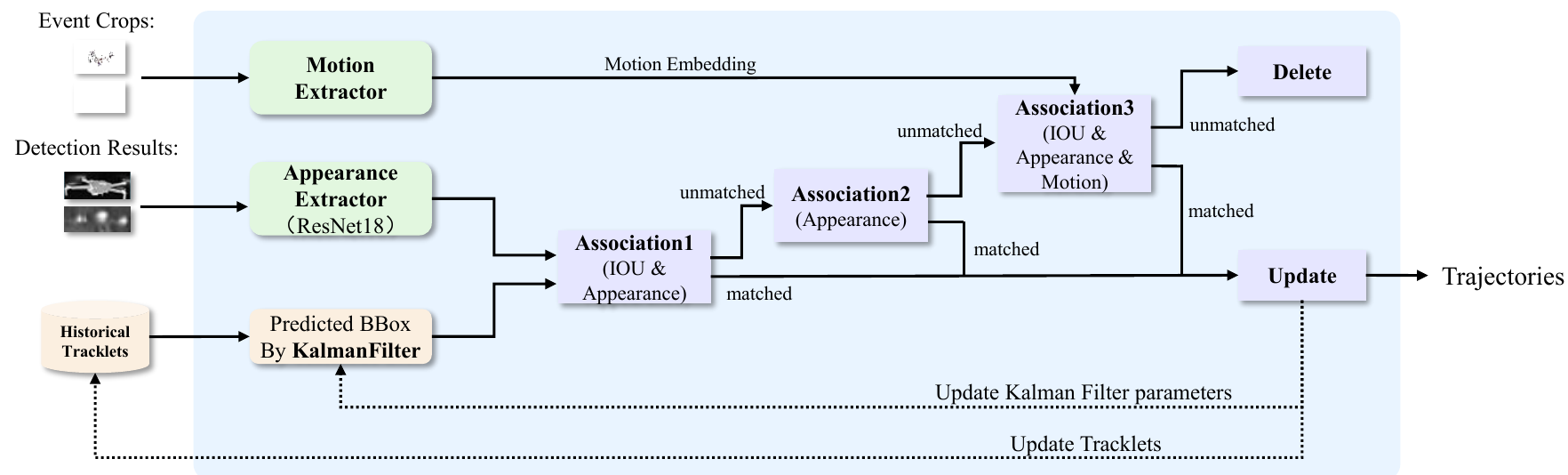}
    \caption{The tracking pipeline of our solution: MMA-SORT.}
    \label{fig:MMA-SORT}
\end{figure*}

Existing multi-object tracking models (such as SORT~\cite{sort} and its variants~\cite{DeepSORT,deep-ocsort,ocsort,bytetrack,botsort,boostTrack,boostTrack++}) exhibit notable limitations in UAV-target tracking tasks. 
First, the small size and complex motion patterns of UAV targets pose a significant challenge to motion modelling. 
The linear motion assumption inherent to the Kalman filter often leads to substantial discrepancies between the predicted bounding boxes (based on previous frames) and the actual detections in the current frame.
This renders trackers centred on IOU metrics ineffective in achieving cross-frame identity association, thereby resulting in a large number of ID switches (IDSW). 
To address this issue, we downplay the role of IOU metrics in the process of target identity association.
Unlike traditional trackers that treat IOU matching as an essential prerequisite for identity association, we argue that Kalman filters are less reliable in modelling the motion of small, irregularly moving UAV targets, compared to targets like pedestrians and vehicles. 
Thus, rigidly adhering to IOU matching as a prerequisite in multi-UAV tracking tasks is overly restrictive.

Second, UAV targets provide insufficient discriminative appearance clues. 
Specifically, when the targets are small, relying solely on appearance embedding makes it challenging to distinguish between UAVs of different identities.
It is also difficult to separate UAV targets from background objects effectively. 
To tackle this, we introduce the Event modality to depict the motion states of UAVs, using motion embedding to assist appearance embedding in target identity discrimination. 
Considering the strict real-time requirements of UAV tracking tasks, we adopt a learning-free approach, directly leveraging pixel-level statistical information from raw Event modality images to generate motion embeddings. 
This avoids any speed degradation that would be caused by introducing time consuming ReID modules. 

Based on the above analysis, we propose a new tracker incorporating the Event modality, with its detailed pipeline illustrated in Fig.~\ref{fig:MMA-SORT}. 
The tracker receives the detection results exceeding the confidence threshold, with the association process being divided into three stages.

\noindent \textbf{In the first stage}, for all input detections and confirmed trajectories, IOU matching is combined with appearance embedding to distinguish the targets. 
The distance between the detections and the trajectories is calculated as follows:
\begin{equation}
     \mathbf{d} = \min(d_\text{iou}, d_\text{app}), 
\end{equation}
where \(\mathbf{d}\) represents the final distance, \(d_{\text{iou}}\) is the IOU distance between the detection box and the trajectory's predicted box, and \(d_{\text{app}}\) is the appearance embedding distance between the detection and the trajectory. 
The calculation for IOU distance and appearance distance follows those in BoT-SORT~\cite{botsort}.

\noindent \textbf{In the second stage}, for unmatched detections and remaining trajectories, appearance embedding is used to recover the targets with IOU mismatches. 
The calculation of the appearance distance \(d_{\text{app}}\) between detection boxes and trajectories in this stage is the same as  in the first stage.

\noindent \textbf{In the third stage}, for the remaining detections and unconfirmed trajectories, matching is performed by combining IOU, appearance embedding, and motion embedding. 
Considering the sparsity of information in Event modality images and the real-time requirements of the task, we do not employ ReID models for layer-wise feature extraction. 
Instead, we directly compute statistical features of non-background pixels within the target region and explicitly model them as a 7-dimensional vector:
\begin{equation}
\mathbf{E}_{\text{motion}} = \left[ 
\begin{array}{c} 
N_{\text{non-bg}},
\mu_x,
\sigma_x,
\mu_y,
\sigma_y,
S_x,
S_y 
\end{array} 
\right],
\end{equation}
where each dimension of the motion embedding represents the following statistics, \ie the number of non-background pixels \(N_{\text{non-bg}}\), normalized X mean \(\mu_x\), normalized X standard deviation \(\sigma_x\), normalized Y mean \(\mu_y\), normalized Y standard deviation \(\sigma_y\), normalized X range \(S_x\), and normalized Y range \(S_y\). 
The computation of the motion embedding distance follows the cosine similarity definition:
\begin{equation}
d_{\text{motion}} = 1 - \frac{\mathbf{E}_{\text{motion}}^i \cdot \mathbf{E}_{\text{motion}}^j}{\|\mathbf{E}_{\text{motion}}^i\| \cdot \|\mathbf{E}_{\text{motion}}^j\|},
\end{equation}
where \( d_{\text{motion}} \) denotes the motion embedding distance between two targets, \( \mathbf{E}_{\text{motion}}^i \) and \( \mathbf{E}_{\text{motion}}^j \) represent the motion embeddings of the \( i \)-th and \( j \)-th targets respectively, \( \cdot \) indicates the dot product operation, and \( \|\cdot\| \) denotes the L2 norm of the vector.
After computing the IOU distance \(d_{\text{iou}}\), appearance distance\(d_{\text{app}}\), and motion distance \(d_{\text{motion}}\), analogous to the IOU distance and appearance distance, we define the motion distance \(\boldsymbol{d}_m\) with a threshold for filtering, formulated as:
\begin{equation}
    d_\text{motion} = 
    \begin{cases} 
    d_m, & \text{if } d_m \leq \theta_\text{motion} \\
    1, & \text{otherwise}
    \end{cases}.
\end{equation}
The final distance \(\mathbf{d}\) for this stage is:
\begin{equation}
    \mathbf{d} = 
    \begin{cases}
        \max(d_{\text{app}}, d_{\text{motion}}), & \text{if } d_{\text{iou}} \leq \theta_\text{iou} \\
        1, & \text{otherwise}
    \end{cases}
\end{equation}

The purpose of introducing motion embedding in this stage is to address the issue where, in complex backgrounds, false positive samples of UAVs with similar appearance easily emerge. 
These interfering objects are initialised as unconfirmed trajectories, disrupting subsequent matching processes. 
The motion embedding can prevent incorrect associations between targets and interfering backgrounds, reduce label misalignment, and thereby mitigate background noise interference.
In each stage, the Hungarian matching algorithm is applied to the obtained distance matrix to achieve optimal matching between the detections and trajectories. 

It should be noted that the Event modality exhibits certain spatio-temporal misalignment with the RGB and IR modalities. 
Additionally, its lack of detailed texture information makes pre-alignment via traditional methods difficult.
Meanwhile, considering that the dataset is tailored for UAV tracking tasks in sparse, open scenes, instances of dense target overlap are extremely rare. 
Therefore, when cropping the targets from Event images, we alleviate sensor-induced spatio-temporal misalignment between modalities by expanding the cropping range (extending 20 pixels outward in both width and height directions). 

\section{Evaluation}
To provide a comprehensive reference for future studies, we conduct extensive evaluations of state-of-the-art multi-object tracking methods on our dataset. 
Furthermore, the superiority of the proposed multimodal tracker is validated, demonstrating that our multimodal fusion framework significantly enhances the robustness of anti-UAV tracking.

\subsection{Implementation Details}
In the proposed framework, the detector’s backbone, the FPN module, and the detection head follow the original YOLOX~\cite{YOLOX}. 
For RGB images (640×360) and IR images (640×512) from the two modalities, we first resize them to 640×640 using the standard YOLOX preprocessing pipeline before feeding them into the backbone for feature extraction.
The proposed OGAA and ADFM modules are inserted between the backbone and FPN. 
They are responsible for cross-modal alignment and fusion of the three-scale feature maps extracted independently by the two modality-specific backbones. 
The fused multi-scale features are then forwarded to the respective FPN modules for subsequent detection, and the final detection results are passed to the MMA-SORT tracker for multi-object tracking.

During training, the dual-stream architecture produces independent detection outputs for each modality. 
Consequently, the detection losses of the RGB and IR heads are computed separately and backpropagated simultaneously, while the loss functions for each modality remain consistent with YOLOX.

In principle, YOLO-based detectors typically rely on diverse data augmentation strategies (\eg random colour jittering, Mosaic, and Mixup) to improve robustness. 
However, such multi-image mixing and random geometric transformations impose additional challenges for aligning and fusing unregistered multimodal features.
This issue is particularly critical for STN-based alignment, as these augmentations, while enhancing robustness to complex scenarios, often disrupt the inherent spatial structure of the input, preventing the STN from learning an effective affine transformation. 
To address this, we adopt a two-stage training strategy. 
In the first stage, we train the network for 100 epochs with data augmentation, updating only the backbones, FPNs, and detection heads of each modality independently, without cross-modal interaction. 
In the second stage, we freeze the backbone and FPN weights, disable data augmentation, and train the OGAA and ADFM modules for 50 epochs while fine-tuning the detection heads. 
The first stage uses SGD with an initial learning rate of 5e-3, momentum of 0.9, and batch size of 32. 
The second stage employs AdamW, with the same learning rate and batch size. 
All training and inference are performed on a single NVIDIA RTX 4090 GPU.

In terms of the STN-based alignment strategy, the initial affine transformation matrix is computed via keypoint matching. 
A pair of RGB and IR images is selected from the dataset, and a conventional keypoint detection method is used to derive the matrix. 
Since the relative positions of the RGB and IR cameras remain fixed during data collection, the same matrix can be applied across all sequences, serving as a global and unique initialisation. 
Finally, two affine matrices are computed, one aligning RGB to IR and the other aligning IR to RGB.

For the appearance-based ReID module, given the limited visual information of UAV targets, we directly adopt the ReID model and training strategy of DeepSORT~\cite{DeepSORT}, using a ResNet backbone to extract discriminative appearance features.

During inference, the detection confidence threshold is 0.3. 
For the tracker, the IoU threshold \( \theta_\text{iou} \) is 0.9, the appearance threshold is 0.2, the motion threshold \( \theta_{\text{motion}} \) is 0.2, and the confidence threshold for initialising new tracks is 0.7.

\subsection{Quantitative Comparison}
\begin{table*}[t]
\centering
\caption{A comparison of different multi-object tracking methods on RGB and IR datasets. The best results are shown in \textbf{bold}, and second best are \underline{underlined}. In addition, within the ``Tracking-By-Detection'' paradigm, methods marked with an asterisk ($*$) denote the use of a ReID module.
}
\label{tab:comparative_experiments_results}
\resizebox{\textwidth}{!}{
\begin{tabular}{lc|cccc|cccc|c}
\toprule
\multirow{2}{*}{Method} & \multirow{2}{*}{Year} &
\multicolumn{4}{c|}{RGB} & \multicolumn{4}{c|}{IR} & \multirow{2}{*}{FPS} \\
\cmidrule(lr){3-6} \cmidrule(lr){7-10}
 & & MOTA$\uparrow$ & HOTA$\uparrow$ & IDF1$\uparrow$ & IDs$\downarrow$ 
 & MOTA$\uparrow$ & HOTA$\uparrow$ & IDF1$\uparrow$ & IDs$\downarrow$ & \\
\midrule
\multicolumn{11}{l}{\textbf{Tracking by Detection}} \\
DeepSORT*       & 2017 &  8.31 & 29.91 & 30.73 & 2342 & 41.50 & 38.30 & 44.09 & 2038 &  18 \\
BoT-SORT        & 2022 & 60.88 & 44.37 & 51.89 &  988 & 78.18 & 53.93 & 58.52 &  968 & 128 \\
BoT-SORT*       & 2022 & 60.87 & 44.38 & 51.90 &  977 & 78.16 & 53.96 & 58.56 &  968 &  90 \\
OC-SORT         & 2023 & 59.51 & 42.71 & 47.68 & 1592 & 76.40 & 52.97 & 55.82 & 1453 & \textbf{183} \\
Deep OC-SORT*   & 2023 & 59.51 & 42.70 & 47.67 & 1594 & 76.38 & 52.98 & 55.82 & 1460 & 153 \\
BoostTrack      & 2024 & 54.66 & 40.13 & 43.55 & 2040 & 71.64 & 49.02 & 51.04 & 1734 & \underline{171} \\
BoostTrack*     & 2024 & 57.03 & 43.78 & 50.15 & 1308 & 73.63 & 52.11 & 56.39 & 1209 & 153 \\
BoostTrack++    & 2024 & 55.35 & 40.63 & 43.98 & 2024 & 72.44 & 49.45 & 51.70 & 1711 & 165 \\
BoostTrack++*   & 2024 & 57.80 & 44.33 & 50.68 & 1288 & 74.51 & 52.78 & 57.51 & 1193 & 156 \\
\midrule
\multicolumn{11}{l}{\textbf{Tracking by Query}} \\
TransTrack      & 2020 & 54.33 & 38.65 & 40.62 & 2237 & 70.53 & 46.95 & 48.90 & 1963 &  16 \\
MOTR            & 2021 & 27.26 & 24.60 & 27.14 & 3243 & 18.99 & 40.70 & 44.39 & 1492 &  17 \\
MOTRv2          & 2022 & 58.84 & 46.36 & 55.24 &  656 & 69.95 & 52.51 & 58.10 &  754 &  23 \\
MeMOTR          & 2023 & 26.52 & 14.02 & 14.22 & 1964 & 33.14 & 45.26 & 53.87 &  530 &  21 \\
MOTIP           & 2025 & 62.72 & 53.04 & 65.06 &  581 & 76.54 & 59.67 & 70.23 &  832 &  23 \\
\midrule
\multicolumn{11}{l}{\textbf{Ours}} \\
MMA-SORT (DefConv) & 2025 & \textbf{63.26} & \underline{56.15} & \underline{74.92} & \underline{159} & \underline{79.48} & \underline{66.57} & \underline{83.75} & \textbf{131} &  47 \\
MMA-SORT (STN) & 2025 & \underline{63.08} & \textbf{56.43} & \textbf{74.99} & \textbf{151} & \textbf{79.91} & \textbf{66.81} & \textbf{84.62} & \underline{138} &  81 \\
\bottomrule
\end{tabular}}
\end{table*}
For comparative evaluation, two major categories of state-of-the-art multi-object tracking methods were systematically evaluated, including Tracking by Detection approaches (DeepSORT~\cite{DeepSORT}, BoT-SORT~\cite{botsort}, OC-SORT~\cite{ocsort}, Deep OC-SORT~\cite{deep-ocsort}, BoostTrack~\cite{boostTrack}, BoostTrack++~\cite{boostTrack++}) and Tracking by Query approaches (TransTrack~\cite{transtrack}, MOTR~\cite{motr}, MOTRv2~\cite{motrv2}, MeMOTR~\cite{memotr}, MOTIP~\cite{motip}). 
All trackers are retrained and tested on our MM-UAV to enable a fair comparison. 
Since existing methods are designed for unimodal tracking, experiments are conducted independently on the RGB and IR modalities. 
The results are summarised in Table~\ref{tab:comparative_experiments_results}. 
For fairness, all Tracking by Detection methods share the same detector weights, while Tracking by Query methods were retrained following their default configurations.

The performance is evaluated using standard MOT metrics~\cite{mota,hota}, including MOTA, IDF1, HOTA, and ID switch. 
With the deformable convolution-based alignment, our solution achieves 63.26 MOTA, 56.15 IDF1, 74.92 HOTA, and 159 IDs on the RGB modality, and 79.48, 66.57, 83.75, and 131 on the IR modality. 
With the STN-based alignment, the results were 63.08, 56.43, 74.99, and 151 on RGB, and 79.91, 66.81, 84.62, and 138 on IR. 
These results demonstrate that the proposed multimodal fusion method consistently outperforms existing unimodal approaches across all metrics, with particularly notable improvements in HOTA and IDF1. 
For example, compared with the strongest unimodal baseline MOTIP, the deformable alignment strategy improves HOTA and IDF1 by 3.11\% and 9.86\% on the RGB modality, while the STN-based strategy achieves gains of 3.39\% and 9.93\%, respectively.

Besides, as real-time performance is crucial for UAV tracking applications, inference speed (FPS) was also reported.
Compared with transformer-based Tracking by Query approaches, Tracking by Detection approaches generally exhibit higher speed thanks to the lightweight YOLO-based detectors. 
Although the proposed multi-modal method introduces additional alignment and fusion operations, delivering a slower speed than unimodal Tracking by Detection methods, it achieves substantially more robust tracking, while still satisfying the real-time requirements.
\begin{figure}
    \centering
    \includegraphics[width=1.0\linewidth]{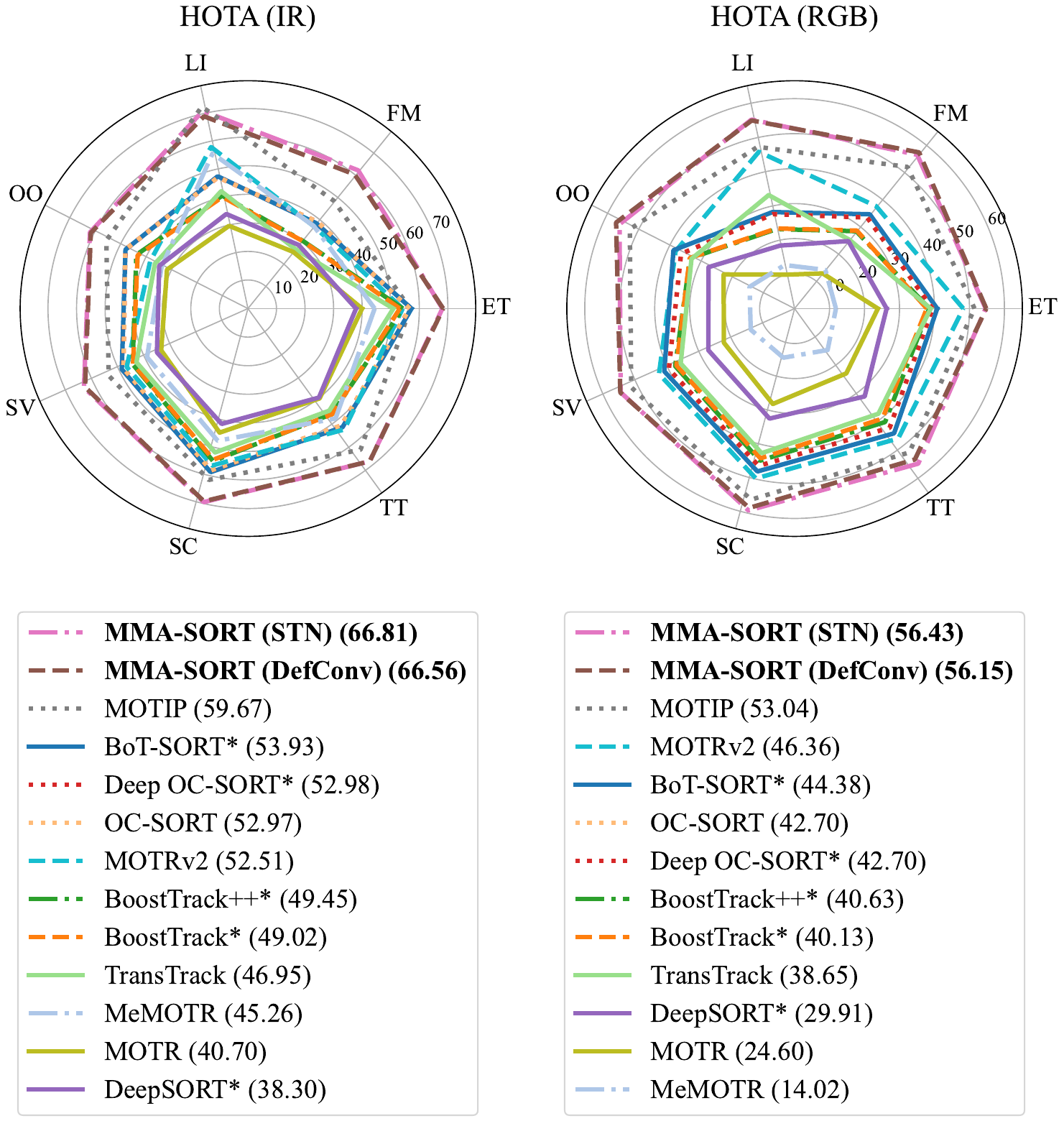}
    \caption{A performance comparison of all trackers across different challenging scenarios. The values in parentheses represent the overall HOTA scores, and all trackers are sorted and displayed in descending order based on performance.}
    \label{fig:results-challenges}
\end{figure}
To provide a comprehensive assessment of tracker performance in various challenging scenarios, we conduct a detailed evaluation of all methods under multiple representative conditions, as presented in Fig.~\ref{fig:results-challenges}. 
To fairly and intuitively quantify the effectiveness of tracking, we employ the holistic HOTA metric. 
The experimental results demonstrate that our approach consistently outperforms existing multi-object trackers across nearly all scenarios, with particularly pronounced superiority under fast motion (FM) and low illumination (LI) conditions. 
These findings strongly indicate that the proposed multimodal fusion framework exploits the complementary information across modalities, thereby achieving a stronger identity association.

\subsection{Ablation Studies}

\begin{table*}[t]          
\centering
\caption{An ablation Study of Different Components under the Alignment Strategy Based on Deformable Convolution.}  
\label{tab:components-def}     
\resizebox{\textwidth}{!}{
\begin{tabular}{l c c c c c c c c c c c c c c}  
\toprule
\multirow{2}{*}{Method} & \multicolumn{2}{c}{Detector} & \multicolumn{3}{c}{Tracker} & \multicolumn{4}{c}{RGB} & \multicolumn{4}{c}{IR} & \multirow{2}{*}{FPS} \\
\cmidrule(lr){2-3}  \cmidrule(lr){4-6}  \cmidrule(lr){7-10}  \cmidrule(lr){11-14} 
& Align & Fusion & ReID & MMA-SORT & Event & MOTA$\uparrow$ & HOTA$\uparrow$ & IDF1$\uparrow$ & IDs$\downarrow$ & MOTA$\uparrow$ & HOTA$\uparrow$ & IDF1$\uparrow$ & IDs$\downarrow$  \\
\midrule
DualYOLOX+Bot-SORT       &        &        &        &        &        & 59.97 & 43.66 & 50.78 & 989  & 77.69 & 52.93 & 58.21 & 995  & \textbf{130} \\
+Deformable Align        & $\boldsymbol{\checkmark}$ &        &        &        &        & 60.77 & 43.95 & 50.51 & 1024 & 77.67 & 53.50 & 58.12 & 1024 & 58  \\
+Adaptive Weight Fusion  & $\boldsymbol{\checkmark}$ & $\boldsymbol{\checkmark}$ &        &        &        & 60.89 & 44.05 & 50.65 & 1020 & 77.75 & 53.76 & 58.37 & 966  & 56  \\
+Appearance ReID        & $\boldsymbol{\checkmark}$ & $\boldsymbol{\checkmark}$ & $\boldsymbol{\checkmark}$ &        &        & 61.71 & 45.44 & 51.65 & 1277 & 78.04 & 53.50 & 58.02 & 1102 & 48  \\
+MMA-SORT (without Event) & $\boldsymbol{\checkmark}$ & $\boldsymbol{\checkmark}$ & $\boldsymbol{\checkmark}$ & $\boldsymbol{\checkmark}$ &        & 62.97 & 56.07 & 74.65 & 163  & 79.41 & 66.38 & 83.44 & 144  & 48  \\
+Motion Embedding        & $\boldsymbol{\checkmark}$ & $\boldsymbol{\checkmark}$ & $\boldsymbol{\checkmark}$ & $\boldsymbol{\checkmark}$ & $\boldsymbol{\checkmark}$ & \textbf{63.26} & \textbf{56.15} & \textbf{74.92} & \textbf{159}  & \textbf{79.48} & \textbf{66.57} & \textbf{83.75} & \textbf{131}  & 47  \\
\bottomrule
\end{tabular}}
\end{table*}

\begin{table*}[t]          
\centering
\caption{An ablation Study of Different Components under the Alignment Strategy Based on STN.}  
\label{tab:components-stn}     
\resizebox{\textwidth}{!}{
\begin{tabular}{l c c c c c c c c c c c c c c}
\toprule
\multirow{2}{*}{Method} & \multicolumn{2}{c}{Detector} & \multicolumn{3}{c}{Tracker} & \multicolumn{4}{c}{RGB} & \multicolumn{4}{c}{IR} & \multirow{2}{*}{FPS} \\
\cmidrule(lr){2-3}  \cmidrule(lr){4-6}  \cmidrule(lr){7-10}  \cmidrule(lr){11-14}
& Align & Fusion & ReID & MMA-SORT & Event & MOTA$\uparrow$ & HOTA$\uparrow$ & IDF1$\uparrow$ & IDs$\downarrow$ & MOTA$\uparrow$ & HOTA$\uparrow$ & IDF1$\uparrow$ & IDs$\downarrow$  \\
\midrule
DualYOLOX+Bot-SORT       &        &        &        &        &        & 59.97 & 43.66 & 50.78 & 989  & 77.70 & 52.93 & 58.21 & 995  & \textbf{130} \\
+STN Align               & $\boldsymbol{\checkmark}$ &        &        &        &        & 58.04 & 41.68 & 48.72 & 963  & 76.74 & 52.55 & 57.71 & 1051 & 107  \\
+Adaptive Weight Fusion  & $\boldsymbol{\checkmark}$ & $\boldsymbol{\checkmark}$ &        &        &        & 60.88 & 44.50 & 51.29 & 1029 & 77.83 & 53.82 & 58.40 & 1017 & 102  \\
+Appearance ReID        & $\boldsymbol{\checkmark}$ & $\boldsymbol{\checkmark}$ & $\boldsymbol{\checkmark}$ &        &        & 60.92 & 44.40 & 51.24 & 1018 & 77.80 & 53.81 & 58.23 & 1010 & 82  \\
+MMA-SORT (without Event) & $\boldsymbol{\checkmark}$ & $\boldsymbol{\checkmark}$ & $\boldsymbol{\checkmark}$ & $\boldsymbol{\checkmark}$ &        & 63.07 & 56.37 & 74.90 & 153  & 79.86 & 66.65 & 84.30 & 149  & 82  \\
+Motion Embedding        & $\boldsymbol{\checkmark}$ & $\boldsymbol{\checkmark}$ & $\boldsymbol{\checkmark}$ & $\boldsymbol{\checkmark}$ & $\boldsymbol{\checkmark}$ & \textbf{63.08} & \textbf{56.43} & \textbf{74.99} & \textbf{151}  & \textbf{79.91} & \textbf{66.81} & \textbf{84.62} & \textbf{138}  & 81  \\
\bottomrule
\end{tabular}}
\end{table*}

\begin{figure}
    \centering
    \includegraphics[trim={00mm 20mm 110mm 00mm},clip,width=1.0\linewidth]{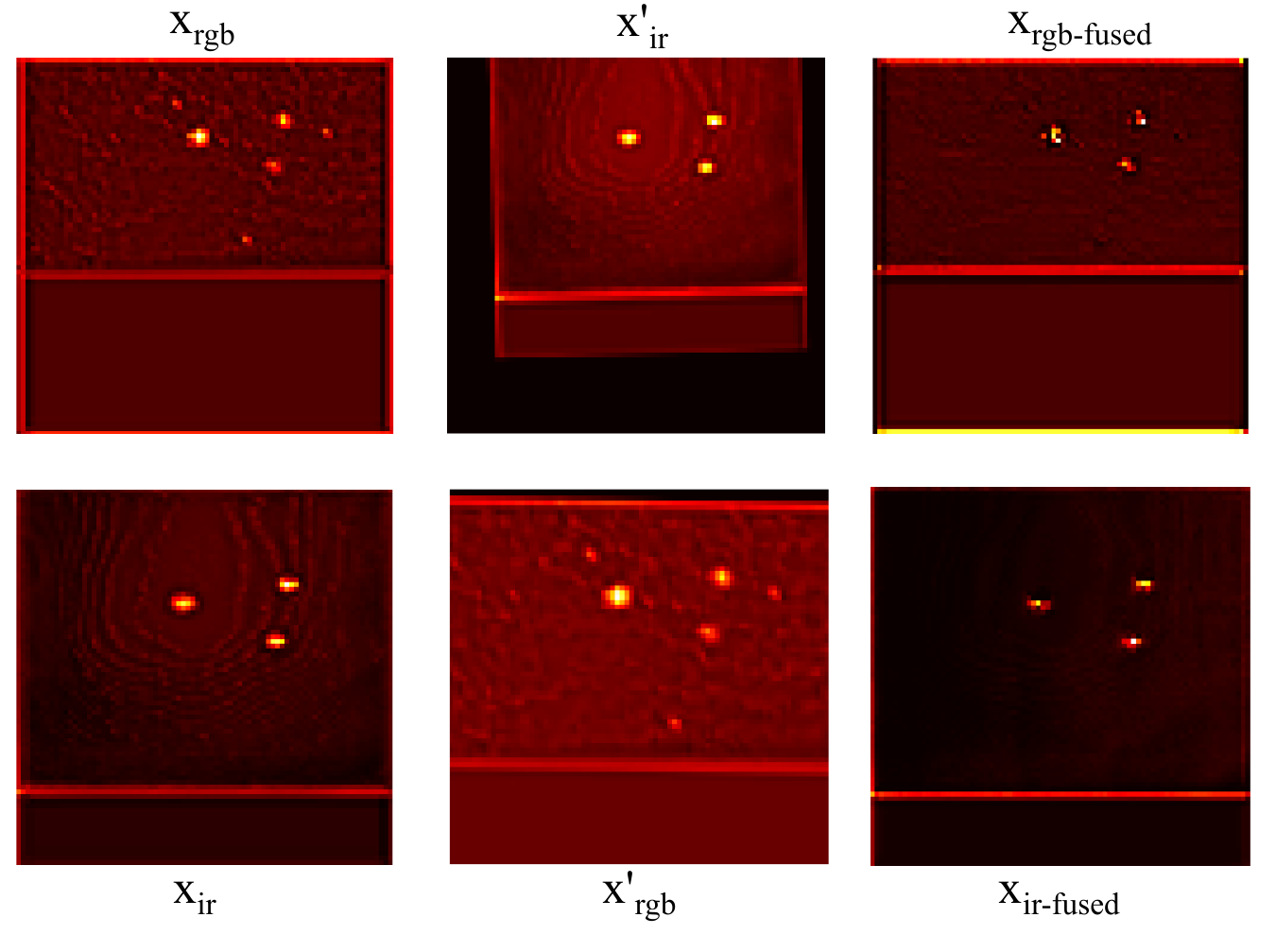}
    \caption{Visualisation of feature alignment and fusion. Inputs from two modalities, $\mathbf{X}_\text{rgb}$ and $\mathbf{X}_\text{ir}$, are first aligned to each other's coordinate space via the OGAA module, producing $\mathbf{X}'_\text{rgb}$ and $\mathbf{X}'_\text{ir}$. The aligned features are then combined with the original features and fed into the ADFM module, yielding modality-specific fused features $\mathbf{X}_\text{rgb-fused}$ and $\mathbf{X}_\text{ir-fused}$.}
    \label{fig:vis_features}
\end{figure}

To validate the effectiveness of different components in our framework, we conduct systematic ablation studies based on both the Deformable Conv-based alignment strategy and the STN-based alignment strategy. 
The results are reported in Table~\ref{tab:components-def} and~\ref{tab:components-stn}, respectively.
Specifically, the first row represents the baseline, where no cross-modal alignment or dynamic fusion is applied at the detection stage. 
Instead, unaligned features from both modalities are directly summed together, and the vanilla BoT-SORT tracker without a ReID module is used at the tracking stage. 
On this basis, the second row incorporates the alignment module.
The third row reports the impact of replacing a simple summation with the dynamic weighted fusion. 
Rows four to six focus on ablations at the tracking stage.
The fourth row augments BoT-SORT with an appearance-based ReID module to compensate for the limitations of IOU-only matching under complex UAV motion.
The fifth row adopts our proposed MMA-SORT with a three-stage association pipeline but uses only IOU similarity and appearance embeddings without incorporating the Event modality.
The sixth row presents the final version, where motion embeddings from the Event modality are introduced to highlight its potential for Multi-UAV tracking.

\begin{figure*}
    \centering
    \includegraphics[width=1.0\linewidth]{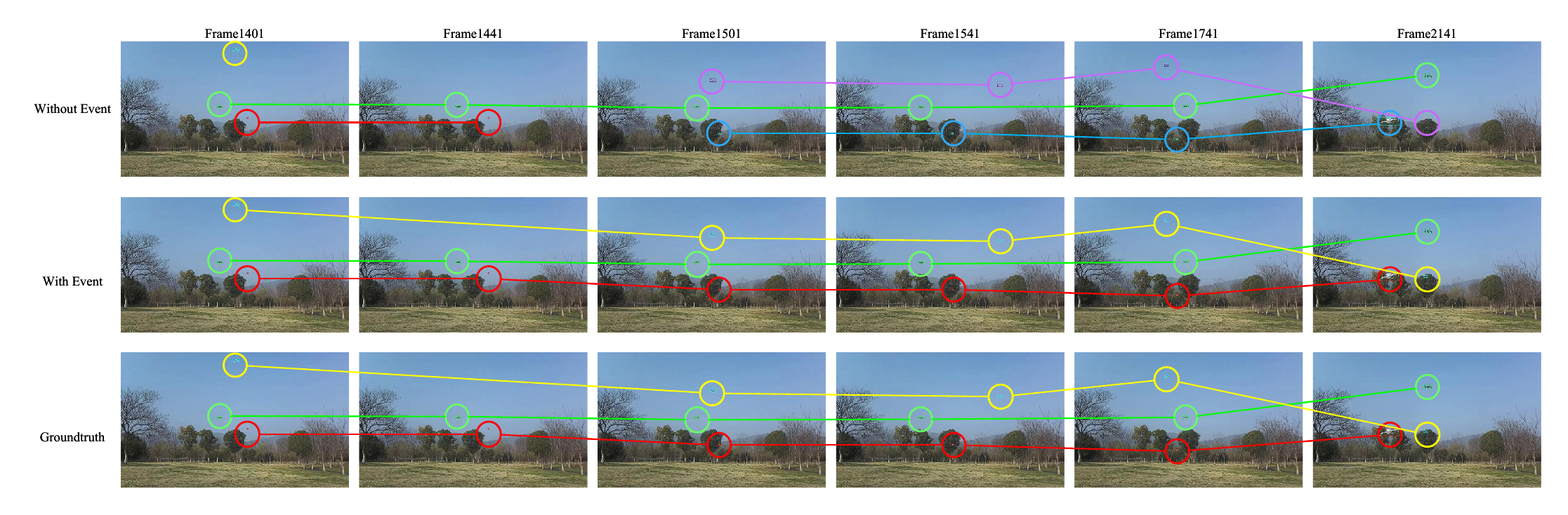}\\
    \includegraphics[width=1.0\linewidth]{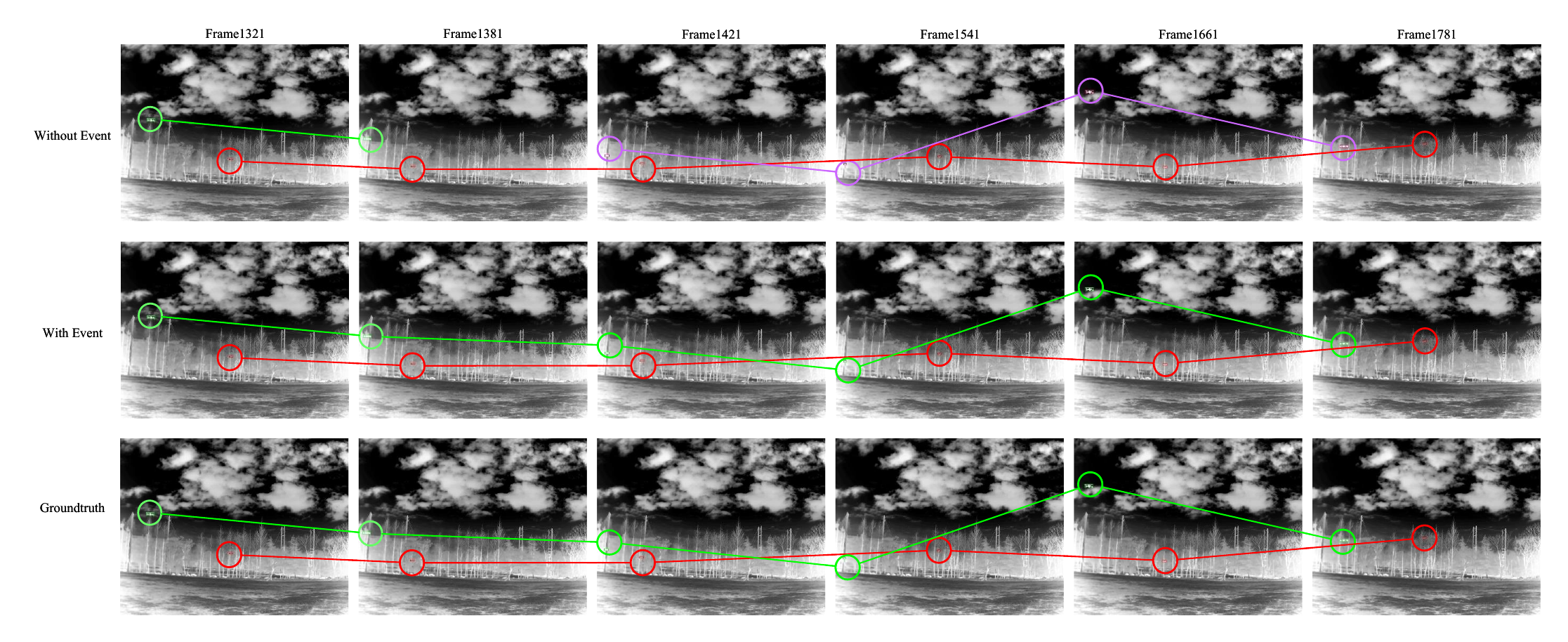}
    \caption{A visual comparison of the tracking performance w/o and w the Event modality, where trajectories with the same colour correspond to the same target ID.}
    \label{fig:vis_tracking_results}
\end{figure*}

\emph{Effect of Align \& Fusion Module:} The first three rows in Table~\ref{tab:components-def} and Table~\ref{tab:components-stn} demonstrate that incorporating the alignment module together with the adaptive dynamic fusion leads to consistent performance improvements, albeit at the cost of some runtime overhead. 
The alignment module aims to mitigate any spatial discrepancies between the modalities. 
In Table~\ref{tab:components-def}, introducing the Deformable Conv-based alignment yields noticeable gains, especially for the less robust RGB modality, where MOTA increases by 0.8\% and HOTA by 0.29\%. 
However, directly summing the modality features amplifies noise, producing more false positives and increasing ID switches. 
By contrast, the STN module performs global affine transformations to achieve alignment but lacks the local adaptive learning capability of Deformable Conv, making it more vulnerable to noise when features are directly summed. 
This explains the performance degradation observed in Table~\ref{tab:components-stn} when only the STN alignment is used. 
Notably, when the alignment module is coupled with the dynamic fusion (third rows in both tables), the two modalities are better balanced, resulting in significant performance improvements. 
Moreover, since STN explicitly applies affine transformations for alignment, we further visualise the feature variations, as shown in Fig.~\ref{fig:vis_features}.

\emph{The effect of the ReID Module:} Many existing trackers heavily rely on motion clues, using Kalman filter predictions and IOU-based bounding box similarity for association. 
However, such strategies are unreliable for UAV targets, whose trajectories are unstable and subject to frequent disappearances and reappearances.
Therefore, the ReID module is essential for tracking UAV targets. 
The fourth row in Table~\ref{tab:components-def} and Table~\ref{tab:components-def} confirms the effectiveness of this design, showing consistent improvements in UAV tracking accuracy.

\emph{Effect of Association Strategy:} An appropriate association strategy is crucial for multi-object tracking. 
Most existing trackers rely heavily on IOU similarity between bounding boxes, but the small size and irregular motion of UAV targets significantly reduce the reliability of IOU-based association. 
To overcome this, our proposed MMA-SORT weakens the dominance of IOU by (1) increasing the IOU filtering threshold, and (2) introducing motion embeddings captured from the Event modality to assist appearance-based ReID. 
The results in Table~\ref{tab:components-def} and Table~\ref{tab:components-stn} show that MMA-SORT provide substantial advantages in identity association. 
For instance, under the Deformable Conv-based alignment, for the RGB modality, HOTA improves from 45.44 to 56.07, IDF1 from 51.65 to 74.92, and ID switches are reduced from 1277 to 159. 
For the IR modality, HOTA increases from 53.50 to 66.57, IDF1 from 58.02 to 83.75, and ID switches from 1102 to 131. 
Similarly, under the STN-based alignment, for the RGB modality, HOTA rises from 44.40 to 56.43, IDF1 from 51.24 to 74.99, and ID switches decrease from 1018 to 151.
For the IR modality, HOTA improves from 53.81 to 66.81, IDF1 from 58.23 to 84.62, and ID switches from 1010 to 138.

\emph{The Effect of Motion Embedding:} Finally, we evaluate the merit of the Event modality in Multi-UAV tracking. 
As shown in the sixth row of Table~\ref{tab:components-def} and Table~\ref{tab:components-stn}, incorporating motion embeddings from the Event modality yields consistent overall improvements across all metrics. 
To illustrate its contribution, we present  qualitative visualisation of the tracking results in Fig.~\ref{fig:vis_tracking_results}, where the integration of the Event information leads to more accurate trajectory association.

\subsection{Limitations and Discussion} 
Despite the promising results achieved by the proposed multimodal tracking framework, several limitations remain to be addressed in future studies. 

\noindent \textbf{Limitations of modality alignment.} A key bottleneck of multimodal fusion lies in the risk of imperfect modality alignment. 
Although the proposed OGAA and ADFM modules alleviate cross-modal misalignment to a certain extent, they cannot totally eliminate discrepancies caused by sensor placement, resolution differences, or environmental variations. 
As a result, the complementary strengths of different modalities, particularly the Event modality, are not yet fully exploited. 
This limitation suggests that more advanced spatio-temporal alignment techniques are required to unlock the full potential of multi-modal multi-UAV tracking. 

\noindent \textbf{The challenges of cross-modal annotation for small UAVs.} Unlike existing multi-modal benchmarks, where object sizes are relatively consistent across modalities, UAV targets are extremely small and exhibit significant variations in apparent size between different modalities. 
This disparity makes the sharing of a unified cross-modal annotation unreliable. 
To mitigate this, we provide independent annotations for both RGB and IR frames and adopt a dual-stream network architecture that processes each modality separately. 
While this design ensures fairness and reliability in evaluation, it remains a compromise rather than a principled solution. 

\section{Conclusion}
This paper presents MM-UAV, the first large-scale multi-modal multi-UAV tracking benchmark, established to address the absence of an authoritative dataset in the field. 
MM-UAV encompasses three critical visual modalities—RGB, IR, and Event—and includes 1,321 video sequences (1,200 for training, 121 for testing), with over 2.8 million frames per modality. 
Beyond its unprecedented scale and diversity, the benchmark introduces rigorous annotation protocols that maintain consistent identity labels across occlusions and reappearances. 
Furthermore, it defines seven challenging scenario attributes, ensuring a faithful representation of real-world conditions and providing a reliable ground for evaluating algorithmic robustness.

Building upon this benchmark, we propose the first multi-modal multi-object tracking framework specifically designed for anti-UAV applications, enabling efficient cross-modal fusion and collaboration. 
Adhering to the Tracking-by-Detection paradigm, our framework introduces an Offset-Guided Adaptive Alignment Module (OGAA) and an Adaptive Dynamic Fusion Module (ADFM) at the detection stage. 
The OGAA leverages both deformable convolution and Spatial Transformer Network (STN) strategies to resolve spatio-temporal misalignments across different sensors, thereby achieving precise localisation of small UAV targets. 
The ADFM employs a channel-attention mechanism to dynamically balance the contributions of RGB and infrared features, fully exploiting their complementary strengths. 
At the tracking stage, we design the MMA-SORT tracker to mitigate identity switches caused by similar target appearances and irregular motion patterns. 
MMA-SORT is the first to incorporate motion embeddings derived from event data, which work in concert with appearance embeddings to enhance the accuracy of identity association and suppress false detections and matching errors in complex backgrounds very effectively.

Looking forward, we plan to investigate more advanced fusion strategies for the event modality and continuously refine the proposed multi-modal multi-UAV tracking framework to advance research in multi-modal UAV tracking.

\section*{Acknowledgement}
This work was supported in part by the National Natural Science Foundation of China (62020106012, 62576152), the Basic Research Program of Jiangsu (BK20250104), the Fundamental Research Funds for the Central Universities (JUSRP202504007), and by Leverhulme Trust Emeritus Fellowship EM-2025-06-09.


\bibliographystyle{ieeetr} 
\bibliography{myref.bib}    

@inproceedings{radar1,
  title={3d motion capture of an unmodified drone with single-chip millimeter wave radar},
  author={Zhao, Peijun and Lu, Chris Xiaoxuan and Wang, Bing and Trigoni, Niki and Markham, Andrew},
  booktitle={2021 IEEE International Conference on Robotics and Automation (ICRA)},
  pages={5186--5192},
  year={2021},
  organization={IEEE}
}

@article{RF,
  title={DroneRF dataset: A dataset of drones for RF-based detection, classification and identification},
  author={Allahham, MHD Saria and Al-Sa'd, Mohammad F and Al-Ali, Abdulla and Mohamed, Amr and Khattab, Tamer and Erbad, Aiman},
  journal={Data in brief},
  volume={26},
  pages={104313},
  year={2019}
}

@article{YOLOX,
  title={Yolox: Exceeding yolo series in 2021},
  author={Ge, Zheng and Liu, Songtao and Wang, Feng and Li, Zeming and Sun, Jian},
  journal={arXiv preprint arXiv:2107.08430},
  year={2021}
}

@article{yolov12,
  title={YOLOv12: Attention-Centric Real-Time Object Detectors},
  author={Tian, Yunjie and Ye, Qixiang and Doermann, David},
  journal={arXiv preprint arXiv:2502.12524},
  year={2025}
}

@inproceedings{detr,
  title={End-to-end object detection with transformers},
  author={Carion, Nicolas and Massa, Francisco and Synnaeve, Gabriel and Usunier, Nicolas and Kirillov, Alexander and Zagoruyko, Sergey},
  booktitle={European conference on computer vision},
  pages={213--229},
  year={2020},
  organization={Springer}
}

@inproceedings{DeepSORT,
  title={DeepSORT: real Time \& multi-object detection and tracking with YOLO and TensorFlow},
  author={Pujara, Abhijeet and Bhamare, Mamta},
  booktitle={2022 International conference on augmented intelligence and sustainable systems (ICAISS)},
  pages={456--460},
  year={2022},
  organization={IEEE}
}

@inproceedings{bytetrack,
  title={Bytetrack: Multi-object tracking by associating every detection box},
  author={Zhang, Yifu and Sun, Peize and Jiang, Yi and Yu, Dongdong and Weng, Fucheng and Yuan, Zehuan and Luo, Ping and Liu, Wenyu and Wang, Xinggang},
  booktitle={European conference on computer vision},
  pages={1--21},
  year={2022},
  organization={Springer}
}

@article{botsort,
  title={BoT-SORT: Robust associations multi-pedestrian tracking},
  author={Aharon, Nir and Orfaig, Roy and Bobrovsky, Ben-Zion},
  journal={arXiv preprint arXiv:2206.14651},
  year={2022}
}

@inproceedings{trackformer,
  title={Trackformer: Multi-object tracking with transformers},
  author={Meinhardt, Tim and Kirillov, Alexander and Leal-Taixe, Laura and Feichtenhofer, Christoph},
  booktitle={Proceedings of the IEEE/CVF conference on computer vision and pattern recognition},
  pages={8844--8854},
  year={2022}
}

@inproceedings{SiamSTA,
  title={Siamsta: Spatio-temporal attention based siamese tracker for tracking uavs},
  author={Huang, Bo and Chen, Junjie and Xu, Tingfa and Wang, Ying and Jiang, Shenwang and Wang, Yuncheng and Wang, Lei and Li, Jianan},
  booktitle={Proceedings of the IEEE/CVF international conference on computer vision},
  pages={1204--1212},
  year={2021}
}

@inproceedings{GLTF-MA,
  title={A global-local tracking framework driven by both motion and appearance for infrared anti-uav},
  author={Li, Yifan and Yuan, Dian and Sun, Meng and Wang, Hongyu and Liu, Xiaotao and Liu, Jing},
  booktitle={Proceedings of the IEEE/CVF Conference on Computer Vision and Pattern Recognition},
  pages={3026--3035},
  year={2023}
}

@inproceedings{UTTracker,
  title={A unified transformer based tracker for anti-uav tracking},
  author={Yu, Qianjin and Ma, Yinchao and He, Jianfeng and Yang, Dawei and Zhang, Tianzhu},
  booktitle={Proceedings of the IEEE/CVF Conference on Computer Vision and Pattern Recognition},
  pages={3036--3046},
  year={2023}
}

@misc{2th-antiuav,
      title={The 2nd Anti-UAV Workshop \& Challenge: Methods and Results}, 
      author={Jian Zhao and Gang Wang and Jianan Li and Lei Jin and Nana Fan and Min Wang and Xiaojuan Wang and Ting Yong and Yafeng Deng and Yandong Guo and Shiming Ge and Guodong Guo},
      year={2021},
      eprint={2108.09909},
      archivePrefix={arXiv},
      primaryClass={cs.CV},
      url={https://arxiv.org/abs/2108.09909}, 
}

@misc{3th-antiuav,
      title={The 3rd Anti-UAV Workshop \& Challenge: Methods and Results}, 
      author={Jian Zhao and Jianan Li and Lei Jin and Jiaming Chu and Zhihao Zhang and Jun Wang and Jiangqiang Xia and Kai Wang and Yang Liu and Sadaf Gulshad and Jiaojiao Zhao and Tianyang Xu and Xuefeng Zhu and Shihan Liu and Zheng Zhu and Guibo Zhu and Zechao Li and Zheng Wang and Baigui Sun and Yandong Guo and Shin ichi Satoh and Junliang Xing and Jane Shen Shengmei},
      year={2023},
      eprint={2305.07290},
      archivePrefix={arXiv},
      primaryClass={cs.CV},
      url={https://arxiv.org/abs/2305.07290}, 
}

@inproceedings{Dist-Tracker,
  title={Dist-Tracker: A Small Object-aware Detector and Tracker for UAV Tracking},
  author={Wang, Wenzhen and Fu, Jing and Song, Jiayi and Li, Kaiyu and Qiao, Hui and Liu, Jiang and Sun, Hao and Cao, Xiangyong},
  booktitle={Proceedings of the Computer Vision and Pattern Recognition Conference},
  pages={6601--6609},
  year={2025}
}

@misc{strong-baseline,
      title={Strong Baseline: Multi-UAV Tracking via YOLOv12 with BoT-SORT-ReID}, 
      author={Yu-Hsi Chen},
      year={2025},
      eprint={2503.17237},
      archivePrefix={arXiv},
      primaryClass={cs.CV},
      url={https://arxiv.org/abs/2503.17237}, 
}

@article{FL-Drones,
  title={Detecting flying objects using a single moving camera},
  author={Rozantsev, Artem and Lepetit, Vincent and Fua, Pascal},
  journal={IEEE transactions on pattern analysis and machine intelligence},
  volume={39},
  number={5},
  pages={879--892},
  year={2016},
  publisher={IEEE}
}

@inproceedings{NPS-Drones,
  title={Multi-target detection and tracking from a single camera in Unmanned Aerial Vehicles (UAVs)},
  author={Li, Jing and Ye, Dong Hye and Chung, Timothy and Kolsch, Mathias and Wachs, Juan and Bouman, Charles},
  booktitle={2016 IEEE/RSJ international conference on intelligent robots and systems (IROS)},
  pages={4992--4997},
  year={2016},
  organization={IEEE}
}

@article{Det-Fly,
  title={Air-to-air visual detection of micro-uavs: An experimental evaluation of deep learning},
  author={Zheng, Ye and Chen, Zhang and Lv, Dailin and Li, Zhixing and Lan, Zhenzhong and Zhao, Shiyu},
  journal={IEEE Robotics and automation letters},
  volume={6},
  number={2},
  pages={1020--1027},
  year={2021},
  publisher={IEEE}
}

@inproceedings{Drone-vs-Bird,
  title={Drone-vs-bird detection challenge at IEEE AVSS2021},
  author={Coluccia, Angelo and Fascista, Alessio and Schumann, Arne and Sommer, Lars and Dimou, Anastasios and Zarpalas, Dimitrios and Akyon, Fatih Cagatay and Eryuksel, Ogulcan and Ozfuttu, Kamil Anil and Altinuc, Sinan Onur and others},
  booktitle={2021 17th IEEE International Conference on Advanced Video and Signal Based Surveillance (AVSS)},
  pages={1--8},
  year={2021},
  organization={IEEE}
}

@article{UETT4K-Anti-UAV,
  title={UETT4K Anti-UAV: A Large Scale 4K Benchmark Dataset for Vision-Based Drone Detection in High-Resolution Imagery},
  author={Awan, Mughees Sarwar and Zaidi, Syed Azhar Ali and Mir, Junaid},
  journal={IEEE Access},
  year={2025},
  volume={13},
  pages={73553-73568},
  publisher={IEEE}
}

@inproceedings{USC-Drone,
  title={A deep learning approach to drone monitoring},
  author={Chen, Yueru and Aggarwal, Pranav and Choi, Jongmoo and Kuo, C-C Jay},
  booktitle={2017 Asia-Pacific Signal and Information Processing Association Annual Summit and Conference (APSIPA ASC)},
  pages={686--691},
  year={2017},
  organization={IEEE}
}

@article{Anti-UAV,
  title={Anti-UAV: A large-scale benchmark for vision-based UAV tracking},
  author={Jiang, Nan and Wang, Kuiran and Peng, Xiaoke and Yu, Xuehui and Wang, Qiang and Xing, Junliang and Li, Guorong and Guo, Guodong and Ye, Qixiang and Jiao, Jianbin and others},
  journal={IEEE Transactions on Multimedia},
  volume={25},
  pages={486--500},
  year={2021},
  publisher={IEEE}
}

@article{Anti-UAV600,
  author       = {Xuefeng Zhu and
                  Tianyang Xu and
                  Jian Zhao and
                  Jia{-}Wei Liu and
                  Kai Wang and
                  Gang Wang and
                  Jianan Li and
                  Zhihao Zhang and
                  Qiang Wang and
                  Lei Jin and
                  Zheng Zhu and
                  Junliang Xing and
                  Xiao{-}Jun Wu},
  title        = {Evidential Detection and Tracking Collaboration: New Problem, Benchmark
                  and Algorithm for Robust Anti-UAV System},
  journal      = {CoRR},
  volume       = {abs/2306.15767},
  year         = {2023},
  url          = {https://doi.org/10.48550/arXiv.2306.15767},
  doi          = {10.48550/ARXIV.2306.15767},
  eprinttype    = {arXiv},
  eprint       = {2306.15767},
  bibsource    = {dblp computer science bibliography, https://dblp.org}
}

@article{Anti-UAV410,
  title={Anti-UAV410: A thermal infrared benchmark and customized scheme for tracking drones in the wild},
  author={Huang, Bo and Li, Jianan and Chen, Junjie and Wang, Gang and Zhao, Jian and Xu, Tingfa},
  journal={IEEE Transactions on Pattern Analysis and Machine Intelligence},
  volume={46},
  number={5},
  pages={2852--2865},
  year={2023},
  publisher={IEEE}
}

@inproceedings{Halmstad-Drone,
  title={Real-time drone detection and tracking with visible, thermal and acoustic sensors},
  author={Svanstr{\"o}m, Fredrik and Englund, Cristofer and Alonso-Fernandez, Fernando},
  booktitle={2020 25th International Conference on Pattern Recognition (ICPR)},
  pages={7265--7272},
  year={2021},
  organization={IEEE}
}

@ARTICLE{DUT-Anti-UAV,
  author={Zhao, Jie and Zhang, Jingshu and Li, Dongdong and Wang, Dong},
  journal={IEEE Transactions on Intelligent Transportation Systems}, 
  title={Vision-Based Anti-UAV Detection and Tracking}, 
  year={2022},
  volume={23},
  number={12},
  pages={25323-25334},
  doi={10.1109/TITS.2022.3177627}}

@INPROCEEDINGS{MOT-FLY,
  author={Chu, Zhaochen and Song, Tao and Jin, Ren and Jiang, Tao},
  booktitle={2023 IEEE International Conference on Unmanned Systems (ICUS)}, 
  title={An Experimental Evaluation Based on New Air-to-Air Multi-UAV Tracking Dataset}, 
  year={2023},
  volume={},
  number={},
  pages={671-676},
  doi={10.1109/ICUS58632.2023.10318461}}

@article{UAVSwarm,
  title={UAVSwarm dataset: An unmanned aerial vehicle swarm dataset for multiple object tracking},
  author={Wang, Chuanyun and Su, Yang and Wang, Jingjing and Wang, Tian and Gao, Qian},
  journal={Remote Sensing},
  volume={14},
  number={11},
  pages={2601},
  year={2022},
  publisher={MDPI}
}

@misc{4th-Anti-UAV-workshop,
  title        = {4th Anti-UAV Challenge},
  year         = {2025},
  note         = {https://anti-uav.github.io/},
  urldate      = {2025-6}
}

@inproceedings{sort,
  title={Simple online and realtime tracking with a deep association metric},
  author={Wojke, Nicolai and Bewley, Alex and Paulus, Dietrich},
  booktitle={2017 IEEE international conference on image processing (ICIP)},
  pages={3645--3649},
  year={2017},
  organization={IEEE}
}

@inproceedings{ocsort,
  title={Observation-centric sort: Rethinking sort for robust multi-object tracking},
  author={Cao, Jinkun and Pang, Jiangmiao and Weng, Xinshuo and Khirodkar, Rawal and Kitani, Kris},
  booktitle={Proceedings of the IEEE/CVF conference on computer vision and pattern recognition},
  pages={9686--9696},
  year={2023}
}

@inproceedings{deep-ocsort,
  title={Deep oc-sort: Multi-pedestrian tracking by adaptive re-identification},
  author={Maggiolino, Gerard and Ahmad, Adnan and Cao, Jinkun and Kitani, Kris},
  booktitle={2023 IEEE International conference on image processing (ICIP)},
  pages={3025--3029},
  year={2023},
  organization={IEEE}
}

@article{boostTrack,
  title={BoostTrack: Boosting the similarity measure and detection confidence for improved multiple object tracking},
  author={Stanojevic, Vukasin D and Todorovic, Branimir T},
  journal={Machine Vision and Applications},
  volume={35},
  number={3},
  pages={53},
  year={2024},
  publisher={Springer}
}

@article{boostTrack++,
  title={Boosttrack++: using tracklet information to detect more objects in multiple object tracking},
  author={Stanojevi{\'c}, Vuka{\v{s}}in and Todorovi{\'c}, Branimir},
  journal={arXiv preprint arXiv:2408.13003},
  year={2024}
}

@misc{transtrack,
      title={TransTrack: Multiple Object Tracking with Transformer}, 
      author={Peize Sun and Jinkun Cao and Yi Jiang and Rufeng Zhang and Enze Xie and Zehuan Yuan and Changhu Wang and Ping Luo},
      year={2021},
      eprint={2012.15460},
      archivePrefix={arXiv},
      primaryClass={cs.CV},
      url={https://arxiv.org/abs/2012.15460}, 
}

@inproceedings{motr,
  title={Motr: End-to-end multiple-object tracking with transformer},
  author={Zeng, Fangao and Dong, Bin and Zhang, Yuang and Wang, Tiancai and Zhang, Xiangyu and Wei, Yichen},
  booktitle={European conference on computer vision},
  pages={659--675},
  year={2022},
  organization={Springer}
}

@inproceedings{motrv2,
  title={Motrv2: Bootstrapping end-to-end multi-object tracking by pretrained object detectors},
  author={Zhang, Yuang and Wang, Tiancai and Zhang, Xiangyu},
  booktitle={Proceedings of the IEEE/CVF conference on computer vision and pattern recognition},
  pages={22056--22065},
  year={2023}
}

@inproceedings{memotr,
  title={MeMOTR: Long-term memory-augmented transformer for multi-object tracking},
  author={Gao, Ruopeng and Wang, Limin},
  booktitle={Proceedings of the IEEE/CVF International Conference on Computer Vision},
  pages={9901--9910},
  year={2023}
}

@inproceedings{motip,
  title={Multiple object tracking as id prediction},
  author={Gao, Ruopeng and Qi, Ji and Wang, Limin},
  booktitle={Proceedings of the Computer Vision and Pattern Recognition Conference},
  pages={27883--27893},
  year={2025}
}

@article{KF,
  title={A New Approach To Linear Filtering and Prediction Problems},
  author={ Kalman, R. E. },
  journal={Journal of Basic Engineering},
  volume={82D},
  pages={35-45},
  year={1960},
}

@misc{mot15,
      title={MOTChallenge 2015: Towards a Benchmark for Multi-Target Tracking}, 
      author={Laura Leal-Taixé and Anton Milan and Ian Reid and Stefan Roth and Konrad Schindler},
      year={2015},
      eprint={1504.01942},
      archivePrefix={arXiv},
      primaryClass={cs.CV},
      url={https://arxiv.org/abs/1504.01942}, 
}

@misc{mot20,
      title={MOT20: A benchmark for multi object tracking in crowded scenes}, 
      author={Patrick Dendorfer and Hamid Rezatofighi and Anton Milan and Javen Shi and Daniel Cremers and Ian Reid and Stefan Roth and Konrad Schindler and Laura Leal-Taixé},
      year={2020},
      eprint={2003.09003},
      archivePrefix={arXiv},
      primaryClass={cs.CV},
      url={https://arxiv.org/abs/2003.09003}, 
}

@ARTICLE{visevent,
  author={Wang, Xiao and Li, Jianing and Zhu, Lin and Zhang, Zhipeng and Chen, Zhe and Li, Xin and Wang, Yaowei and Tian, Yonghong and Wu, Feng},
  journal={IEEE Transactions on Cybernetics}, 
  title={VisEvent: Reliable Object Tracking via Collaboration of Frame and Event Flows}, 
  year={2024},
  volume={54},
  number={3},
  pages={1997-2010},
  doi={10.1109/TCYB.2023.3318601}}

@misc{ceutrack,
      title={Revisiting Color-Event based Tracking: A Unified Network, Dataset, and Metric}, 
      author={Chuanming Tang and Xiao Wang and Ju Huang and Bo Jiang and Lin Zhu and Jianlin Zhang and Yaowei Wang and Yonghong Tian},
      year={2024},
      eprint={2211.11010},
      archivePrefix={arXiv},
      primaryClass={cs.CV},
      url={https://arxiv.org/abs/2211.11010}, 
}

@INPROCEEDINGS{eventvot,
  author={Wang, Xiao and Wang, Shiao and Tang, Chuanming and Zhu, Lin and Jiang, Bo and Tian, Yonghong and Tang, Jin},
  booktitle={2024 IEEE/CVF Conference on Computer Vision and Pattern Recognition (CVPR)}, 
  title={Event Stream-Based Visual Object Tracking: A High-Resolution Benchmark Dataset and A Novel Baseline}, 
  year={2024},
  volume={},
  number={},
  pages={19248-19257},
  doi={10.1109/CVPR52733.2024.01821}}

@InProceedings{fe108,
    author    = {Zhang, Jiqing and Yang, Xin and Fu, Yingkai and Wei, Xiaopeng and Yin, Baocai and Dong, Bo},
    title     = {Object Tracking by Jointly Exploiting Frame and Event Domain},
    booktitle = {Proceedings of the IEEE/CVF International Conference on Computer Vision (ICCV)},
    month     = {October},
    year      = {2021},
    pages     = {13043-13052}
}

@article{TENet,
title = {TENet: Targetness entanglement incorporating with multi-scale pooling and mutually-guided fusion for RGB-E object tracking},
journal = {Neural Networks},
volume = {183},
pages = {106948},
year = {2025},
issn = {0893-6080},
doi = {https://doi.org/10.1016/j.neunet.2024.106948},
url = {https://www.sciencedirect.com/science/article/pii/S0893608024008773},
author = {Pengcheng Shao and Tianyang Xu and Zhangyong Tang and Linze Li and Xiao-Jun Wu and Josef Kittler}
}

@INPROCEEDINGS{vipt,
  author={Zhu, Jiawen and Lai, Simiao and Chen, Xin and Wang, Dong and Lu, Huchuan},
  booktitle={2023 IEEE/CVF Conference on Computer Vision and Pattern Recognition (CVPR)}, 
  title={Visual Prompt Multi-Modal Tracking}, 
  year={2023},
  volume={},
  number={},
  pages={9516-9526},
  doi={10.1109/CVPR52729.2023.00918}}

@inproceedings{afnet,
  title={Frame-event alignment and fusion network for high frame rate tracking},
  author={Zhang, Jiqing and Wang, Yuanchen and Liu, Wenxi and Li, Meng and Bai, Jinpeng and Yin, Baocai and Yang, Xin},
  booktitle={Proceedings of the IEEE/CVF conference on computer vision and pattern recognition},
  pages={9781--9790},
  year={2023}
}

@INPROCEEDINGS{CSOM,
  author={Chen, Chen and Qi, Jiahao and Liu, Xingyue and Bin, Kangcheng and Fu, Ruigang and Hu, Xikun and Zhong, Ping},
  booktitle={2024 IEEE/CVF Conference on Computer Vision and Pattern Recognition (CVPR)}, 
  title={Weakly Misalignment-Free Adaptive Feature Alignment for UAVs-Based Multimodal Object Detection}, 
  year={2024},
  volume={},
  number={},
  pages={26826-26835},
  doi={10.1109/CVPR52733.2024.02534}}

@INPROCEEDINGS{ARCNN,
  author={Zhang, Lu and Zhu, Xiangyu and Chen, Xiangyu and Yang, Xu and Lei, Zhen and Liu, Zhiyong},
  booktitle={2019 IEEE/CVF International Conference on Computer Vision (ICCV)}, 
  title={Weakly Aligned Cross-Modal Learning for Multispectral Pedestrian Detection}, 
  year={2019},
  volume={},
  number={},
  pages={5126-5136},
  doi={10.1109/ICCV.2019.00523}}

@INPROCEEDINGS{align2,
  author={Chen, Yongxin and Guan, Yong and Shao, Zhenzhou},
  booktitle={2023 IEEE International Conference on Real-time Computing and Robotics (RCAR)}, 
  title={Real-Time Multispectral Pedestrian Detection with Weakly Aligned Cross-Modal Learning}, 
  year={2023},
  volume={},
  number={},
  pages={829-834},
  doi={10.1109/RCAR58764.2023.10249327}}

@INPROCEEDINGS{def-conv,
  author={Dai, Jifeng and Qi, Haozhi and Xiong, Yuwen and Li, Yi and Zhang, Guodong and Hu, Han and Wei, Yichen},
  booktitle={2017 IEEE International Conference on Computer Vision (ICCV)}, 
  title={Deformable Convolutional Networks}, 
  year={2017},
  volume={},
  number={},
  pages={764-773},
  doi={10.1109/ICCV.2017.89}}

@misc{STN,
      title={Spatial Transformer Networks}, 
      author={Max Jaderberg and Karen Simonyan and Andrew Zisserman and Koray Kavukcuoglu},
      year={2016},
      eprint={1506.02025},
      archivePrefix={arXiv},
      primaryClass={cs.CV},
      url={https://arxiv.org/abs/1506.02025}, 
}

@INPROCEEDINGS{siamfc,
  author={Cen, Miaobin and Jung, Cheolkon},
  booktitle={2018 25th IEEE International Conference on Image Processing (ICIP)}, 
  title={Fully Convolutional Siamese Fusion Networks for Object Tracking}, 
  year={2018},
  volume={},
  number={},
  pages={3718-3722},
  doi={10.1109/ICIP.2018.8451102}}

@INPROCEEDINGS{siamRPN,
  author={Li, Bo and Yan, Junjie and Wu, Wei and Zhu, Zheng and Hu, Xiaolin},
  booktitle={2018 IEEE/CVF Conference on Computer Vision and Pattern Recognition}, 
  title={High Performance Visual Tracking with Siamese Region Proposal Network}, 
  year={2018},
  volume={},
  number={},
  pages={8971-8980},
  doi={10.1109/CVPR.2018.00935}}

@INPROCEEDINGS{siamRPN++,
  author={Li, Bo and Wu, Wei and Wang, Qiang and Zhang, Fangyi and Xing, Junliang and Yan, Junjie},
  booktitle={2019 IEEE/CVF Conference on Computer Vision and Pattern Recognition (CVPR)}, 
  title={SiamRPN++: Evolution of Siamese Visual Tracking With Very Deep Networks}, 
  year={2019},
  volume={},
  number={},
  pages={4277-4286},
  doi={10.1109/CVPR.2019.00441}}

@INPROCEEDINGS {stark,
author = { Yan, Bin and Peng, Houwen and Fu, Jianlong and Wang, Dong and Lu, Huchuan },
booktitle = { 2021 IEEE/CVF International Conference on Computer Vision (ICCV) },
title = {{ Learning Spatio-Temporal Transformer for Visual Tracking }},
year = {2021},
volume = {},
ISSN = {},
pages = {10428-10437},
doi = {10.1109/ICCV48922.2021.01028},
url = {https://doi.ieeecomputersociety.org/10.1109/ICCV48922.2021.01028},
publisher = {IEEE Computer Society},
address = {Los Alamitos, CA, USA},
month =Oct}

@inproceedings{ostrack,
  title={Joint feature learning and relation modeling for tracking: A one-stream framework},
  author={Ye, Botao and Chang, Hong and Ma, Bingpeng and Shan, Shiguang and Chen, Xilin},
  booktitle={European conference on computer vision},
  pages={341--357},
  year={2022},
  organization={Springer}
}

@INPROCEEDINGS{mixformer,
  author={Cui, Yutao and Jiang, Cheng and Wang, Limin and Wu, Gangshan},
  booktitle={2022 IEEE/CVF Conference on Computer Vision and Pattern Recognition (CVPR)}, 
  title={MixFormer: End-to-End Tracking with Iterative Mixed Attention}, 
  year={2022},
  volume={},
  number={},
  pages={13598-13608},
  doi={10.1109/CVPR52688.2022.01324}}

@ARTICLE{got10k,
  author={Huang, Lianghua and Zhao, Xin and Huang, Kaiqi},
  journal={IEEE Transactions on Pattern Analysis and Machine Intelligence}, 
  title={GOT-10k: A Large High-Diversity Benchmark for Generic Object Tracking in the Wild}, 
  year={2021},
  volume={43},
  number={5},
  pages={1562-1577},
  doi={10.1109/TPAMI.2019.2957464}}

@inproceedings{trackingnet,
  title={Trackingnet: A large-scale dataset and benchmark for object tracking in the wild},
  author={Muller, Matthias and Bibi, Adel and Giancola, Silvio and Alsubaihi, Salman and Ghanem, Bernard},
  booktitle={European Conference on Computer Vision},
  pages={300--317},
  year={2018}
}

@inproceedings{lasot,
  title={Lasot: A high-quality benchmark for large-scale single object tracking},
  author={Fan, Heng and Lin, Liting and Yang, Fan and Chu, Peng and Deng, Ge and Yu, Sijia and Bai, Hexin and Xu, Yong and Liao, Chunyuan and Ling, Haibin},
  booktitle={IEEE/CVF Conference on Computer Vision and Pattern Recognition},
  pages={5374--5383},
  year={2019}
}

@article{hota,
  title={Hota: A higher order metric for evaluating multi-object tracking},
  author={Luiten, Jonathon and Osep, Aljosa and Dendorfer, Patrick and Torr, Philip and Geiger, Andreas and Leal-Taix{\'e}, Laura and Leibe, Bastian},
  journal={International Journal of Computer Vision},
  volume={129},
  number={2},
  pages={548--578},
  year={2021},
  publisher={Springer}
}

@article{mota,
  title={Evaluating multiple object tracking performance: the clear mot metrics},
  author={Bernardin, Keni and Stiefelhagen, Rainer},
  journal={EURASIP Journal on Image and Video Processing},
  volume={2008},
  number={1},
  pages={246309},
  year={2008},
  publisher={Springer}
}

@inproceedings{fan2020visdrone,
  title={VisDrone-SOT2020: The vision meets drone single object tracking challenge results},
  author={Fan, Heng and Wen, Longyin and Du, Dawei and Zhu, Pengfei and Hu, Qinghua and Ling, Haibin and Shah, Mubarak and Wang, Biao and Dong, Bin and Yuan, Di and others},
  booktitle={European Conference on Computer Vision},
  pages={728--749},
  year={2020},
  organization={Springer}
}

@inproceedings{cheng2025one,
  title={One model for all: Low-level task interaction is a key to task-agnostic image fusion},
  author={Cheng, Chunyang and Xu, Tianyang and Feng, Zhenhua and Wu, Xiaojun and Tang, Zhangyong and Li, Hui and Zhang, Zeyang and Atito, Sara and Awais, Muhammad and Kittler, Josef},
  booktitle={Proceedings of the IEEE/CVF Conference on Computer Vision and Pattern Recognition},
  pages={28102--28112},
  year={2025}
}

@inproceedings{kristan2023first,
  title={The first visual object tracking segmentation vots2023 challenge results},
  author={Kristan, Matej and Matas, Ji{\v{r}}{\'\i} and Danelljan, Martin and Felsberg, Michael and Chang, Hyung Jin and Zajc, Luka {\v{C}}ehovin and Luke{\v{z}}i{\v{c}}, Alan and Drbohlav, Ondrej and Zhang, Zhongqun and Tran, Khanh-Tung and others},
  booktitle={Proceedings of the IEEE/CVF international conference on computer vision},
  pages={1796--1818},
  year={2023}
}

@article{xu2020accelerated,
  title={An accelerated correlation filter tracker},
  author={Xu, Tianyang and Feng, Zhen-Hua and Wu, Xiao-Jun and Kittler, Josef},
  journal={Pattern recognition},
  volume={102},
  pages={107172},
  year={2020},
  publisher={Elsevier}
}




\begin{IEEEbiography}[{\includegraphics[width=1in,height=1.25in,clip,keepaspectratio]{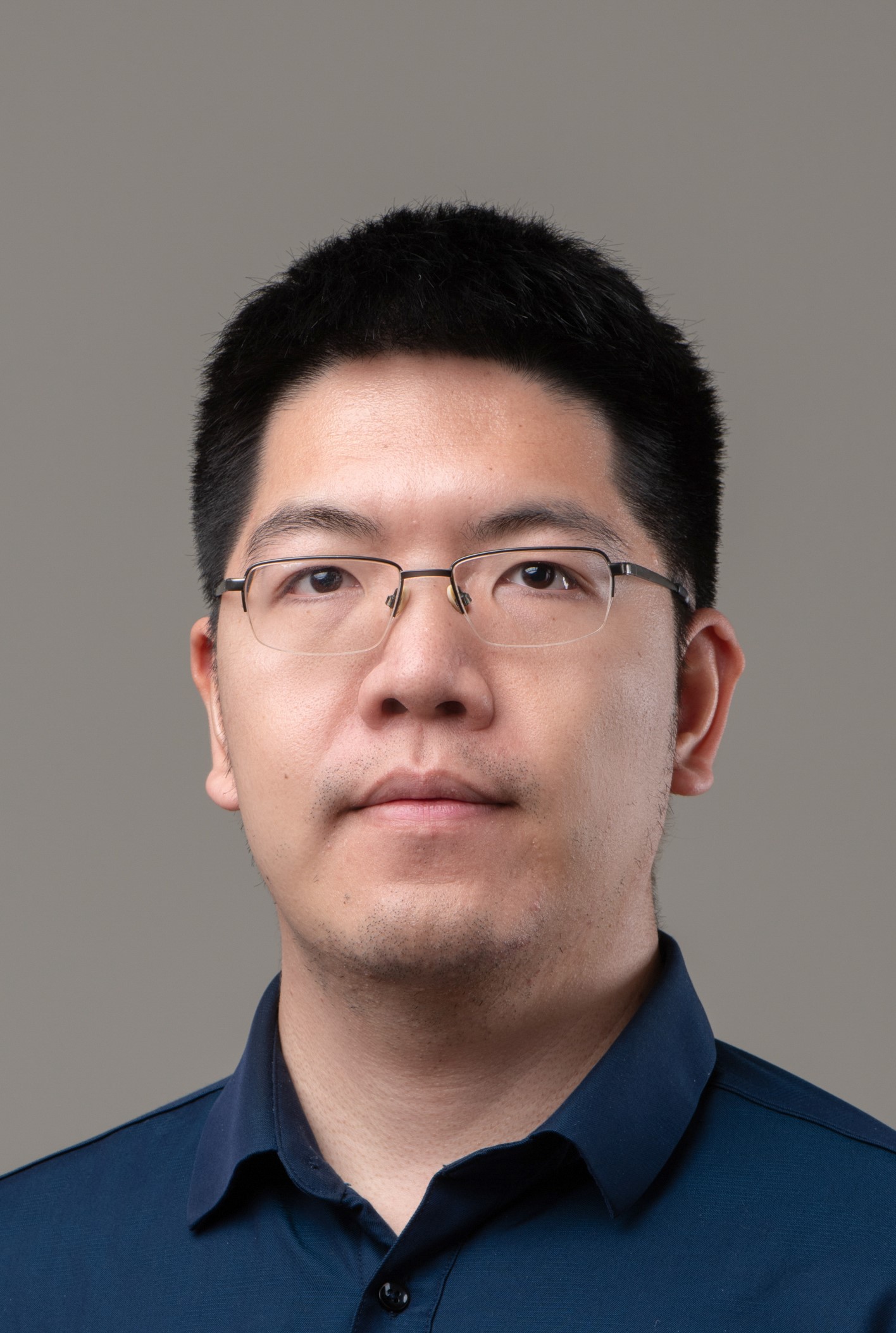}}]{Tianyang Xu}
(Member, IEEE) 
received the B.Sc. degree in electronic
science and engineering from Nanjing University,
Nanjing, China, in 2011. He received his PhD degree
at the School of Artificial Intelligence and Computer Science, Jiangnan University, Wuxi, China, in
2019. He was a Research Fellow in the Centre for Vision, Speech and Signal Processing (CVSSP), University of Surrey.
He is currently an Associate Professor at the
School of Artificial Intelligence and Computer Science, Jiangnan University, Wuxi, China. 
His research
interests include visual understanding and multi-modal learning.
He has published several scientific papers, including
IEEE TPAMI, IJCV, IEEE TIP, CVPR, ICCV, ICML, NeurIPS, etc. 
His publications have been cited more than 6000 times. 
He achieved top 1 performance in several competitions, including the
VOT2020 RGBT challenge, Anti-UAV challenge, and
MMVRAC challenges.
\end{IEEEbiography}

\begin{IEEEbiography}[{\includegraphics[width=1in,height=1.25in,clip,keepaspectratio]{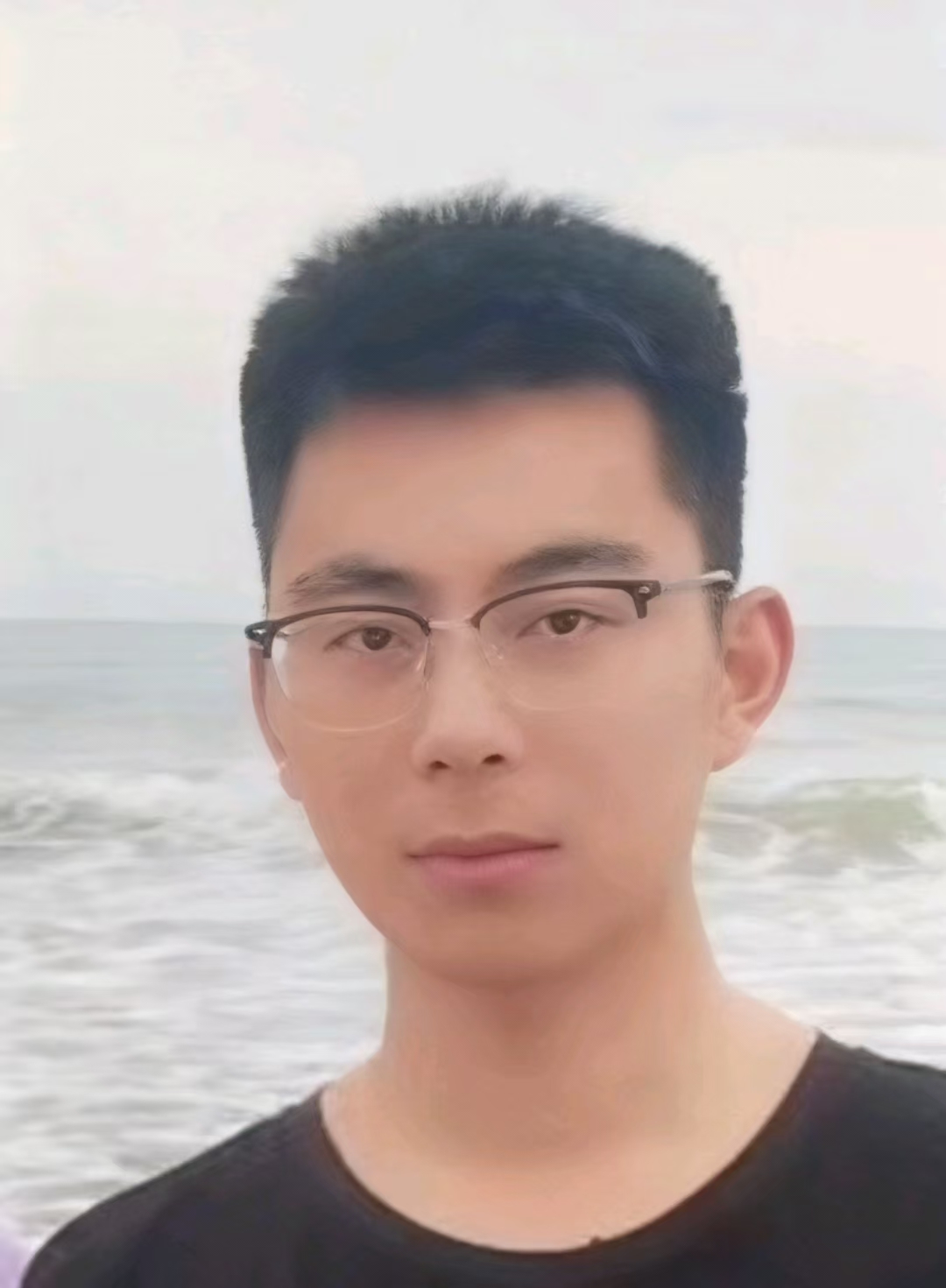}}]{Jinjie Gu} is currently pursuing his Master’s degree at Jiangnan University. His research focuses on object detection and multi-modal visual object tracking, including single-object tracking, multi-object tracking, and anti-UAV tracking. He served as one of the organisers of the 4th Anti-UAV Workshop at CVPR2025. He participated in several challenges on countering maneuvering targets, achieving top-two rankings in the multi-object tracking track of the PETS2025 Challenge.
\end{IEEEbiography}

\begin{IEEEbiography}[{\includegraphics[width=1in,height=1.25in,clip,keepaspectratio]{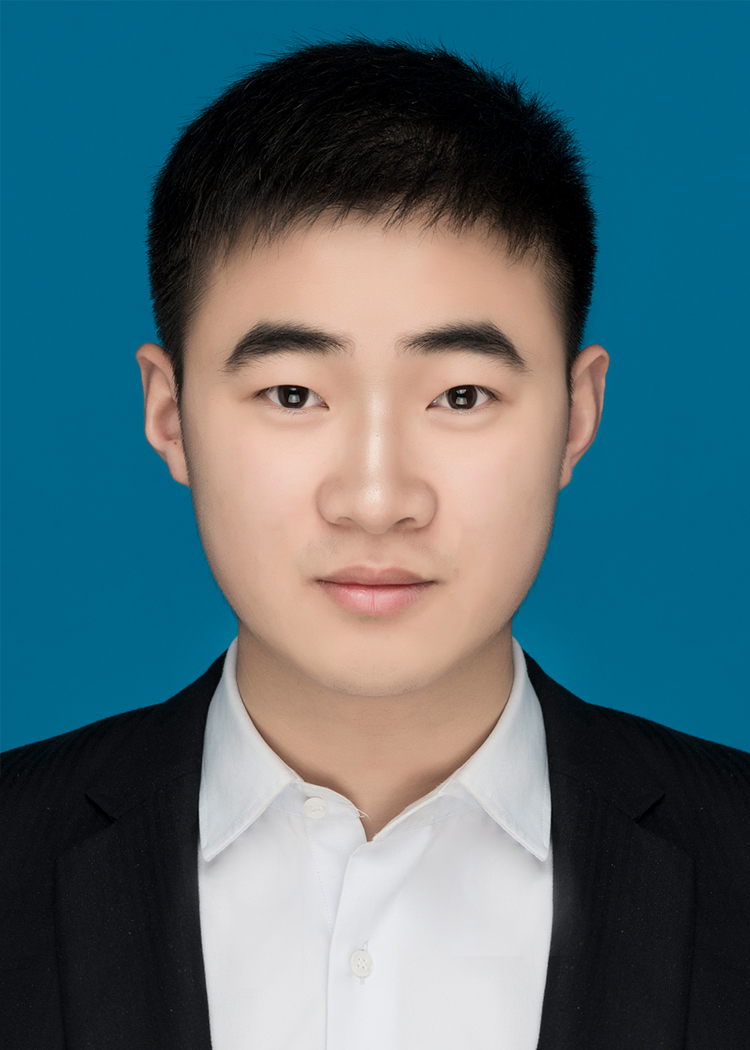}}]{Xuefeng Zhu} received the B.Eng. degree in Internet of Things Engineering from Chongqing University of Posts and Telecommunications, Chongqing, China, in 2017. He received the PhD degree in pattern recognition and intelligent system, from Jiangnan University, Wuxi, China, in 2024. He was a visiting PhD student with the Centre for Vision, Speech and Signal Processing (CVSSP), University of Surrey, Guildford, U.K. from 2022 to 2023.
Currently, he is a lecturer with the School of Artificial Intelligence and Computer Science, Jiangnan University, Wuxi, China. His research interests include visual object tracking and multi-modal learning. 
He has published several scientific papers, including IJCV, IEEE TMM, IEEE TCSVT, PR, AAAI, etc. 
He achieved the 1st place award in the VOT2020 RGB-T challenge (ECCV2020).
\end{IEEEbiography}

\begin{IEEEbiography}[{\includegraphics[width=1in,height=1.25in,clip,keepaspectratio]{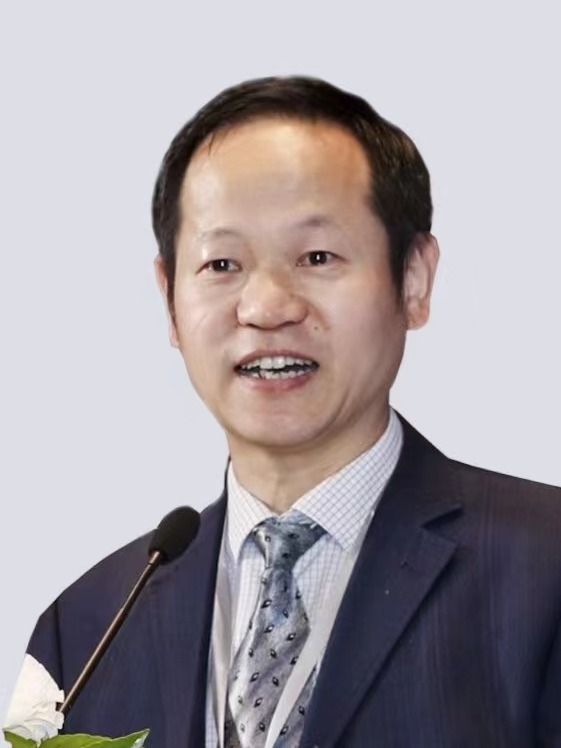}}]{Xiao-Jun Wu} received the BS degree in mathematics from Nanjing Normal University, Nanjing, PR China, in 1991, the MS degree from the Nanjing Univer-sity of Science and Technology, Nanjing, China, in 1996, and the PhD degree in pattern recognition and intelligent system, from the Nanjing University of Science and Technology, in 2002. He was a fellow of United Nations University, International Institute for Software Technology (UNU/IIST) from 1999 to 2000. He joined Jiangnan University in 2006 where he is currently a distinguished professor with the School of Artiﬁcial Intelligence and Computer Science, Jiangnan University. He has published more than 400 papers in his ﬁelds of research. He was a visiting postdoctoral Researcher in the Centre for Vision, Speech, and Signal Processing (CVSSP), University of Surrey, U.K. from 2003 to 2004, under the supervision of Professor Josef Kittler. His current research interests are pattern recognition, computer vision, fuzzy systems, and neural networks. He owned several domestic and international awards because of his research achievements. Currently, he is a fellow of IAPR and AAIA.
\end{IEEEbiography}


\begin{IEEEbiography}[{\includegraphics[width=1in,height=1.25in,clip,keepaspectratio]{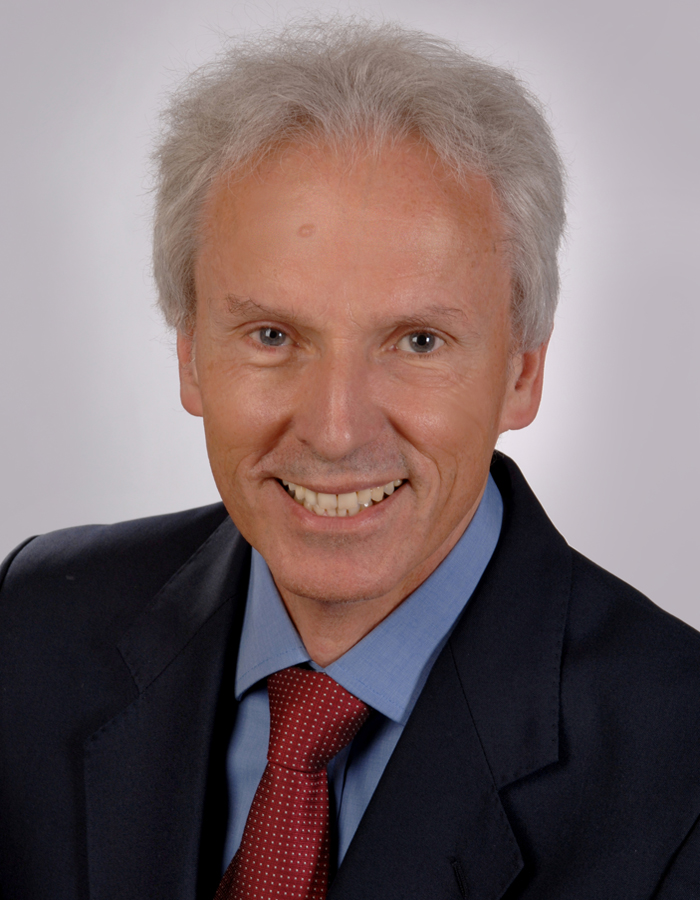}}]{Josef Kittler}(M'74-LM'12) received the B.A., Ph.D., and D.Sc. degrees from the University of Cambridge, in 1971, 1974, and 1991, respectively. He is a distinguished Professor of Machine Intelligence at the Centre for Vision, Speech and Signal Processing, University of Surrey, Guildford, U.K. He conducts research in biometrics, video and image database retrieval, medical image analysis, and cognitive vision. He published the textbook Pattern Recognition: A Statistical Approach and over 700 scientific papers. His publications have been cited more than 78,000 times (Google Scholar).

He is series editor of Springer Lecture Notes on Computer Science. He currently serves on the Editorial Boards of Pattern Recognition Letters, Pattern Recognition and Artificial Intelligence, Pattern Analysis and Applications. He also served as a member of the Editorial Board of IEEE Transactions on Pattern Analysis and Machine Intelligence during 1982-1985. He served on the Governing Board of the International Association for Pattern Recognition (IAPR) as one of the two British representatives during the period 1982-2005, President of the IAPR during 1994-1996.
\end{IEEEbiography}
\end{document}